\documentclass[11pt]{article}
\usepackage{fullpage}
\usepackage{amsmath,amsthm,amsfonts,amssymb,dsfont,bm,tikz,caption}
\usepackage{enumerate,color,xcolor} 
\usepackage[colorlinks,linkcolor=cyan,citecolor=blue]{hyperref}
\usepackage{url} 
\usepackage{scalerel}
\usepackage[normalem]{ulem} 
\usepackage{booktabs}
\usepackage{todonotes}

\numberwithin{equation}{section}
\theoremstyle{plain}

\newtheorem{theorem}{Theorem} 
\newtheorem{corollary}[theorem]{Corollary}

\newtheorem{lemma}[theorem]{Lemma}
\newtheorem{proposition}[theorem]{Proposition}
\newtheorem{assumption}[theorem]{Assumption}

\newcommand{\ep}{\varepsilon}
\newcommand{\nul}{\mathrm{null}}

\begin{document}
\begin{center}  
{\bf\Large Generalized EXTRA stochastic gradient Langevin dynamics}
\end{center}

\author{}
\begin{center}
{Mert G\"{u}rb\"{u}zbalaban}\,
\footnote{Department of Management Science
 and Information Systems, Rutgers Business School, Piscataway, New Jersey, United States of America;
  mg1366@rutgers.edu},
    {Mohammad Rafiqul Islam}\,\footnote{Department of Mathematics, Florida State University, Tallahassee, Florida, United States of America;
  rislam@fsu.edu},
  {Xiaoyu Wang}\,\footnote{FinTech Thrust,
  Hong Kong University of Science and Technology (Guangzhou), Guangzhou, Guangdong, People's Republic of China; 
  xiaoyuwang@hkust-gz.edu.cn},
  Lingjiong Zhu\,\footnote{Department of Mathematics, Florida State University, Tallahassee, Florida, United States of America; zhu@math.fsu.edu
 }
\end{center}

\begin{center}
 \today
\end{center}

\begin{abstract} 
Langevin algorithms are popular Markov Chain Monte Carlo methods for Bayesian learning, particularly when the aim is to sample from the posterior distribution of a parametric model, given the input data and the prior distribution over the model parameters. Their stochastic versions such as stochastic gradient Langevin dynamics (SGLD) allow iterative learning based on randomly sampled mini-batches of large datasets and are scalable to large datasets. However, when data is decentralized across a network of agents subject to communication and privacy constraints, standard SGLD algorithms cannot be applied. Instead, we employ decentralized SGLD (DE-SGLD) algorithms, where Bayesian learning is performed collaboratively by a network of agents without sharing individual data. Nonetheless, existing DE-SGLD algorithms induce a bias at every agent that can negatively impact performance; this bias persists even when using full batches and is attributable to network effects. Motivated by the EXTRA algorithm and its generalizations for decentralized optimization, we propose the generalized EXTRA stochastic gradient Langevin dynamics, which eliminates this bias in the full-batch setting. Moreover, we show that, in the mini-batch setting, our algorithm provides performance bounds that significantly improve upon those of standard DE-SGLD algorithms in the literature. Our numerical results also demonstrate the efficiency of the proposed approach.
\end{abstract}

\section{Introduction}

In our era of big data, the amount of data collected and stored has seen exponential growth with ever-increasing rates. Given the rapid pace at which data are generated, often exceeding our ability to analyze it—-particularly due to limitations in computational resources—-there is a growing interest in developing scalable machine learning algorithms that can efficiently handle large datasets.
Very often, because of communication constraints and privacy constraints, gathering all these data for centralized processing is often impractical or infeasible. Decentralized machine learning algorithms have received a lot of attention for such applications where agents can collaboratively learn a predictive model without sharing their own data but sharing only their local models with their immediate neighbors at some frequency to generate a global model; see e.g. \cite{he2018cola,hendrikx2019accelerated,arjevani2020ideal}. 

Although there is a large body of literature on scaleable first-order decentralized learning methods have been proposed in the literature such as decentralized stochastic approximation and optimization algorithms (see e.g. \cite{uribe2017optimal,gorbunov2019optimal,scaman2019optimal,nedic2020distributed}),
very few of them deal with decentralized Bayesian learning (inference) \cite{PBGG2020,gurbuzbalaban2021decentralized}. With this context, we now introduce the problem of decentralized Bayesian inference.
Assume there are $N$ agents connected over a network $\mathcal{G}=(\mathcal{V},\mathcal{E})$ with $\mathcal{V}=\{1,2,\dots,N\}$ representing the agents and $\mathcal{E}\subseteq \mathcal{V}\times \mathcal{V}$ being the set of edges; i.e. $i$ and $j$ are connected if $(i,j) \in \mathcal{E}$ where the network is undirected, i.e. $(i,j) \in \mathcal{E}$ then $(j,i) \in \mathcal{E}$. 
Let $Z = [z_1,\dots,z_n]$ be a dataset consisting of $n$ independent and identically distributed (i.i.d.) data vectors sampled from a parameterized distribution $p(Z|x)$ where the parameter $x\in \mathbb{R}^{d}$ has a common prior distribution $p(x)$. Due to the decentralization in the data collection, each agent $i$ possesses a subset $Z_i$ of the data where $Z_i = \{z_1^i, z_2^i, \dots, z_{n_i}^i \}$ and $n_i$ is the number of samples of the agent $i$. The data is held disjointly over agents; i.e. $Z = \cup_i Z_i$ with $Z_i \cap Z_j = \emptyset$ for $j\neq i$. The goal is to
sample from the posterior distribution $p(x|Z) \propto p(Z|x) p(x)$.
Since the data points are independent, the log-likelihood function will be additive; $\log p(Z|x) = \sum_{i=1}^N \sum_{j=1}^{n_i} \log p(z_j^i | x)$. Thus, if we set
\begin{equation}\label{def-bayesian-inf} 
f(x) := \sum_{i=1}^N f_i(x), \quad f_i(x) := -\sum_{j=1}^{n_i}\log p\left(z_j^i | x\right) - \frac{1}{N} \log p(x),
\end{equation}
the aim is to sample from the posterior distribution 
with density $\pi(x):= p(x|Z)\propto e^{-f(x)}$, 
where the functions $f_i(x)$ are called \emph{component functions} with $f_i(x)$ being associated to the local data of agent $i$ that is only accessible by the agent $i$. Different choices of the log-likelihood function and therefore the component functions result in different problems, including for example Bayesian linear regression \cite{hoff2009first}, Bayesian logistic regression \cite{hoff2009first}, Bayesian principal component analysis \cite{dubey2016variance} and Bayesian deep learning \cite{wang2016towards,polson2017}. 

Decentralized Langevin algorithms have been proposed in the recent literature
that can be used in the large-scale decentralized sampling problems \cite{PBGG2020,gurbuzbalaban2021decentralized}. 
In this paper, 
we propose and study a new class of Langevin algorithms for decentralized Bayesian inference. For these algorithms, we provide a non-asymptotic convergence analysis alongside numerical experiments. More specifically, our contributions are as follows:

First, inspired by the EXTRA algorithm and its extensions in the decentralized optimization literature \cite{shi2015extra,jakovetic2018unification}, we propose a new algorithm, termed the \emph{generalized EXTRA stochastic gradient Langevin dynamics}, enabling collaborative Bayesian learning across a network of agents without requiring them to share individual data. Our algorithm eliminates the network-induced bias present in existing DE-SGLD algorithms that rely on full-batch processing \cite{gurbuzbalaban2021decentralized}.
We provide non-asymptotic performance guarantees for generalized EXTRA SGLD when each of the components $f_i(x)$ is strongly convex and smooth in which case the target distribution has density $\pi(x) \propto e^{-f(x)}$ where $f$ is strongly convex and smooth. More specifically, we show that the distribution of the iterates $x_i^{(k)}$ converges to a neighborhood of the posterior distribution $\pi(x)$ linearly (geometrically fast in $k$) in the 2-Wasserstein distance with a properly chosen stepsize and communication matrices (Theorem~\ref{thm:main}). We can also show similar results for the averaged iterates $\overline{x}^{(k)} = \frac{1}{N}\sum_{i=1}^N x_i^{(k)}$. Our proof technique relies on analyzing generalized EXTRA SGLD as a perturbed version of the Euler-Maruyama discretization of an overdamped Langevin diffusion
where the perturbation effect is due to the stochastic nature of the gradients and the network effect where agents are only able to communicate with their immediate neighbors. 
The proof technique relies on developing novel bounds on the $L^2$ distance between $x_{i}^{(k)}$ and their average $\overline{x}^{(k)}$, as well as the $L^2$ distance between the average iterate $\overline{x}^{(k)}$ and iterates based on the Euler-Maruyama discretization of overdamped diffusion.

Second, we rigorously compare the iteration complexity of our generalized EXTRA SGLD algorithm to the existing iteration complexity results for the DE-SGLD algorithm, and show improvement by a factor of at least $L$, where $L$ is the smoothness coefficient of $f_{i}$'s (Proposition~\ref{prop:comparison}).

Finally, we provide numerical experiments that illustrate our theory and showcase the practical performance of the EXTRA SGLD algorithm: We show on Bayesian linear regression with synthetic data and Bayesian logistic regression tasks with both synthetic and real data that our method allows each agent to sample from the posterior distribution efficiently without communicating local data. We compare the numerical results of the EXTRA SGLD
with DE-SGLD in the literature and show superior performance.

\section{Preliminaries and Background}

\paragraph{Langevin algorithms.} 
One of the most widely used Markov Chain Monte Carlo methods in statistics are \emph{Langevin algorithms}, 
that allow one to sample from a given density $\pi(x)$ of interest. The classical one is based on the
\emph{overdamped Langevin SDE}; see e.g. \cite{Dalalyan,DM2016,DM2017,dalalyan2019user}:
\begin{equation}\label{eq:overdamped-2}
dX(t)=-\nabla f(X(t))dt+\sqrt{2}dW_{t},
\end{equation}
where $f:\mathbb{R}^{d}\rightarrow\mathbb{R}$
and $W_{t}$ is a standard $d$-dimensional Brownian motion that starts at zero
at time zero. Under some mild assumptions on $f$, the diffusion \eqref{eq:overdamped-2} admits a unique stationary distribution with the density $\pi(x) \propto e^{-f(x)}$,
also known as the \emph{Gibbs distribution} \cite{pavliotis2014stochastic}. For computational purposes, this diffusion is simulated by considering its discretization. 
Although various discretization schemes are proposed, Euler-Maruyama discretization is the simplest one and is known as the unadjusted Langevin algorithm in the literature \cite{DM2017,DM2016}:
\begin{equation}\label{discrete:overdamped}
x_{k+1}=x_{k}-\eta \nabla f(x_k)+\sqrt{2\eta}w_{k+1}\,,
\end{equation}
where $\eta>0$ is the stepsize parameter, and $w_k \in \mathbb{R}^d$ is a sequence of i.i.d. standard Gaussian random vectors $\mathcal{N}(0,I_{d})$. But then the discretized chain \eqref{discrete:overdamped} does not converge to the target $\pi$ and has a bias that needs to be properly characterized to provide performance guarantees \cite{dalalyan2019user}.
The unadjusted Langevin algorithm \eqref{discrete:overdamped} assumes availability of the gradient $\nabla f$. 
On the other hand, in many settings in machine learning, computing the full gradient $\nabla f$ is either infeasible or impractical. For example, in Bayesian regression or classification problems, $f$ can have a finite-sum form as the sum of many component functions over all the data points
and the number of data points can be large (see, e.g., \cite{gurbuzbalaban2021decentralized,xu2018global}). In such settings, algorithms that rely on \emph{stochastic gradients}, i.e., unbiased stochastic estimates of the gradient obtained by a randomized sampling of the data points, is often more efficient \cite{bottou2010large}. This fact motivated the development of Langevin algorithms that can support stochastic gradients. 
In particular, if one replaces the full gradient $\nabla f$ in \eqref{discrete:overdamped} by a stochastic gradient, 
the resulting algorithm
is known as the \emph{stochastic gradient Langevin dynamics} (SGLD) (see, e.g., \cite{welling2011bayesian}).
There has been growing recent interest in the non-asymptotic analysis of Langevin algorithms, motivated by applications to large-scale data analysis and Bayesian inference. The Langevin algorithms admit convergence guarantees to a stationary distribution in a variety of metrics and under various assumptions on $f$; see e.g. \cite{Dalalyan,DM2017,DM2016,CB2018,EH2020,dalalyan2019user,Barkhagen2021,Raginsky,xu2018global,Chau2019,Zhang2019}.

\paragraph{Decentralized setting.}

We consider decentralized algorithm where the agent is connected over a connected network by $N$ nodes, and $W = [W_{ij}] \in \mathbb{R}^{N \times N}$ is a symmetric, doubly stochastic matrix such that, for $i\neq j$, $W_{ij} = W_{ji} > 0$ if $\{i,j\} \in \mathcal{E}$, and $W_{ij} = W_{ji} = 0$ if $\{i,j\} \notin \mathcal{E}$, and $W_{ii} = 1 - \sum_{j \neq i}W_{ij} > 0$. Moreover, we have
$\sigma_{\max}(W - \frac{1}{n}1_N1_N^T) < 1$, where $\sigma_{\max}$ denotes the largest singular value and $1_N \in \mathbb R^N$ is a column vector of ones. We aim to sample from a target distribution with density $\pi(x) \propto e^{-f(x)}$ on $\mathbb{R}^d$ with
$f(x):=\sum_{i=1}^Nf_i(x)$.

In decentralized optimization, decentralized gradient descent (DGD) \cite{nedic2009distributed} carries the following iterative algorithm
\begin{equation}\label{ref-DGD}
x^{(k+1)} = \mathcal{W}x^{(k)} - \eta\nabla F\left(x^{(k)}\right), \quad \mathcal{W} = W \otimes I_d,
\end{equation}
where $
x^{(k)} = \left[\left(x_1^{(k)}\right)^T, \ldots, \left(x_N^{(k)}\right)^T\right]^{T} \in \mathbb{R}^{Nd}$
with $\mathcal{W} = W \otimes I_d$ as the Kronecker product of matrices $W$ and $I_d$ where $I_d$ is a $d\times d$ identity matrix
and $F(x):\mathbb{R}^{Nd} \rightarrow \mathbb{R}$
is defined as:
\begin{equation}
\label{eqn:cap:F}
F(x)= F(x_1,\ldots,x_N) := \sum_{i=1}^Nf_i(x_i),\qquad\text{for any $x=(x_{1},\ldots,x_{N})\in\mathbb{R}^{Nd}$.}
\end{equation}
Let $\mathbf{x}_* := \left[x_*^T,\ldots,x_*^T\right]^T \in \mathbb{R}^{Nd}$, where $x_{\ast}$ is the minimizer of $f(x)$. It satisfies the conditions (1) $\mathbf{x}_{*} = \mathcal{W}\mathbf{x}_{*}$, (2) $1_{Nd}^T \nabla F(\mathbf{x}_{*}) = \sum_{i=1}^N \nabla f_i\left(x_{*}\right) = 0$; these are referred to as \emph{consensus} and \emph{optimality} conditions respectively.

\paragraph{Inexactness and exact algorithms.} 
If we take the limit over $k$ in DGD iterations \eqref{ref-DGD}, we get
\begin{equation}
x^{\infty} = \mathcal{W}x^{\infty} - \eta\nabla F(x^{\infty}),
\end{equation} 
if $x^\infty = \mathbf{x}_{*}$, then we must have $ \mathcal{W}x^{\infty}=x^{\infty}$ by consensus, then we can get
\begin{equation}
x^{\infty} = x^{\infty} - \eta\nabla F(x^{\infty}),
\end{equation}
which means $\nabla F(x^{\infty}) = \mathbf{0}$, i.e. $\nabla f_i\left( x_i^{\infty} \right) = 0$ for every $i$, that implies for any agent $i$, $x_i^{\infty}$ simultaneously minimizes the objective function $f_i$, which is impossible in general. Hence, $x^\infty \neq \mathbf{x}_{*}$ in general and DGD is inexact. Although it is inexact, it is shown $\left\Vert x^{\infty} - \mathbf{x}_* \right\Vert \leq \mathcal{O}(\eta\sqrt{N})$, see~\cite{gurbuzbalaban2021decentralized},~\cite{yuan2016convergence},~\cite{aybat2019universally} and \cite{robust-network-asg}. A decentralized exact first-order algorithm (EXTRA) proposed by~\cite{shi2015extra} can solve the consensus optimization problem and converges to the exact solution. \cite{jakovetic2018unification} unified and generalized this exact distributed first-order algorithm.

\paragraph{Decentralized Langevin algorithms.}

The \emph{decentralized stochastic gradient Langevin dynamics} (DE-SGLD) algorithm \cite{swenson2020distributed,gurbuzbalaban2021decentralized} consists of a weighted averaging with the local variables $x_j^{(k)}$ of node $i$'s immediate neighbors $j\in\Omega_{i}:=\{j : (i,j) \in \mathcal{E}\}$, where
$x_i^{(k)}$ denotes the local variable of node $i$ at iteration $k$, as well as a stochastic gradient step over the node's component function $f_i(x)$, i.e. 
\begin{equation}
x_{i}^{(k+1)}=\sum_{j\in\Omega_{i}}W_{ij}x_{j}^{(k)}-\eta\tilde\nabla f_{i} \left(x_{i}^{(k)}\right)+\sqrt{2\eta}w_{i}^{(k+1)},
\end{equation} 
where $\eta>0$ is the stepsize, $W_{ij}$ are the entries of a doubly stochastic weight matrix $W$ with $W_{ij}>0$ only if $i$ is connected to $j$, $w_{i}^{(k)}$
are independent and identically distributed (i.i.d.) Gaussian random variables with zero mean and identity covariance matrix for every $i$ and $k$, and $\tilde\nabla f_{i} \left(x_{i}^{(k)}\right)$ is an unbiased stochastic estimate of the deterministic gradient $\nabla f_{i} \left(x_{i}^{(k)}\right)$ with a bounded variance. When the number of data points $n_i$ is large, stochastic estimates $\tilde\nabla f_i(x)$ are cheaper to compute compared to actual gradients $\nabla f_i(x)$ and can for instance be estimated from a mini-batch of data, i.e. from randomly selected smaller subsets of data. This allows the DE-SGLD method to be scaleable to big data settings when $n_i$ can be large. 
Without Gaussian noise, iterations are also equivalent to the decentralized stochastic gradient algorithm \cite{swenson-journal,robust-network-asg} which has its origins in the decentralized gradient descent (DGD) methods introduced in \cite{nedic2009distributed}.
\paragraph{Strong convexity and smoothness.} Let $\mathcal{S}_{\mu,L}(\mathbb{R}^d)$ denote the set of functions from $\mathbb{R}^d$ to $\mathbb{R}$ that are $\mu$-strongly convex and $L$-smooth, that is, for any $g \in \mathcal{S}_{\mu,L}(\mathbb{R}^d)$, it holds that 
\begin{equation}
\frac{L}{2}\left\Vert x-y \right\Vert^2 \geq g(x) - g(y) - \nabla g(y)^T(x-y) \geq \frac{\mu}{2}\left\Vert x-y\right\Vert^2, \quad \text{for every $x,y \in \mathbb R^d$}. 
\end{equation}

\paragraph{Notations.} 
Define $\mathcal{P}_{2}(\mathbb{R}^{d})$
as the space consisting of all the Borel probability measures $\mu$
on $\mathbb{R}^{d}$ with the finite second moment
(based on the Euclidean norm).
For any $\mu_{1},\mu_{2}\in\mathcal{P}_{2}(\mathbb{R}^{d})$, 
the $2$-Wasserstein
distance $\mathcal{W}_{2}$ (see e.g. \cite{villani2008optimal}) between $\mu_{1}$ and $\mu_{2}$ is defined as:
$\mathcal{W}_{2}(\mu_{1},\mu_{2}):=\left(\inf\mathbb{E}\left[\Vert Y_{1}-Y_{2}\Vert^{2}\right]\right)^{1/2},$
where the infimum is taken over all joint distributions of the random variables $Y_{1},Y_{2}$ with marginal distributions
$\mu_{1},\mu_{2}$ respectively. For any $x,y\in\mathbb{R}$, denote
$x\vee y:=\max\{x,y\}$ and $x\wedge y:=\min\{x,y\}$. For any random variable $X$, denote 
$\mathcal{L}(X)$ the law of $X$.

\section{EXTRA Langevin Algorithms} 

We aim to sample from a target distribution with density $\pi(x) \propto e^{-f(x)}$ on $\mathbb{R}^d$ with
$f(x):=\sum_{i=1}^Nf_i(x)$. Now we make the first assumption on the objective function.
\begin{assumption}
\label{assumption:f}
We assume the component functions $f_i$ are $\mu$-strongly convex and $L$-smooth with $L > \mu$, i.e. $f_{i}\in\mathcal{S}_{\mu,L}(\mathbb{R}^d)$ for every $i=1,2,\ldots,N$.
\end{assumption}
\noindent Under Assumption~\ref{assumption:f}, it follows
that $F$ is also $\mu$-strongly convex and $L$-smooth
where we recall from the definition in~\eqref{eqn:cap:F} such that $F:\mathbb{R}^{Nd}\rightarrow\mathbb{R}$ with $F(x_1, x_2,\ldots, x_N) = \sum_{i=1}^Nf_i(x_i)$ for any $x=(x_{1},\ldots,x_{N})\in\mathbb{R}^{Nd}$. 

We propose \emph{EXTRA stochastic gradient Langevin dynamics} (EXTRA SGLD) to target $\pi$ that is defined as follows 
\begin{align}
x_i^{(k+2)} & = \sum_{j \in \Omega_i}W_{ij}x_j^{(k+1)} - \eta \widetilde\nabla f_i\left(x_i^{(k+1)}\right)  + \sqrt{2\eta}w_i^{(k+2)},
\label{extra:alg:1}
\\
x_i^{(k+1)} & = \sum_{j \in \Omega_i}\widetilde{W}_{ij}x_j^{(k)} - \eta \widetilde\nabla f_i\left(x_i^{(k)}\right) + \sqrt{2\eta}w_i^{(k+1)},
\label{extra:alg:2}
\end{align}
where $w_i^{(k)}$ are standard $d$-dimensional Gaussian random vectors that are i.i.d.
in both $i=1,2,\ldots,N$ and $k=1,2,3,\ldots$. 
In this algorithm, $x_i^{(k)}$ denotes the local variable of node $i$ at iteration $k$ for every node $i=1,2,\ldots,N$ and iteration $k=0,1,2,\ldots$. At the iteration $k$, node $i$ accesses $\widetilde{\nabla}f_i\left(x_i^{(k)}\,,\,z_i^{(k)}\right)$ where $z_i^{(k)}$ is a random variable independent of $\left\{z_j^{(t)} \right\}_{j = 1,2,\ldots,i-1,i+1,\ldots,N;\,t = 1,\ldots,k-1}$. We let $\widetilde \nabla f_i\left(x_i^{(k)}\right)$ denote $\widetilde \nabla f_i\left(x_i^{(k)}\,,\,z_i^{(k)}\right)$ and define the \emph{gradient noise} as 
\begin{equation}
\label{eqn:grad:noise}
\xi_i^{(k)} := \widetilde \nabla f_i\left(x_i^{(k)}\right) - \nabla f_i\left(x_i^{(k)}\right), \quad i = 1,2,\ldots,N,
\end{equation}
and we assume the stochastic gradient noise satisfies the following assumption. 
\begin{assumption}
\label{assumption:noise}
For every $i=1,2,\ldots,N$ and $k=0,1,2,\ldots$, the gradient noise defined in~\eqref{eqn:grad:noise} is conditionally unbiased with a finite second moment such that
\begin{equation}
\mathbb E\left[\xi_i^{(k+1)} \bigg\vert \mathcal F_k\right] = 0,\qquad \mathbb E\left\Vert \xi_i^{(k+1)} \right\Vert^2 \leq \sigma^2,
\end{equation}
where $\mathcal{F}_k$ is the natural filtration of the iterates $\left(x_i^{(k)}\right)_{i=1}^{N},\left(z_{i}^{(k)}\right)_{i=1}^{N}$ up to (and including) time $k$.
\end{assumption}

\noindent Then, we re-formulate EXTRA stochastic gradient Langevin dynamics as follows.
\begin{align}
x_i^{(k+2)} & = \sum_{j \in \Omega_i}W_{ij}x_j^{(k+1)} - \eta \nabla f_i\left(x_i^{(k+1)}\right) - \eta \xi_i^{(k+1)} + \sqrt{2\eta}w_i^{(k+2)},
\\
x_i^{(k+1)} & = \sum_{j \in \Omega_i}\widetilde{W}_{ij}x_j^{(k)} - \eta \nabla f_i\left(x_i^{(k)}\right)  - \eta \xi_i^{(k)} + \sqrt{2\eta}w_i^{(k+1)}.
\end{align}
These updates for $N$ agents can also be expressed as
\begin{align}
x^{(k+2)} & = \mathcal{W}x^{(k+1)} - \eta \nabla F\left(x^{(k+1)}\right) - \eta \xi^{(k+1)} + \sqrt{2\eta}w^{(k+2)},\label{EXTRA:matrix:1}
\\
x^{(k+1)} & = \widetilde{\mathcal{W}}x^{(k)} - \eta \nabla F\left(x^{(k)}\right)  - \eta \xi^{(k)} + \sqrt{2\eta}w^{(k+1)},\label{EXTRA:matrix:2}
\end{align}
where $\mathcal{W} = W \otimes I_d$, $\widetilde{\mathcal{W}} = \widetilde{W} \otimes I_d$, $
x^{(k)} = \left[\left(x_1^{(k)}\right)^T, \ldots, \left(x_N^{(k)}\right)^T\right]^{T} \in \mathbb{R}^{Nd}$ and 
$$
w^{(k)} = \left[\left(w_1^{(k)}\right)^T, \ldots, \left(w_N^{(k)}\right)^T\right]^{T},\quad k=0,1,2,\ldots,
$$
and we assume that the mixing matrices $W, \widetilde W$ satisfy the following assumption. Such assumptions are made for analyzing the EXTRA methods and its generalizations \cite{shi2015extra,jakovetic2018unification}.

\begin{assumption}
\label{assumption:mixing}
Consider a connected network $\mathcal{G}=(\mathcal{V}, \mathcal{E})$ consisting of a set of agents $\mathcal{V}=\{1,2, \cdots, n\}$ and a set of undirected edges $\mathcal{E}$. The doubly stochastic matrices $W=\left[W_{i j}\right] \in \mathbb{R}^{N \times N}$ and $\widetilde{W}=\left[\widetilde{W}_{i j}\right] \in \mathbb{R}^{N \times N}$ satisfy

(1) Null space property: 
$$
\nul \left\{W-\widetilde{W}\right\}=\operatorname{span}\{1_N\},\qquad \nul\left\{I_{N}-\widetilde{W}\right\} \supseteq \operatorname{span}\{1_N\},
$$
where $\operatorname{span}\{1_N\}$ is the span of the vector space supported by all-one vector $\left[1^T_N,1^T_N,\ldots,1^T_N\right]$.

(2) Spectral property:
$$
\widetilde{W} \succ 0, \qquad \frac{I_{N}+W}{2} \succcurlyeq \widetilde{W} \succcurlyeq W.
$$
\end{assumption}

The assumption implies $W \succ -I_{N}$ and $\frac{I_{N}+W}{2} \succcurlyeq W$, so the eigenvalues of $W$ lie in $(-1,1]$ and the eigenvalues of $\widetilde{W}$ lie in $(0,1]$. We will only consider $\widetilde{W} \neq W$; if $\widetilde{W} = W$, then the EXTRA SGLD iterate in~\eqref{alg} reduces to DE-SGLD algorithm studied by~\cite{gurbuzbalaban2021decentralized}. 
We assume that 
\begin{equation}
\label{choice:tilde:W}
\widetilde{W} = hI_N + \left(1-h\right)W, \quad\text{$h \in (0,1/2]$}.
\end{equation}
Note that the definition of $\widetilde{W}$ satisfies Assumption~\ref{assumption:mixing}, where we can compute that $h(I_{N} - W) \succcurlyeq 0$, which implies $hI_{N} + \left(1-h\right)W \succcurlyeq W$, and it is clear that $\frac{I_N + W}{2} \succcurlyeq hI_{N} + \left(1-h\right)W$ with $h \leq 1/2$.

In the noiseless case, EXTRA has a primal-dual interpretation as a gradient descent ascent on a particular energy function \cite{jakovetic2018unification}.
EXTRA was proposed by~\cite{shi2015extra}, and its unification and generalization was studied by~\cite{jakovetic2018unification} to solve the dual optimization problem, where the author showed the iterates converges to the exact optimal solution if the parameters are chosen appropriately.
Motivated by this work for decentralized optimization, we aim to propose a generalized EXTRA stochastic gradient Langevin dynamics which can produce the exact target distribution. We can use some algebraic transformation to generalize EXTRA SGLD algorithm in~\eqref{EXTRA:matrix:1}-\eqref{EXTRA:matrix:2}. 
By subtracting \eqref{EXTRA:matrix:2} from \eqref{EXTRA:matrix:1}, the updating iterates follow
\begin{align}
 x^{(k+2)} - x^{(k+1)} 
& = \mathcal{W}x^{(k+1)} - \widetilde{\mathcal{W}}x^{(k)} 
 - \eta\left(\nabla F\left(x^{(k+1)}\right) - \nabla F\left(x^{(k)}\right)\right) 
\nonumber
\\
&\qquad\qquad\qquad\qquad - \eta\left(\xi^{(k+1)} - \xi^{(k)}\right) + \sqrt{2\eta}\left(w^{(k+2)}-w^{(k+1)}\right),\label{after:subtract}
\end{align}
where $\xi^{(k)} = \left[\left(\xi_{1}^{(k)}\right)^T,\left(\xi_{2}^{(k)}\right)^T,\ldots,\left(\xi_{N}^{(k)}\right)^T\right]^T$
for every $k$. Next, we sum up the subtraction terms $\left(x^{(2)} - x^{(1)}\right), \left(x^{(3)} - x^{(2)}\right), \ldots, \left(x^{(k+2)} - x^{(k+1)}\right)$ in \eqref{after:subtract}, and by telescopic cancellation, we obtain
\begin{equation}
\label{alg}
x^{(k+2)} = \mathcal{W}x^{(k+1)} - \eta\nabla F\left(x^{(k+1)}\right) - \eta \xi^{(k+1)} + \sum_{h=0}^{k}\left(\mathcal W - \widetilde{\mathcal W}\right)x^{(h)} + \sqrt{2\eta}w^{(k+1)},
\end{equation}
and equivalently, 
\begin{equation}
x^{(k+2)} = \widetilde{\mathcal{W}}x^{(k+1)} - \eta\nabla F\left(x^{(k+1)}\right) - \eta \xi^{(k+1)} - \sum_{h=0}^{k+1}\left(\widetilde{\mathcal W}-\mathcal W\right)x^{(h)} + \sqrt{2\eta}w^{(k+1)},
\end{equation}
provided $x^{(1)} = \mathcal W x^{(0)} - \eta \nabla f\left(x^{(0)}\right) - \eta\xi^{(0)} + \sqrt{2\eta}w^{(1)}$. Since $U=\widetilde W - W$ is positive semi-definite, we are able to have the following matrix decomposition $U^{1/2} = PD^{1/2}P^T$, where $D$ is diagonal with non-negative diagonal entries
and $P$ is an orthogonal matrix. Thus, we can introduce an auxiliary sequence as follows:
\begin{equation}
\label{def:q}
q^{(k)} = \mathcal{U}^{1/2}\sum_{h=0}^{(k)}x^{(h)}, \quad \mathcal U = U \otimes I_d, \quad U = \widetilde W - W \in \mathbb R^{N \times N}.
\end{equation}
Moreover, it is easy to observe that 
\begin{equation}
q^{(k+1)} = q^{(k)} + \mathcal{U}^{1/2}x^{(k+1)}.
\end{equation}
Thus, we obtain the following $2(Nd)$-dimensional recursive expression:
\begin{align}
x^{(k+1)} & = x^{(k)}  - \eta\left(\frac{1}{\eta}\left(I_{Nd} - \widetilde{\mathcal{W}}\right)x^{(k)} + \nabla F\left(x^{(k)}\right) + \xi^{(k)} + \frac{1}{\eta}\mathcal{U}^{1/2}q^{(k)}\right) + \sqrt{2\eta}w^{(k+1)}, \label{upt1}\\
q^{(k+1)} & = q^{(k)} + \mathcal{U}^{1/2}x^{(k+1)}, \quad \mathcal{U=\widetilde{W}-W}.
\label{upt2}
\end{align}
By denoting 
\begin{equation}
\label{def:v}
v^{(k)} = \frac{1}{\eta}\mathcal{U}^{1/2}q^{(k)}, \quad k=0,1,2,\ldots,
\end{equation} 
we get from~\eqref{upt1} that
\begin{equation}
v^{(k)}+\nabla F\left(x^{(k)}\right)+\xi^{(k)} - \frac{1}{\eta} \widetilde{\mathcal{W}} x^{(k)} - \sqrt{\frac{2}{\eta}}w^{(k+1)} = -\frac{1}{\eta} x^{(k+1)}.
\end{equation}
Moreover, we can compute from~\eqref{upt2} and~\eqref{def:v} that
\begin{align}
\label{eqn:mid:v}
v^{(k+1)} - v^{(k)} = \frac{1}{\eta}\mathcal{U}^{1/2}q^{(k+1)} - \frac{1}{\eta}\mathcal{U}^{1/2}q^{(k)}
= \frac{1}{\eta}\mathcal{U}^{1/2}\left(q^{(k)} + \mathcal{U}^{1/2}x^{(k+1)}\right) - \frac{1}{\eta}\mathcal{U}^{1/2}q^{(k)}
= \frac{1}{\eta}\mathcal{U}x^{(k+1)}.
\end{align}
Hence, we can re-write~\eqref{upt2} as the following update:
\begin{align}
v^{(k+1)} = v^{(k)} + \frac{1}{\eta}\mathcal{U}x^{(k+1)},
 = v^{(k)} -\mathcal{U}\left(v^{(k)}+\nabla F\left(x^{(k)}\right)+\xi^{(k)} - \frac{1}{\eta} \widetilde{\mathcal{W}} x^{(k)} - \sqrt{\frac{2}{\eta}}w^{(k+1)}  \right).
\end{align}

\noindent We introduce the \emph{generalized EXTRA stochastic gradient Langevin dynamics} as follows
\begin{align}
x^{(k+1)} & = \widetilde{\mathcal{W}}x^{(k)}  - \eta\left(\nabla F\left(x^{(k)}\right)  + v^{(k)}\right) - \eta\xi^{(k)} + \sqrt{2\eta}w^{(k+1)}, \label{iter1}\\
v^{(k+1)} & =  v^{(k)} -\mathcal{U}\left(v^{(k)} + \nabla F\left(x^{(k)}\right) - \mathcal{B}x^{(k)} \right) - \mathcal{U}\xi^{(k)} + \mathcal{U}\sqrt{\frac{2}{\eta}}w^{(k+1)},\, \quad \mathcal{U=\widetilde{W}-W}.
\label{iter2}
\end{align}
We can observe from the iterates \eqref{iter1}-\eqref{iter2} that if we choose $\mathcal{U} = 0_N \otimes I_d$, then it reduces iterative updates to the DE-SGLD algorithm in~\cite{gurbuzbalaban2021decentralized}. We will study the algorithm with the matrix $\widetilde{\mathcal{W}} = \widetilde{W} \otimes I_d$ defined in~\eqref{choice:tilde:W}, 
and the matrix $\mathcal{B} = B \otimes I_d$ has the property 
\begin{equation}
1^T_NB = c \quad\text{with $c \in \mathbb{R}$.} 
\end{equation}
The corresponding deterministic optimization algorithm without gradient noise was studied in~\cite{shi2015extra} and~\cite{jakovetic2018unification}\footnote{We note~\cite{jakovetic2018unification} exchanged the notations $\widetilde{\mathcal{W}}$ and $\mathcal{W}$ to get their Equations~(16)-(17) and Lemma~3. In their notation, they have $\mathcal{L} := I_{Nd}-\mathcal{W} = \mathcal{W - \widetilde{W}}$ by using $W = \frac{I_{N} + \widetilde{W}}{2}$ in their algorithm, it corresponds to $\mathcal{U} = \widetilde{\mathcal{W}} - \mathcal{W}$ in ours. In their algorithm, they only have the notation $\mathcal{W}$, so they further denote $\widetilde{\mathcal{W}} = \mathcal{W - J}$ in their proofs which is different from our choice on $\widetilde{W}$ in~\eqref{choice:tilde:W}.} with $\widetilde{W} = \frac{I_{N}+W}{2}$,  which corresponds to our choice on $\widetilde{W}$ when $h = 1/2$.~\cite{gurbuzbalaban2021decentralized} studied decentralized SGLD, it corresponds to let $h = 0$ in our algorithm, that is $U = 0_{N}$. We also note that by taking $\mathcal{B} = \widetilde{\mathcal{W}}/\eta$, the algorithm \eqref{iter1}-\eqref{iter2} reduces to EXTRA SGLD algorithm in~\eqref{extra:alg:1}-\eqref{extra:alg:2}. In particular,~\cite{jakovetic2018unification} considered the case $\mathcal{B} = b I_{Nd}$ with $b > 0$.

\section{Convergence Analysis}

In this section, we provide the main results of the paper. 
Our non-asymptotic convergence analysis provides
the convergence guarantees for
the 2-Wasserstein distance between 
the law of the average of the iterates $\overline{x}^{(K)}$
and the target distribution $\pi$, 
as well as the average of the 2-Wasserstein distance
between the law of the individual iterates $x_{i}^{(K)}$
and the target distribution $\pi$.

\begin{theorem}\label{thm:main}
Consider the generalized EXTRA Langevin dynamics with the network averaging matrix $\widetilde{W} = hI_N + \left(1-h\right)W$ where 
\begin{equation}
0 < h \leq \frac{1-\overline{\gamma}_{{\scaleto{W}{3pt}}}}{4\overline{\gamma}_{{\scaleto{I_{N}-W}{5pt}}}^2}\wedge\frac{1}{2} \wedge \frac{1}{\gamma_1\gamma_2},
\end{equation}
and assume that the stepsize $\eta$ is chosen satisfying
\begin{equation}
0<\eta < \frac{1}{h\gamma_1\gamma_2} \wedge \frac{\gamma_{{\scaleto{\widetilde{W}}{5pt}}}}{6(L+\mu) \vee 2A}\wedge 1\wedge\frac{1}{L+\mu}\wedge \frac{\gamma_{{\scaleto{\widetilde{W}}{5pt}}}}{6(L + \mu)},
\end{equation} 
where $\gamma_1, \gamma_2, \gamma_{{\scaleto{\widetilde{W}}{5pt}}}, \overline{\gamma}_{{\scaleto{I_{N}-W}{5pt}}}^2$ are constants defined in Table~\ref{table_constants}. Then, for any $K\geq K_{0}$, the following bound holds: 
\begin{align}
\label{decompose:t2:main}
\mathcal{W}_2\left(\mathcal{L}\left(\overline{x}^{(K)}\right)\,,\,\pi \right) & \leq  \left(\frac{\overline{\gamma}_{{\scaleto{\widetilde{W}}{5pt}}}^{2 K}-\left(1-\eta \mu\left(1-\frac{\eta L}{2}\right)\right)^K}{\overline{\gamma}_{{\scaleto{\widetilde{W}}{5pt}}}^2-1+\eta \mu\left(1-\frac{\eta L}{2}\right)}\right)^{1 / 2} \frac{2 L \overline{\gamma}_{{\scaleto{\widetilde{W}}{5pt}}}}{\sqrt{N}}\left(\mathbb{E}\left\|x^{(0)}\right\|^2\right)^{1 / 2}
\nonumber 
\\
& \qquad\qquad
 + (1-\mu \eta)^{K} \mathcal{W}_2\left(\mathcal{L}\left(x_0\right), \pi\right) + \sqrt{\eta}\mathcal{E}_1,
\end{align}
where we have
\begin{align}
\label{defn:mathcal:E:1}
\mathcal{E}_1:=& \left(\frac{\eta}{\mu\left(1-\frac{\eta L}{2}\right)}+\frac{(1+\eta L)^2}{\mu^2\left(1-\frac{\eta L}{2}\right)^2}\right)^{1 / 2} \cdot\left(\frac{4 L^2 \left(R_h + R_h'\right) \eta}{N(1-\overline{\gamma}_{{\scaleto{\widetilde{W}}{5pt}}})^2}+\frac{4 L^2 \sigma^2 \eta}{1-\overline{\gamma}_{{\scaleto{\widetilde{W}}{5pt}}}^2}+\frac{8 L^2 d}{1-\overline{\gamma}_{{\scaleto{\widetilde{W}}{5pt}}}^2}\right)^{1 / 2} 
\nonumber 
\\
& \qquad +\frac{\sigma}{\sqrt{\mu\left(1-\frac{\eta L}{2}\right) N}}+\frac{1.65 L}{\mu} \sqrt{d N^{-1}}.
\end{align}
Moreover,
\begin{align}
& \frac{1}{N}\sum_{i=1}^N\mathcal{W}_2\left(\mathcal{L}\left(x_i^{(K)}\right)\,,\, \pi \right) 
\nonumber 
\\
& \, \leq \eta \cdot \frac{D_1}{\sqrt{N}} + \sqrt{\eta} \cdot \left(D_2 + \mathcal{E}_1\right) + \left(\frac{\overline{\gamma}_{{\scaleto{\widetilde{W}}{5pt}}}^{2 K}-\left(1-\eta \mu\left(1-\frac{\eta L}{2}\right)\right)^K}{\overline{\gamma}_{{\scaleto{\widetilde{W}}{5pt}}}^2-1+\eta \mu\left(1-\frac{\eta L}{2}\right)}\right)^{1 / 2} \frac{2 L \overline{\gamma}_{{\scaleto{\widetilde{W}}{5pt}}}}{\sqrt{N}}\left(\mathbb{E}\left\|x^{(0)}\right\|^2\right)^{1 / 2}
\nonumber 
\\
& \qquad\qquad\qquad\qquad\qquad + (1-\mu \eta)^{K} \mathcal{W}_2\left(\mathcal{L}\left(x_0\right), \pi\right)
+ \frac{2\left(\overline{\gamma}_{{\scaleto{\widetilde{W}}{5pt}}}\right)^{K} }{\sqrt{N}}\sqrt{\mathbb{E}\left[ \left\Vert x^{(0)}\right\Vert^2 \right]},
\end{align}
where the constants $R_h$ and $R_h'$ are made explicit and given in Table~\ref{table_constants}.
\end{theorem}

\section{Comparison with DE-SGLD}

In this section, we are interested in comparing our generalized EXTRA SGLD method
 with the DE-SGLD method in the literature \cite{gurbuzbalaban2021decentralized}. In particular, we highlight the dependence
 on strong-convexity constant $\mu$, the smoothness constant $L$, the dimension $d$ and the accuracy level $\varepsilon$.
 We have the following proposition.

\begin{proposition}\label{prop:comparison}
For DE-SGLD, under the assumptions in Theorem~1 in~\cite{gurbuzbalaban2021decentralized}, as $\varepsilon\rightarrow 0$,
\begin{equation}
\frac{1}{N} \sum_{i=1}^N \mathcal{W}^{\mathrm{de-sgld}}_2\left(\mathcal{L}\left(x_i^{(K)}\right), \pi\right) 
\leq\mathcal{O}(\varepsilon),
\end{equation}
provided that
\begin{align}
\label{result:sgld}
K\geq K^{\mathrm{de-sgld}} & =\tilde{\mathcal{O}}\left(\frac{L^4d}{\varepsilon^{2}\mu^{3}}\right)
\end{align}
where $\tilde{\mathcal{O}}$ hides the logarithmic dependence on $\varepsilon$.

For generalized EXTRA SGLD, under the assumptions in Theorem~\ref{thm:main}, as $\ep \rightarrow 0$, by taking $h\geq\Omega(\eta\mu)$ and
$h< \frac{1}{(L/\mu)^4(L + \Vert B \Vert^2)} \wedge \frac{1}{2\gamma_1\gamma_2}$,
it holds that
\begin{equation}
\frac{1}{N} \sum_{i=1}^N \mathcal{W}^{\mathrm{extra-sgld}}_2\left(\mathcal{L}\left(x_i^{(K)}\right), \pi\right) 
\leq\mathcal{O}(\varepsilon),
\end{equation}
provided that
\begin{equation}\label{result:extra}
K\geq K^{\mathrm{extra-sgld}}= \tilde{\mathcal{O}}\left(\frac{L^2d}{\ep^2\mu^3}\right),
\end{equation}
where $\tilde{\mathcal{O}}$ hides the logarithmic dependence on $\varepsilon$ and the constants $\gamma_1,\gamma_2$ are provided in Table~\ref{table_constants}.
\end{proposition} 

By comparing the complexities of DE-SGLD in~\eqref{result:sgld} to generalized EXTRA SGLD in~\eqref{result:extra}, we find that generalized EXTRA SGLD achieves an improvement on the order of $\tilde{\mathcal{O}}(L^2)$. When $\mu$ is large (and therefore $L$ is large), the Gibbs distribution \(\pi \propto e^{-f}\) becomes concentrated around the minimizer of \(f\), making the sampling problem approximately equivalent to the global optimization problem \(\min_{x \in \mathbb{R}^d} f(x)\). In the decentralized optimization setting, EXTRA is known to improve upon decentralized gradient descent, and the improvement achieved with generalized EXTRA SGLD in the sampling setting is analogous. Intuitively, DE-SGLD introduces a bias, which is evident in the constant \(D\) in~\eqref{sgld:D} that results from the agents' gradient updates. The generalized EXTRA SGLD algorithm corrects this bias in the agents' gradient updates, leading to improved performance.



\begin{table}{\emph{Constants} \hfill \emph{Source}}

\centering

\rule{\textwidth}{\heavyrulewidth}

\vspace{-1.6\baselineskip}

\begin{flalign*}
&
\gamma_{{\scaleto{\widetilde{W}}{5pt}}}
=\vert \lambda_2^{{\scaleto{\widetilde{W}}{5pt}}} \vert^2 1_{0 < \vert \lambda_2^{{\scaleto{\widetilde{W}}{5pt}}} \vert^2 < \frac{1}{2}}
+\frac{\vert \lambda_2^{{\scaleto{\widetilde{W}}{5pt}}} \vert^2(\vert \lambda_2^{{\scaleto{\widetilde{W}}{5pt}}} \vert^2 - \frac{1}{2})}{1 - \vert \lambda_2^{{\scaleto{\widetilde{W}}{5pt}}} \vert^2} 1_{\frac{1}{2} \leq \vert \lambda_2^{{\scaleto{\widetilde{W}}{5pt}}} \vert^2 \leq \frac{2}{3}}
+\frac{5\vert \lambda_2^{{\scaleto{\widetilde{W}}{5pt}}}\vert^2 - 3\vert \lambda_2^{{\scaleto{\widetilde{W}}{5pt}}} \vert^4-2}{3\vert \lambda_2^{{\scaleto{\widetilde{W}}{5pt}}} \vert^2-1}
 1_{\frac{2}{3} \leq \vert \lambda_2^{{\scaleto{\widetilde{W}}{5pt}}}\vert^2 < 1}
&&&\eqref{const:gamma:tilde:W}
\end{flalign*}

\vspace{-1.6\baselineskip}
\begin{flalign*}
&
A= \left(\frac{L}{\mu} - 1 + \frac{\gamma_{{\scaleto{\widetilde{W}}{5pt}}}}{2(1+\mu/L)}\right)\cdot \frac{4L^2}{N^2}\left(1 + \frac{2 + 2L}{\mu} \right)
&&&\eqref{const:A}
\end{flalign*}

\vspace{-1.6\baselineskip}
\begin{flalign*}
&
\overline{\gamma}_{{\scaleto{I_{N}-W}{5pt}}}=\max\left\{1 - \left\vert \lambda_2^W \right\vert, 1-\left\vert \lambda_N^W \right\vert \right\}, 
\quad 
\overline{\gamma}_{{\scaleto{W}{3pt}}}:=\max\left\{\left\vert \lambda_2^W \right\vert, \left\vert \lambda_N^W \right\vert \right\}
&&&\eqref{defn:bar:gamma}
\end{flalign*}

\vspace{-1.6\baselineskip}
\begin{flalign*}
&
\gamma_1 = \frac{1}{\gamma_{{\scaleto{\widetilde{W}}{5pt}}}}\left(\frac{1}{L} + 2 + \frac{1}{L\mu}\right), 
\quad
\gamma_2 = \frac{12\left(L^2 + L\left\Vert B \right\Vert^2 \right)}{(1-\overline{\gamma}_{{\scaleto{W}{3pt}}})\left(1 - \overline{\gamma}_{{\scaleto{I_{N}-W}{5pt}}}^2\right)}\left(1 + \frac{4L^2\left(1 + \frac{2 + 2L}{\mu} \right)}{N^2\mu}\right)
&&\eqref{defn:gamma:1:2}
\end{flalign*}

\vspace{-1.6\baselineskip}
\begin{flalign*}
&
w_1 = 2\left(\frac{N^2 + 1}{\gamma_{{\scaleto{\widetilde{W}}{5pt}}}} + \frac{4}{\gamma_{{\scaleto{\widetilde{W}}{5pt}}}}\cdot\left(\frac{L}{\mu} + 3\eta L - 1\right) \right), 
\quad
w_2 =\frac
{8\left(6\left(L^2 + L\left\Vert B \right\Vert^2 \right)+N^{2}\mu\right)}{N\mu(1-\overline{\gamma}_{{\scaleto{W}{3pt}}})\left(1 - \overline{\gamma}_{{\scaleto{I_{N}-W}{5pt}}}^2\right)}
&&&\eqref{defn:w:1},\eqref{defn:w:2}
\end{flalign*}

\vspace{-1.6\baselineskip}
\begin{flalign*}
&E_1 = \frac{8}{\gamma_{{\scaleto{\widetilde{W}}{5pt}}}}\left(L/\mu + 3\eta L - 1\right), 
\quad E_2 = \frac{2}{\gamma_{{\scaleto{\widetilde{W}}{5pt}}}}
&&&\eqref{defn:E:1:2}
\end{flalign*}

\vspace{-1.6\baselineskip}
\begin{flalign*}
&E_3 = \frac{12\left(L^2 + L\left\Vert B \right\Vert^2 \right)}{\mu(1-\overline{\gamma}_{{\scaleto{W}{3pt}}})\left(1 - \overline{\gamma}_{{\scaleto{I_{N}-W}{5pt}}}^2\right)} , 
\quad E_4 = \frac{4}{(1-\overline{\gamma}_{{\scaleto{W}{3pt}}})\left(1 - \overline{\gamma}_{{\scaleto{I_{N}-W}{5pt}}}^2\right)}
&&&\eqref{defn:E:3:4}
\end{flalign*}

\vspace{-1.6\baselineskip}
\begin{flalign*}
&R_h=   h\delta^2\left(\frac{C_1\gamma_2}{2L^2} +  \frac{C_0\gamma_1\gamma_2}{2L^2} \right) + (h/\eta)\delta^{2}\left( \gamma_2D_0 + \frac{w_2}{N}\left(\eta\sigma^2 + 2d\right) \right) + \left\Vert \nabla F(\mathbf{x}_*) \right\Vert^2
&&&\eqref{R:h}
\end{flalign*}

\vspace{-1.6\baselineskip}
\begin{flalign*}
&R_h'= \eta\delta^2\left(C_1 + C_3 + \gamma_1C_0 + D_0C_2\right) + \delta^2\eta^2\left(\frac{C_1C_2}{2L^2} + \frac{\gamma_1C_0C_2}{2L^2}\right) 
 + 3\left\Vert \nabla F(\mathbf{x}_*)\right\Vert^2
&&&\eqref{R:h:prime}
\end{flalign*}

\vspace{-1.6\baselineskip}
\begin{flalign*}
&K_0=\frac{\delta^2}{1 - \delta^2}\left[\left(1 - \frac{\left\Vert \nabla F(\mathbf{x}_*) \right\Vert^2}{D_0 + C_4}\right)  \vee \left(1 - \frac{\left\Vert \nabla F(\mathbf{x}_*) \right\Vert^2}{C_0}\right)  \right] \vee 0
&&&\eqref{const:K:0}
\end{flalign*}

\vspace{-1.6\baselineskip}
\begin{flalign*}
& C_0 = \left((h/\eta)E_3\mathbb{E}\left[ \left\Vert \overline{e}_{x}^{(0)}\right\Vert^2 \right] + E_4\mathbb{E}\left[ \left\Vert \widetilde{v}^{(0)}\right\Vert^2 \right] \right)\cdot \frac{2L^2}{1 - \eta\gamma_1\gamma_2}
&&&\eqref{defn:C:0}
\end{flalign*}

\vspace{-1.6\baselineskip}
\begin{flalign*}
& C_1 = \frac{2L^2\left(\eta \sigma^2 + 2d\right)}{N} \cdot \frac{w_2\gamma_1(h/\eta) + w_1}{1 - h\gamma_1\gamma_2},
\quad C_2 = \frac{2L^4\left(\eta + \frac{1 + \eta L}{\mu\left(1-\frac{\eta L}{2}\right)} \right)}{N^2\left(\delta^2 +\eta\mu\left(1 - \frac{\eta L}{2}\right) - 1 \right)}
&&&\eqref{defn:C:1},\eqref{defn:C:2}
\end{flalign*}

\vspace{-1.6\baselineskip}
\begin{flalign*}
& C_3 = \frac{2L^2}{N} \cdot \frac{\eta\sigma^2 + 2d}{\delta^2 +\eta\mu\left(1 - \frac{\eta L}{2}\right) - 1},
\quad
C_4 =  \frac{2L^2}{\delta^2 +\eta\mu\left(1 - \frac{\eta L}{2}\right) - 1}\mathbb{E}\left[ \left\Vert \overline{e}_{x}^{(0)}\right\Vert^2 \right]
&&&\eqref{defn:C:3},\eqref{defn:C:4}
\end{flalign*}

\vspace{-1.6\baselineskip}
\begin{flalign*}
&D_0 = \frac{1}{1-h\gamma_1\gamma_2}\left(E_1\mathbb{E}\left[ \left\Vert \widetilde{x}^{(0)}\right\Vert^2 \right] + E_2\mathbb{E}\left[ \left\Vert \overline{e}_{x}^{(0)}\right\Vert^2 \right]\right)
&&&\eqref{defn:D:0}
\end{flalign*}

\vspace{-1.6\baselineskip}
\begin{flalign*}
&D_1 = 2\frac{\sqrt{2(R_{h} + R'_{h})}}{1 - \overline{\gamma}_{{\scaleto{\widetilde{W}}{5pt}}}} + \frac{2\sigma}{\sqrt{1 - \overline{\gamma}_{{\scaleto{\widetilde{W}}{5pt}}}^2}}, 
\quad
D_2 = 2\sqrt{\frac{2d}{1 - \overline{\gamma}_{{\scaleto{\widetilde{W}}{5pt}}}^2}}
&&&\eqref{defn:D:0},\eqref{defn:D:1:2}
\end{flalign*}

\vspace{-1.6\baselineskip}
\begin{flalign*}
&\delta^2 \in \left[\left(1 - \frac{\eta\mu}{2}\left(1 - \frac{\eta L}{2} \right)\right) \vee \left(1 - h\frac{1-\overline{\gamma}_{{\scaleto{W}{3pt}}}}{4}\left(1 - \overline{\gamma}_{{\scaleto{I_{N}-W}{5pt}}}\right)\right) \,,\,  1\right)
&&&\eqref{delta:choice}
\end{flalign*}

\rule{\textwidth}{\heavyrulewidth}

\caption{Summary of the constants and where they are defined in the text. }\label{table_constants}
\end{table}


\section{Proof of the Main Results}\label{sec:proof:of:main}

In this section, we provide the proofs of Theorem~\ref{thm:main} and Proposition~\ref{prop:comparison}.

\subsection{Proof of Theorem~\ref{thm:main}}

In this section, we provide the proof of Theorem~\ref{thm:main} via establishing a sequence of key technical
results whose proofs will be provided in Appendix~\ref{sec:proofs:key}. In order to derive Theorem~\ref{thm:main}, based on the triangle inequality, we consider the following decomposition:
\begin{align}
\frac{1}{N}\sum_{i=1}^N\mathcal{W}_2\left(\mathcal{L}\left(x_i^{(k)}\right)\,,\, \pi \right) 
&\leq \frac{1}{N}\sum_{i=1}^N\mathcal{W}_2\left(\mathcal{L}\left(x_i^{(k)}\right)\,,\, \mathcal{L}\left(\overline{x}^{(k)}\right) \right) + \mathcal{W}_2\left(\mathcal{L}\left(\overline{x}^{(k)}\right)\,,\,\pi \right)\label{decompose:0},
\end{align}
where
\begin{align}
\mathcal{W}_2\left(\mathcal{L}\left(\overline{x}^{(k)}\right)\,,\,\pi \right)\leq \mathcal{W}_2\left(\mathcal{L}\left(\overline{x}^{(k)}\right),\mathcal{L}(x_{k}) \right)
+\mathcal{W}_{2}\left(\mathcal{L}(x_{k}),\pi\right),
\label{decompose:00}
\end{align}
$\overline{x}^{(k)}:=\frac{1}{N}\sum_{i=1}^{N}x_{i}{(k)}$
is the average iterates and $x_{k}$ has the iterates
\begin{equation}\label{overdamped:iterates:0}
x_{k+1} = x_{k} - \frac{\eta}{N}\nabla f\left(x_k\right) + \sqrt{2\eta}\overline{w}^{(k+1)}.
\end{equation}
These iterates correspond to the Euler-Maruyama discretization of overdamped Langevin diffusion 
\begin{equation}\label{overdamped:SDE}
dX_t = -\frac{1}{N}\nabla f(X_t)dt + \sqrt{2N^{-1}}dW_t,
\end{equation}
where $W_{t}$ is a standard $d$-dimensional Brownian motion, $\overline{w}^{(k)}:=\frac{1}{N}\sum_{i=1}^{N}w_{i}^{(k)}$, 
and $w_{i}^{(k)}$ are $\mathcal{N}(0,I_{d})$ distributed that are i.i.d. 
in both $k\in\mathbb{N}$ and $i=1,2,\ldots,N$.

The main idea of our proof technique is to bound the following three terms: (1) the $L^2$ distance between $x_{i}^{(k)}$ and their average $\bar{x}^{(k)}$; (2) the $L^2$ distance between the average iterate $\bar{x}^{(k)}$ and iterates $x_{k}$ in \eqref{overdamped:iterates:0} obtained from Euler-Maruyama discretization of overdamped SDE \eqref{overdamped:SDE}; and (3) the $\mathcal{W}_{2}$ distance between the law of $x_{k}$ in \eqref{overdamped:iterates:0} and the Gibbs distribution $\pi$. First, we upper bound the $L^2$ distance between $x_{i}^{(k)}$ and their average.

\subsubsection{Uniform $L^2$ bounds between $x_{i}^{(k)}$ and their average $\bar{x}^{(k)}$}

Denoting by $\mathbf{a}= \left\{a^{(0)}, a^{(1)}, \ldots, a^{(k)}, \ldots \right\}$ an infinite sequence of vectors, where $a^{(k)} \in \mathbb{R}^p$, $k=0,1, \ldots$ for some $p\in\mathbb{N}$. 
For a fixed $ \delta \in (0,1) $, we define the following quantity
\begin{equation}
\label{a:seq}
\|\mathbf{a}\|_2^{\delta, K} := \max_{k=0,1,\ldots,K}\mathbb{E}\left[\left\Vert \frac{1}{\delta^{k}}a^{(k)} \right\Vert^2\right].
\end{equation}
We first state a preliminary lemma that will be frequently used in the following analysis, 
and this lemma is a modification of Lemma~6 in~\cite{jakovetic2018unification}, 
and its proof will be provided in Appendix~\ref{sec:proofs:lemmas}.

\begin{lemma}
\label{lemma:inf:seq}
Consider two infinite random sequences $\mathbf{a}= \left\{a^{(0)}, a^{(1)}, \ldots, a^{(k)}, \ldots \right\}$ and \\
$\mathbf{b}= \left\{b^{(0)}, b^{(1)}, \ldots, b^{(k)}, \ldots \right\}$, with $a^{(k)}, b^{(k)} \in \mathbb{R}^p$ for some $p\in\mathbb{N}$ such that $\mathbb{E}\left\Vert a^{(k)}\right\Vert^2< \infty$
and $\mathbb{E}\left\Vert b^{(k)}\right\Vert^2< \infty$ for every $k\geq 0$.
Suppose that, for all $k=0,1, \ldots$, there holds:
\begin{equation}
\label{lemma:6:assumption}
\mathbb{E}\left\Vert a^{(k+1)}\right\Vert^2 \leq c_1\mathbb{E}\left\Vert a^{(k)}\right\Vert^2 +c_2\mathbb{E}\left\Vert b^{(k)}\right\Vert^2 + c_0.
\end{equation}
where $c_i \geq 0, i=0,1,2$. Then, for all $K=0,1, \ldots$, for any $\delta \in(0,1)$, we have:
\begin{equation}
\|\mathbf{a}\|_2^{\delta, K} \leq \frac{c_1}{\delta^2}\|\mathbf{a}\|_2^{\delta, K}+\frac{c_2}{\delta^2}\|\mathbf{b}\|_2^{\delta, K}+ \frac{c_0}{\delta^{2K}} + \mathbb{E}\left\Vert a^{(0)}\right\Vert^2.
\end{equation}
\end{lemma}

Next, we introduce the following technical lemma,
which is an extension of Lemma~\ref{lemma:inf:seq}
and this extension will be used in the proof of Lemma~\ref{lemma:tilde:x}. 

\begin{lemma}\label{lemma:seq}
Given any $n\in\mathbb{N}$ with $n\geq 2$, if
\begin{equation}
\mathbb{E}\left\Vert a^{(k+1)}\right\Vert^2 \leq c_1\mathbb{E}\left\Vert a^{(k)}\right\Vert^2 + \sum_{i=2}^nc_i\mathbb{E}\left\Vert b_i^{(k)}\right\Vert^2 + c_0,
\end{equation}
for every $k = 0,1,\ldots,K$, then
\begin{equation}
\|\mathbf{a}\|_2^{\delta, K} \leq \frac{c_1}{\delta^2}\|\mathbf{a}\|_2^{\delta, K}+\sum_{i=2}^n\frac{c_i}{\delta^2}\|\mathbf{b}_i\|_2^{\delta, K}+ \frac{c_0}{\delta^{2K}} + \mathbb{E}\left\Vert a^{(0)}\right\Vert^2.
\end{equation}
\end{lemma}

Next, we define the error vectors: 
\begin{equation}
\label{def:err}
e_x^{(k)}:=x^{(k)} - \mathbf{x}_*\,, \quad e_v^{(k)} := v^{(k)} + \nabla F\left(\mathbf{x}_*\right)\,, \quad e^{(k)}:= \left(\left(e_x^{(k)}\right)^T,\left(e_v^{(k)}\right)^T\right)^T,
\end{equation}
where $\mathbf{x}_* = \left[\left(x_*\right)^T, \ldots, \left(x_*\right)^T\right]^T \in \mathbb{R}^{Nd}$ is the vector of minimizer of the objective from
the target Gibbs distribution. For any $k=0,1,2,\ldots$, let us further define
\begin{align}
\label{def:avg:x:v}
&\overline{\mathbf{x}}^{(k)} := \left(\left(\overline{x}^{(k)}\right)^T,\left(\overline{x}^{(k)}\right)^T, 
\ldots,\left(\overline{x}^{(k)}\right)^T \right)^T \in \mathbb{R}^{Nd}, 
\\
&\overline{\mathbf{v}}^{(k)} := \left(\left(\overline{v}^{(k)}\right)^T,\left(\overline{v}^{(k)}\right)^T, 
\ldots,\left(\overline{v}^{(k)}\right)^T \right)^T \in \mathbb{R}^{Nd}, 
\end{align}
where $\overline{x}^{(k)} := \frac{1}{N}\sum_{i=1}^Nx_i^{(k)}$
and $\overline{v}^{(k)} := \frac{1}{N}\sum_{i=1}^Nv_i^{(k)}$. By introducing the following quantities:
\begin{equation}
\label{def:tildex:tildev}
\widetilde{x}^{(k)} := x^{(k)} - \overline{\mathbf{x}}^{(k)},\quad \widetilde{v}^{(k)} := v^{(k)} - \overline{\mathbf{v}}^{(k)},
\end{equation}
we define the average errors as follows.
\begin{equation}
\label{def:avi:err}
\overline{e}_x^{(k)} := \frac{1}{N}\sum_{i=1}^N\left(x_i^{(k)} - x_*\right), \quad \overline{e}_v^{(k)} := \frac{1}{N}\sum_{i=1}^N\left(v_i^{(k)} + \nabla f_i\left(x_*\right)\right).
\end{equation}
Now we can decompose the error terms $e_x^{(k)}$ and  $e_v^{(k)}$ in~\eqref{def:err} 
and get the following lemma. 

\begin{lemma}
\label{lemma:err}
For all $k=0,1,2,\ldots$, the error terms $e_x^{(k)}$ and  $e_v^{(k)}$ have the decomposition:
\begin{equation}
\label{def:ex}
e_x^{(k)} = \widetilde{x}^{(k)} + 1_N \otimes \overline{e}_x^{(k)}\,, \quad e_v^{(k)} = \widetilde{v}^{(k)} + 1_N \otimes \overline{e}_v^{(k)}\,,
\end{equation}
with 
\begin{equation}
\label{tilde:v}
\widetilde{v}^{(k)} = e_v^{(k)}, \quad \overline{e}_v^{(k)} = \overline{v}^{(k)} = 0.
\end{equation}
\end{lemma}

Next, to facilitate the presentations, let us define two sequences $\widetilde{\mathbf{x}} := \left\{ \widetilde{x}^{(0)}, \widetilde{x}^{(1)}, \ldots, \widetilde{x}^{(k)},\ldots\right\}$ and $\widetilde{\mathbf{v}} := \left\{ \widetilde{v}^{(0)}, \widetilde{v}^{(1)}, \ldots, \widetilde{v}^{(k)},\ldots\right\}$ where $\widetilde{x}^{(k)}, \widetilde{v}^{(k)} \in \mathbb{R}^{Nd}$ are given in \eqref{def:tildex:tildev}. 
By following the notation in~\eqref{a:seq}, we denote
\begin{equation}
\label{def:tilde:x:v}
\left\Vert \widetilde{\mathbf{x}} \right\Vert_2^{\delta, K} := \max_{k=0,1,\ldots,K}\mathbb{E}\left[\left\Vert \frac{1}{\delta^{k}}\widetilde{x}^{(k)} \right\Vert^2\right], \quad \left\Vert \widetilde{\mathbf{v}} \right\Vert_2^{\delta, K} = \max_{k=0,1,\ldots,K}\mathbb{E}\left[\left\Vert \frac{1}{\delta^{k}}\widetilde{v}^{(k)} \right\Vert^2\right].
\end{equation}
Similarly, we also define the sequence $\overline{\mathbf{e}}_x = \left\{ \overline{e}_x^{(0)}, \overline{e}_x^{(1)}, \ldots, \overline{e}_x^{(k)},\ldots\right\}$ where $\overline{e}^{(k)}_x \in \mathbb{R}^d$ is defined in~\eqref{def:avi:err} and moreover, we denote
\begin{equation}
\left\Vert \overline{\mathbf{e}}_x \right\Vert_2^{\delta, K} := \max_{k=0,1,\ldots,K}\mathbb{E}\left[\left\Vert \frac{1}{\delta^{k}}\overline{e}_x^{(k)} \right\Vert^2\right].
\end{equation}

Now we present a sequence of technical lemmas.
First, we provide an upper bound on $\left\Vert \overline{\mathbf{e}}_x \right\Vert_2^{\delta, K}$ by 
using $\left\Vert \widetilde{\mathbf{x}}\right\Vert_2^{\delta, K}$.

\begin{lemma}
\label{lemma:avi:ex}
Suppose Assumptions~\ref{assumption:f}, \ref{assumption:noise}, and~\ref{assumption:mixing} hold. 
Taking the stepsize $0<\eta < \frac{2}{L}  \wedge 1$, 
then for any $\delta^2 \in \left(1 - \eta\mu\left(1 - \frac{\eta L}{2} \right) \,,\, 1\right)$, the following inequality holds for every $K\geq 0$:
\begin{align}
\left\Vert \overline{\mathbf{e}}_x \right\Vert_2^{\delta, K}  
& \leq \eta\cdot\frac{L^2}{N^2\left(\delta^2 +\eta\mu\left(1 - \frac{\eta L}{2}\right) - 1 \right)}\left(\eta + \frac{1 + \eta L}{\mu\left(1-\frac{\eta L}{2}\right)} \right)\left\Vert \widetilde{\mathbf{x}}\right\Vert_2^{\delta, K} 
\nonumber 
\\
& \qquad\qquad + \frac{\eta}{N\delta^{2K-2}} \cdot \frac{\eta\sigma^2 + 2d}{\delta^2 +\eta\mu\left(1 - \frac{\eta L}{2}\right) - 1}
+ \frac{\delta^2}{\delta^2 +\eta\mu\left(1 - \frac{\eta L}{2}\right) - 1}\mathbb{E}\left[ \left\Vert \overline{e}_{x}^{(0)}\right\Vert^2 \right].
\end{align}
\end{lemma}

Next, we provide an upper bound on $\left\Vert \widetilde{\mathbf{x}}\right\Vert_2^{\delta, K}$
in terms of $\left\Vert \overline{\mathbf{e}}_x \right\Vert_2^{\delta, K}$ and  $\left\Vert \widetilde{\mathbf{v}}\right\Vert_2^{\delta, K}$.

\begin{lemma}
\label{lemma:tilde:x}
Under the assumptions in Lemma~\ref{lemma:avi:ex}, in addition, let the stepsize $
\eta \leq \frac{1}{L}$. Denoting the eigenvalues of matrix $\widetilde W$ such that $1 = \lambda_1^{{\scaleto{\widetilde{W}}{5pt}}} > \lambda_2^{{\scaleto{\widetilde{W}}{5pt}}} \geq \cdots \geq \lambda_N^{{\scaleto{\widetilde{W}}{5pt}}} > 0$. Define the positive constant
\begin{equation}
\label{const:gamma:tilde:W}
\gamma_{{\scaleto{\widetilde{W}}{5pt}}}:=\begin{cases}
\left\vert \lambda_2^{{\scaleto{\widetilde{W}}{5pt}}} \right\vert^2 &\text{if $0 < \left\vert \lambda_2^{{\scaleto{\widetilde{W}}{5pt}}} \right\vert^2 < \frac{1}{2}$},
\\
\frac{\left\vert \lambda_2^{{\scaleto{\widetilde{W}}{5pt}}} \right\vert^2\left(\left\vert \lambda_2^{{\scaleto{\widetilde{W}}{5pt}}} \right\vert^2 - \frac{1}{2}\right)}{1 - \left\vert \lambda_2^{{\scaleto{\widetilde{W}}{5pt}}} \right\vert^2} &\text{if $\frac{1}{2} \leq \left\vert \lambda_2^{{\scaleto{\widetilde{W}}{5pt}}} \right\vert^2 \leq \frac{2}{3}$},
\\
\frac{5\left\vert \lambda_2^{{\scaleto{\widetilde{W}}{5pt}}} \right\vert^2 - 3\left\vert \lambda_2^{{\scaleto{\widetilde{W}}{5pt}}} \right\vert^4-2}{3\left\vert \lambda_2^{{\scaleto{\widetilde{W}}{5pt}}} \right\vert^2-1} &\text{if $ \frac{2}{3} \leq \left\vert \lambda_2^{{\scaleto{\widetilde{W}}{5pt}}} \right\vert^2 < 1$}.
\end{cases}
\end{equation}
By taking
\begin{equation}
0<\eta \leq \frac{\gamma_{{\scaleto{\widetilde{W}}{5pt}}}}{6(L + \mu)},
\end{equation}
it holds that:
\begin{align}
\left\Vert \widetilde{\mathbf{x}} \right\Vert_2^{\delta, K} 
& \leq \frac{\eta\mu}{\gamma_{{\scaleto{\widetilde{W}}{5pt}}}} \left(L/\mu + 3\eta L - 1\right)\left\Vert \overline{\mathbf{e}}_x \right\Vert_2^{\delta, K} 
+ \frac{\eta}{\gamma_{{\scaleto{\widetilde{W}}{5pt}}}}\left(\frac{1}{2L} + \eta + \frac{\eta}{2L\mu}\right)\left\Vert \widetilde{\mathbf{v}} \right\Vert_2^{\delta, K} 
\nonumber 
\\
 & \qquad\qquad + \frac{\eta}{\gamma_{{\scaleto{\widetilde{W}}{5pt}}}\delta^{2K-2}}\left(N + \frac{1}{N} \right)(\eta\sigma^2 + 2d) + \frac{\delta^2}{\gamma_{{\scaleto{\widetilde{W}}{5pt}}}}\mathbb{E}\left[ \left\Vert \widetilde{x}^{(0)}\right\Vert^2 \right],\label{the:result:holds}
\end{align}
where the constant $\delta^2$ depends on $\left\vert \lambda_2^{{\scaleto{\widetilde{W}}{5pt}}} \right\vert^2$ in three regimes:

(1). If $\left\vert \lambda_2^{{\scaleto{\widetilde{W}}{5pt}}} \right\vert^2 < \frac{1}{2}$, then \eqref{the:result:holds} holds for all $\delta$ such that $1 > \delta^2 \geq 2\left\vert \lambda_2^{{\scaleto{\widetilde{W}}{5pt}}} \right\vert^2$; 

(2). If $\frac{2}{3} \geq \left\vert \lambda_2^{{\scaleto{\widetilde{W}}{5pt}}} \right\vert^2 \geq \frac{1}{2}$, then \eqref{the:result:holds} holds for all $\delta$ such that $1 > \delta^2 \geq \frac{\left\vert \lambda_2^{{\scaleto{\widetilde{W}}{5pt}}} \right\vert^2}{2\left(1 - \left\vert \lambda_2^{{\scaleto{\widetilde{W}}{5pt}}} \right\vert^2\right)} \geq \frac{1}{2}$;

(3). If $1 > \left\vert \lambda_2^{{\scaleto{\widetilde{W}}{5pt}}} \right\vert^2 > \frac{2}{3}$, then \eqref{the:result:holds} holds for all $\delta$ such that $1 > \delta^2 \geq \frac{4\left\vert \lambda_2^{{\scaleto{\widetilde{W}}{5pt}}} \right\vert^2-2}{3\left\vert \lambda_2^{{\scaleto{\widetilde{W}}{5pt}}} \right\vert^2-1} > \frac{1}{2}$.
\end{lemma}

As a direct result from Lemmas~\ref{lemma:avi:ex} and~\ref{lemma:tilde:x}, we obtain the following lemma that provides
an upper bound on $\left\Vert \widetilde{\mathbf{x}} \right\Vert_2^{\delta, K}$ in terms of $\left\Vert \widetilde{\mathbf{v}} \right\Vert_2^{\delta, K}$.

\begin{lemma}
\label{lemma:tilde:xv}
Denote
\begin{equation}
\label{const:A}
A:= \left(\frac{L}{\mu} - 1 + \frac{\gamma_{{\scaleto{\widetilde{W}}{5pt}}}}{2(1+\mu/L)}\right)\cdot \frac{4L^2}{N^2}\left(1 + \frac{2 + 2L}{\mu} \right).
 \end{equation}
Given any $\delta^2 \in \left[1 - \frac{\eta\mu}{2}\left(1 - \frac{\eta L}{2} \right) \,,\, 1\right)$ under the conditions for $\delta$ in Lemma~\ref{lemma:tilde:x}, and suppose $\eta \leq \frac{\gamma_{{\scaleto{\widetilde{W}}{5pt}}}}{6(L+\mu)} \wedge \frac{\gamma_{{\scaleto{\widetilde{W}}{5pt}}}}{2A}$, 
there holds,
\begin{align}
\left\Vert \widetilde{\mathbf{x}} \right\Vert_2^{\delta, K} 
& \leq \eta\cdot \frac{2}{\gamma_{{\scaleto{\widetilde{W}}{5pt}}}}\left(\frac{1}{2L} + \eta + \frac{\eta}{2L\mu}\right)\left\Vert \widetilde{\mathbf{v}} \right\Vert_2^{\delta, K}
\nonumber 
\\
& \qquad + \eta \cdot \frac{2}{\delta^{2K-2}}(\eta\sigma^2 + 2d)\left(\frac{N + \frac{1}{N}}{\gamma_{{\scaleto{\widetilde{W}}{5pt}}}} + \frac{4}{N\gamma_{{\scaleto{\widetilde{W}}{5pt}}}}\cdot \left(L/\mu + 3\eta L - 1\right) \right) 
\nonumber 
\\
& \qquad\qquad + \frac{8\delta^2}{\gamma_{{\scaleto{\widetilde{W}}{5pt}}}} \left(L/\mu + 3\eta L - 1\right)\mathbb{E}\left[ \left\Vert \overline{e}_{x}^{(0)}\right\Vert^2 \right]  + \frac{2\delta^2}{\gamma_{{\scaleto{\widetilde{W}}{5pt}}}}\mathbb{E}\left[ \left\Vert \widetilde{x}^{(0)}\right\Vert^2 \right],
\end{align}
where $\gamma_{{\scaleto{\widetilde{W}}{5pt}}}$ defined in~\eqref{const:gamma:tilde:W} depending on three regimes in Lemma~\ref{lemma:tilde:x}.
\end{lemma}

Next, we define the quantities
\begin{equation}
\label{defn:bar:gamma}
\overline{\gamma}_{{\scaleto{I_{N}-W}{5pt}}}:=\max\left\{1 - \left\vert \lambda_2^W \right\vert, 1-\left\vert \lambda_N^W \right\vert \right\}, 
\quad 
\overline{\gamma}_{{\scaleto{W}{3pt}}}:=\max\left\{\left\vert \lambda_2^W \right\vert, \left\vert \lambda_N^W \right\vert \right\},
\end{equation}
so that $1 > \overline{\gamma}_{{\scaleto{I_{N}-W}{5pt}}} \geq 1 - \overline{\gamma}_{{\scaleto{W}{3pt}}} > 0$. 
In the following lemma, we derive an upper bound on $\left\Vert \widetilde{\mathbf{v}}\right\Vert_2^{\delta, K}$
in terms of $\left\Vert \widetilde{\mathbf{x}} \right\Vert_2^{\delta, K}$.

\begin{lemma}
\label{lemma:tilde:vx}
By taking 
\begin{equation}
\label{choose:h}
0 < h \leq \frac{1-\overline{\gamma}_{{\scaleto{W}{3pt}}}}{4\overline{\gamma}_{{\scaleto{I_{N}-W}{5pt}}}^2}\wedge\frac{1}{2},
\end{equation} 
and $\delta^2 \geq 1 - h\frac{1-\overline{\gamma}_{{\scaleto{W}{3pt}}}}{4}\left(1 - \overline{\gamma}_{{\scaleto{I_{N}-W}{5pt}}}\right) > 0$ in three regimes defined in Lemma~\ref{lemma:tilde:x}, the following bound holds:
\begin{align}
\left\Vert \widetilde{\mathbf{v}}\right\Vert_2^{\delta, K} 
& \leq \frac{12(h/\eta)\left(L^2 + L\left\Vert B \right\Vert^2 \right)}{(1-\overline{\gamma}_{{\scaleto{W}{3pt}}})\left(1 - \overline{\gamma}_{{\scaleto{I_{N}-W}{5pt}}}^2\right)}\left(1 + \frac{4L^2\left(1 + \frac{2 + 2L}{\mu} \right)}{N^2\mu}\right)\left\Vert \widetilde{\mathbf{x}}\right\Vert_2^{\delta, K} 
\nonumber 
\\
& \qquad
 + \left(\frac{6\left(L^2 + L\left\Vert B \right\Vert^2 \right)}{N\mu} + N\right)\cdot \frac{8(h/\eta)}{(1-\overline{\gamma}_{{\scaleto{W}{3pt}}})\left(1 - \overline{\gamma}_{{\scaleto{I_{N}-W}{5pt}}}^2\right)} \cdot \frac{\eta\sigma^2 + 2d}{\delta^{2K-2}} 
 \nonumber
 \\
 & \qquad\qquad + \frac{12\delta^2(h/\eta)\left(L^2 + L\left\Vert B \right\Vert^2 \right)}{\eta\mu(1-\overline{\gamma}_{{\scaleto{W}{3pt}}})\left(1 - \overline{\gamma}_{{\scaleto{I_{N}-W}{5pt}}}^2\right)} \mathbb{E}\left[ \left\Vert \overline{e}_{x}^{(0)}\right\Vert^2 \right] + \frac{4\delta^2}{h(1-\overline{\gamma}_{{\scaleto{W}{3pt}}})\left(1 - \overline{\gamma}_{{\scaleto{I_{N}-W}{5pt}}}^2\right)}\left\Vert \widetilde{v}^{(0)} \right\Vert^{2}.
\end{align}
\end{lemma}


Now one can immediately derive from Lemma~\ref{lemma:tilde:xv} and Lemma~\ref{lemma:tilde:vx}
that
\begin{align}
& \left\Vert \widetilde{\mathbf{x}}\right\Vert_2^{\delta,K} 
\leq \eta\gamma_1\left\Vert \widetilde{\mathbf{v}}\right\Vert_2^{\delta,K} + \eta\frac{w_1(\eta\sigma^2 + 2d)}{N\delta^{2K-2}} + \delta^2 E_1\mathbb{E}\left[ \left\Vert \overline{e}_{x}^{(0)}\right\Vert^2 \right] + \delta^2E_2\mathbb{E}\left[ \left\Vert \widetilde{x}^{(0)}\right\Vert^2 \right],
\\
& \left\Vert \widetilde{\mathbf{v}}\right\Vert_2^{\delta,K} \leq (h/\eta)\gamma_2\left\Vert \widetilde{\mathbf{x}}\right\Vert_2^{\delta,K} + (h/\eta)\frac{w_2(\eta\sigma^2 + 2d)}{N\delta^{2K-2}} 
\nonumber
\\
&\qquad\qquad\qquad+ 
\delta^2(h/\eta)(E_3/\eta)\mathbb{E}\left[ \left\Vert \overline{e}_{x}^{(0)}\right\Vert^2 \right] + \delta^2(E_4/h)\mathbb{E}\left[ \left\Vert \widetilde{v}^{(0)}\right\Vert^2 \right]
,
\label{const:twist}
\end{align}
where the constants are defined as:
\begin{align}
\gamma_1 := \frac{1}{\gamma_{{\scaleto{\widetilde{W}}{5pt}}}}\left(\frac{1}{L} + 2 + \frac{1}{L\mu}\right), 
\qquad
\gamma_2 := \frac{12\left(L^2 + L\left\Vert B \right\Vert^2 \right)}{(1-\overline{\gamma}_{{\scaleto{W}{3pt}}})\left(1 - \overline{\gamma}_{{\scaleto{I_{N}-W}{5pt}}}^2\right)}\left(1 + \frac{4L^2\left(1 + \frac{2 + 2L}{\mu} \right)}{N^2\mu}\right), 
\label{defn:gamma:1:2}
\end{align}
and
\begin{align}
& w_1 := 2\left(\frac{N^2 + 1}{\gamma_{{\scaleto{\widetilde{W}}{5pt}}}} + \frac{4}{\gamma_{{\scaleto{\widetilde{W}}{5pt}}}}\cdot \left(L/\mu + 3\eta L - 1\right) \right) 
\label{defn:w:1}
,
\\
& w_2 :=\left(\frac{6\left(L^2 + L\left\Vert B \right\Vert^2 \right)}{N\mu} + N\right)\cdot \frac
{8}{(1-\overline{\gamma}_{{\scaleto{W}{3pt}}})\left(1 - \overline{\gamma}_{{\scaleto{I_{N}-W}{5pt}}}^2\right)},
\label{defn:w:2}
\\
& E_1 := \frac{8}{\gamma_{{\scaleto{\widetilde{W}}{5pt}}}}\left(L/\mu + 3\eta L - 1\right), \quad E_2 := \frac{2}{\gamma_{{\scaleto{\widetilde{W}}{5pt}}}},
\label{defn:E:1:2}
\\
& E_3 := \frac{12\left(L^2 + L\left\Vert B \right\Vert^2 \right)}{\mu(1-\overline{\gamma}_{{\scaleto{W}{3pt}}})\left(1 - \overline{\gamma}_{{\scaleto{I_{N}-W}{5pt}}}^2\right)} , \quad E_4 := \frac{4}{(1-\overline{\gamma}_{{\scaleto{W}{3pt}}})\left(1 - \overline{\gamma}_{{\scaleto{I_{N}-W}{5pt}}}^2\right)},
\label{defn:E:3:4}
\end{align}
where $h \leq \frac{1-\overline{\gamma}_{{\scaleto{W}{3pt}}}}{4\overline{\gamma}_{{\scaleto{I_{N}-W}{5pt}}}^2}\wedge\frac{1}{2}$ from Lemma~\ref{lemma:tilde:vx}. 

We note that if $h = 0$, then $U = \widetilde{W} - W = 0$, and by~\eqref{upt2} and~\eqref{def:tildex:tildev}, we have $\widetilde{v}^{(0)} = v^{(0)} = \mathbf{0}$, and we observe from~\eqref{tilde:v:t} in the proof, $\left\Vert \widetilde{v}^{(k+1)}\right\Vert^2 = \left\Vert \widetilde{v}^{(k)}\right\Vert^2 = \cdots = \left\Vert \widetilde{v}^{(0)}\right\Vert^2 = 0$; hence, we have $\left\Vert \widetilde{\mathbf{v}}\right\Vert_2^{\delta,K} = 0$. In the case $h = 0$, we can get from~\eqref{const:twist} that
\begin{equation}
\left\Vert \widetilde{\mathbf{x}}\right\Vert_2^{\delta,K} 
\leq \eta\frac{w_1(\eta\sigma^2 + 2d)}{N\delta^{2K-2}} + \delta^2 E_1\mathbb{E}\left[ \left\Vert \overline{e}_{x}^{(0)}\right\Vert^2 \right] + \delta^2E_2\mathbb{E}\left[ \left\Vert \widetilde{x}^{(0)}\right\Vert^2 \right],
\end{equation}
and moreover, EXTRA SGLD algorithm reduces to DE-SGLD when $h=0$, and our result implies:
\begin{align}
\mathbb{E}\left[\left\Vert x^{(K)} - \overline{x}^{(K)} \right\Vert^2\right] 
&\leq \delta^{2K}\left\Vert \widetilde{\mathbf{x}}\right\Vert_2^{\delta,K} 
\nonumber 
\\
&\leq \eta^2\delta^2\frac{w_1\sigma^2}{N} + \eta\delta^2\frac{2dw_1}{N} + \delta^{2K + 2} E_1\mathbb{E}\left[ \left\Vert \overline{e}_{x}^{(0)}\right\Vert^2 \right] + \delta^{2K + 2}E_2\mathbb{E}\left[ \left\Vert \widetilde{x}^{(0)}\right\Vert^2 \right].
\label{bound:sgld}
\end{align}
This bound is of the same order as the one shown by Lemma~6 from~\cite{gurbuzbalaban2021decentralized}. Suppose $h \neq 0 $, we present upper bounds on $\Vert\widetilde{\mathbf{x}}\Vert_{2}^{\delta,K}$ and $\Vert\widetilde{\mathbf{v}}\Vert_{2}^{\delta,K}$ for EXTRA SGLD algorithm as follows.

\begin{theorem}
\label{theorem:norm:tilde}
Assume that the stepsize $\eta$ satisfies
\begin{equation}
0 < \eta < \frac{1}{h\gamma_1\gamma_2} \wedge \frac{\gamma_{{\scaleto{\widetilde{W}}{5pt}}}}{6(L+\mu) \vee 2A}\wedge 1,
\end{equation} 
where 
\begin{equation}
0 < h \leq \frac{1-\overline{\gamma}_{{\scaleto{W}{3pt}}}}{4\overline{\gamma}_{{\scaleto{I_{N}-W}{5pt}}}^2}\wedge\frac{1}{2} \wedge \frac{1}{\gamma_1\gamma_2},
\end{equation}
so that 
the condition in~\eqref{choose:h} is satisfied. Moreover, the constant $A$ is defined in~\eqref{const:A} and $\gamma_{{\scaleto{\widetilde{W}}{5pt}}}$ defined by $\left\vert \lambda_2^{{\scaleto{\widetilde{W}}{5pt}}} \right\vert^2$ in~\eqref{const:gamma:tilde:W}. 
%
%
Under the conditions in Lemma~\ref{lemma:tilde:vx}, 
for any constant $\delta$ in three regimes depending on $\left\vert \lambda_2^{{\scaleto{\widetilde{W}}{5pt}}} \right\vert^2$ from Lemma~\ref{lemma:tilde:x}, and furthermore, it satisfies:
\begin{equation}\label{delta:choice}
\delta^2 \in \left[\left(1 - \frac{\eta\mu}{2}\left(1 - \frac{\eta L}{2} \right)\right) \vee \left(1 - h\frac{1-\overline{\gamma}_{{\scaleto{W}{3pt}}}}{4}\left(1 - \overline{\gamma}_{{\scaleto{I_{N}-W}{5pt}}}\right)\right) \,,\,  1\right),
\end{equation} 
then it holds that:
\begin{align}
\label{uniform:bound:x}
\left\Vert \widetilde{\mathbf{x}} \right\Vert_2^{\delta,K} & \leq 
\frac{\eta^2}{\delta^{2K-2}} \cdot \frac{\left(w_2\gamma_1(h/\eta) + w_1\right)\sigma^2/N}{1 - h\gamma_1\gamma_2} + \frac{\eta}{\delta^{2K-2}} \cdot  \Bigg[\frac{2d\left(w_2\gamma_1(h/\eta) + w_1\right)/N}{1 - h\gamma_1\gamma_2} 
\nonumber 
\\
& \qquad+ \frac{\gamma_1}{1-h\gamma_1\gamma_2}\delta^{2K}\left((h/\eta)(E_3/\eta)\mathbb{E}\left[ \left\Vert \overline{e}_{x}^{(0)}\right\Vert^2 \right] + (E_4/h)\mathbb{E}\left[ \left\Vert \widetilde{v}^{(0)}\right\Vert^2 \right] \right)\Bigg] + \delta^2D_0,
\end{align}
and
\begin{align}
\label{uniform:bound:v}
\left\Vert \widetilde{\mathbf{v}} \right\Vert_2^{\delta,K} & \leq  \frac{h\eta}{\delta^{2K-2}} \cdot \frac{\gamma_2\left(w_2\gamma_1(h/\eta) + w_1\right)\sigma^2/N}{1 - h\gamma_1\gamma_2} + \frac{h}{\delta^{2K-2}}\cdot \Bigg[\frac{2\gamma_2d\left(w_2\gamma_1(h/\eta) + w_1\right)/N}{1 - h\gamma_1\gamma_2} + \frac{w_2\sigma^2}{N}
\nonumber 
\\
& \qquad+ \frac{\gamma_1\gamma_2}{1-h\gamma_1\gamma_2}\delta^{2K}\left((h/\eta)(E_3/\eta)\mathbb{E}\left[ \left\Vert \overline{e}_{x}^{(0)}\right\Vert^2 \right] + (E_4/h)\mathbb{E}\left[ \left\Vert \widetilde{v}^{(0)}\right\Vert^2 \right] \right)\Bigg] + (h/\eta)\delta^2\gamma_2D_0
\nonumber 
\\
& \qquad\qquad + (h/\eta)\frac{2dw_2}{N\delta^{2K-2}} + \delta^2(h/\eta)(E_3/\eta)\mathbb{E}\left[ \left\Vert \overline{e}_{x}^{(0)}\right\Vert^2 \right] +\delta^2(E_4/h)\mathbb{E}\left[ \left\Vert \widetilde{v}^{(0)}\right\Vert^2 \right],
\end{align}
where 
\begin{equation}\label{defn:D:0}
D_0 := \frac{1}{1-h\gamma_1\gamma_2}\left(E_1\mathbb{E}\left[ \left\Vert \widetilde{x}^{(0)}\right\Vert^2 \right] + E_2\mathbb{E}\left[ \left\Vert \overline{e}_{x}^{(0)}\right\Vert^2 \right]\right),
\end{equation}
and $\gamma_{1},\gamma_{2},w_{1},w_{2},E_{1},E_{2},E_{3},E_{4}$ are defined in \eqref{defn:gamma:1:2}, \eqref{defn:w:1}, \eqref{defn:w:2}, \eqref{defn:E:1:2} and  \eqref{defn:E:3:4}.
\end{theorem}

Next, we provide the uniform bounds on $\mathbb{E}\left[\left\Vert \widetilde{v}^{(k)} \right\Vert^2\right]$ and $ \mathbb{E}\left[\left\Vert \nabla F\left(x^{(k)}\right) \right\Vert^2\right]$. We first use the upper bounds on $\Vert\widetilde{\mathbf{x}}\Vert_{2}^{\delta,K}$ and $\Vert\widetilde{\mathbf{v}}\Vert_{2}^{\delta,K}$ from Theorem~\ref{theorem:norm:tilde} to get the next lemma.

\begin{lemma}
\label{lemma:bound:grad:tilde:v}
Under the assumptions for Theorem~\ref{theorem:norm:tilde}, the following bounds hold for $\mathbb{E}\left[\left\Vert \widetilde{v}^{(k)} \right\Vert^2\right]$ and $ \mathbb{E}\left[\left\Vert \nabla F\left(x^{(k)}\right) \right\Vert^2\right]$ uniformly for every $k = 1,2,3,\ldots$:
\begin{align}
& \mathbb{E}\left[\left\Vert \widetilde{v}^{(k)}\right\Vert^{2}\right]
\leq h\delta^2 \cdot \frac{C_1\gamma_2}{2L^2} + \delta^{2k+2}h \cdot \frac{C_0\gamma_1\gamma_2}{2L^2}
\nonumber 
\\
& \qquad\qquad\qquad\qquad\qquad + \delta^{2k+2} \cdot \left((h/\eta)\gamma_2D_0 + C_0\right) + (h/\eta)\delta^2 \cdot \frac{w_2}{N}\left(\eta\sigma^2 + 2d\right),
\\
& \mathbb{E}\left[\left\Vert \nabla F\left(x^{(k)}\right) \right\Vert^2 \right] 
\leq \eta\delta^2\left(C_1 + C_3\right) + \delta^2\eta^2\left(\frac{C_1C_2}{2L^2} \right) + \delta^{2K_0}\eta\left(\gamma_1C_0 + D_0C_2\right) 
\nonumber 
\\
& \qquad\qquad \qquad\qquad \qquad  + \delta^{2K_0}\eta^2\left(\frac{\gamma_1C_0C_2}{2L^2}\right) +\delta^{2K_0}\left(D_0 + C_4 \right)+ 2\left\Vert \nabla F(\mathbf{x}_*) \right\Vert^2,
\end{align}
where the constants are given by:
\begin{align}
& C_0 := \frac{2L^2}{1-h\gamma_1\gamma_2}\left((h/\eta)(E_3/\eta)\mathbb{E}\left[ \left\Vert \overline{e}_{x}^{(0)}\right\Vert^2 \right] + (E_4/h)\mathbb{E}\left[ \left\Vert \widetilde{v}^{(0)}\right\Vert^2 \right]\right) ,\label{defn:C:0}
\\
& C_1 := \frac{2L^2\left(\eta \sigma^2 + 2d\right)}{N} \cdot \frac{w_2\gamma_1(h/\eta) + w_1}{1 - h\gamma_1\gamma_2},
\label{defn:C:1}
\\
& C_2 := \frac{2L^4}{N^2\left(\delta^2 +\eta\mu\left(1 - \frac{\eta L}{2}\right) - 1 \right)}\left(\eta + \frac{1 + \eta L}{\mu\left(1-\frac{\eta L}{2}\right)} \right),\label{defn:C:2}
\\
& C_3:= \frac{2L^2}{N} \cdot \frac{\eta\sigma^2 + 2d}{\delta^2 +\eta\mu\left(1 - \frac{\eta L}{2}\right) - 1},\label{defn:C:3}
\\
& C_4:= \frac{2L^2}{\delta^2 +\eta\mu\left(1 - \frac{\eta L}{2}\right) - 1}\mathbb{E}\left[ \left\Vert \overline{e}_{x}^{(0)}\right\Vert^2 \right],\label{defn:C:4}
\end{align}
and $D_0$ is defined in~\eqref{defn:D:0}, $E_3, E_4$ are defined in~\eqref{defn:E:3:4} and $w_1, w_2$ are defined in~\eqref{defn:w:1}-\eqref{defn:w:2}.
\end{lemma}

As an immediate corollary of Lemma~\ref{lemma:bound:grad:tilde:v}, 
we obtain the following upper bounds for $\mathbb{E}\left[\left\Vert \widetilde{v}^{(k)}\right\Vert^{2}\right]$ and $\mathbb{E}\left[\left\Vert \nabla F\left(x^{(k)}\right) \right\Vert^2 \right]$ that are uniform in $k$ when $k$ is larger than a specific lower bound.

\begin{corollary}\label{cor:bound:grad:tilde:v}
Under the assumptions for Theorem~\ref{theorem:norm:tilde},
for any $k\geq K_{0}$, we have
\begin{equation}
\label{uniform:bounds}
\mathbb{E}\left[\left\Vert \widetilde{v}^{(k)} \right\Vert^2\right] \leq R_h,\qquad \mathbb{E}\left[\left\Vert \nabla F\left(x^{(k)}\right) \right\Vert^2\right] \leq R_h', 
\end{equation}
\begin{align}
& R_h:=  h\delta^2\left(\frac{C_1\gamma_2}{2L^2} +  \frac{C_0\gamma_1\gamma_2}{2L^2} \right) + (h/\eta)\delta^{2}\left( \gamma_2D_0 + \frac{w_2}{N}\left(\eta\sigma^2 + 2d\right) \right) + \left\Vert \nabla F(\mathbf{x}_*) \right\Vert^2,
\label{R:h}
\\
& R_h':= \eta\delta^2\left(C_1 + C_3 + \gamma_1C_0 + D_0C_2\right) + \delta^2\eta^2\left(\frac{C_1C_2}{2L^2} + \frac{\gamma_1C_0C_2}{2L^2}\right) 
 + 3\left\Vert \nabla F(\mathbf{x}_*)\right\Vert^2,
\label{R:h:prime}
\end{align}
and 
\begin{equation}\label{const:K:0}
K_0:=\frac{\delta^2}{1 - \delta^2}\left[\left(1 - \frac{\left\Vert \nabla F(\mathbf{x}_*) \right\Vert^2}{D_0 + C_4}\right)  \vee \left(1 - \frac{\left\Vert \nabla F(\mathbf{x}_*) \right\Vert^2}{C_0}\right)  \right] \vee 0.
\end{equation}
\end{corollary}

Now we are ready to present our main technical result on the error between the iterate $x^{(k)}$ and their average (taken over $N$ agents) $\overline{\mathbf{x}}^{(k)}$ after $k$ iterations:  
\begin{equation}
\overline{x}^{(k)} = \frac{1}{N}\sum_{i = 0}^N x_i^{(k)} \in \mathbb R^d, \qquad \overline{\mathbf{x}}^{(k)} = \left[\left(\overline{x}^{(k)}\right)^T,\left(\overline{x}^{(k)}\right)^T,\ldots,\left(\overline{x}^{(k)}\right)^T\right]^T \in \mathbb R^{Nd}.
\end{equation}
We have the next corollary.

\begin{corollary}
\label{corollary:x2avg}
With the assumptions for Theorem~\ref{theorem:norm:tilde}, it holds for $k\geq K_{0}$, with $K_{0}$ given in \eqref{const:K:0},
\begin{equation}
\sum_{i=1}^N\mathbb{E}\left[\left\Vert x_i^{(k)}- \overline{\mathbf{x}}^{(k)}\right\Vert^2\right] 
\leq 4\left(\overline{\gamma}_{{\scaleto{\widetilde{W}}{5pt}}}\right)^{2k} \mathbb{E}\left[ \left\Vert x^{(0)}\right\Vert^2 \right] +  8\eta^2 \cdot \frac{R_{h} + R'_{h}}{\left(1 - \overline{\gamma}_{{\scaleto{\widetilde{W}}{5pt}}}\right)^2} + \frac{4\eta^2\sigma^2N}{1 - \overline{\gamma}_{{\scaleto{\widetilde{W}}{5pt}}}^2} + \frac{8 \eta dN}{1 - \overline{\gamma}_{{\scaleto{\widetilde{W}}{5pt}}}^2},
\end{equation}
where $R_{h},R'_{h}$ are defined in \eqref{R:h}-\eqref{R:h:prime} and $\overline{\gamma}_{{\scaleto{\widetilde{W}}{5pt}}} := \max\left\{\left\vert \lambda_2^{{\scaleto{\widetilde{W}}{5pt}}} \right\vert\,,\, \left\vert \lambda_N^{{\scaleto{\widetilde{W}}{5pt}}} \right\vert \right\} \in [0,1)$.
\end{corollary}

In order to sample from the Gibbs distribution, we recall the decomposition \eqref{decompose:0}:
\begin{equation}
\label{decompose}
\frac{1}{N}\sum_{i=1}^N\mathcal{W}_2\left(\mathcal{L}\left(x_i^{(k)}\right)\,,\, \pi \right) 
\leq \frac{1}{N}\sum_{i=1}^N\mathcal{W}_2\left(\mathcal{L}\left(x_i^{(k)}\right)\,,\, \mathcal{L}\left(\overline{x}^{(k)}\right) \right) + \mathcal{W}_2\left(\mathcal{L}\left(\overline{x}^{(k)}\right)\,,\,\pi \right).
\end{equation}
The first term in \eqref{decompose} can be bounded by Corollary~\ref{corollary:x2avg} as follows:
\begin{align}
\label{decompose:t1} 
\frac{1}{N}\sum_{i=1}^N\mathcal{W}_2\left(\mathcal{L}\left(x_i^{(K)}\right)\,,\, \mathcal{L}\left(\overline{x}^{(K)}\right)  \right) 
&\leq \left(\frac{1}{N}\sum_{i=1}^N\mathbb{E}\left[\left\Vert x_i^{(K)} - \overline{x}^{(K)} \right\Vert^2\right]\right)^{1/2}
\nonumber 
\\
&\leq \eta \cdot \frac{D_1}{\sqrt{N}} + \sqrt{\eta} \cdot D_2 + \frac{2\left(\overline{\gamma}_{{\scaleto{\widetilde{W}}{5pt}}}\right)^{K} }{\sqrt{N}}\sqrt{\mathbb{E}\left[ \left\Vert x^{(0)}\right\Vert^2 \right]},
\end{align}
where 
\begin{align}
D_1 := 2\frac{\sqrt{2(R_{h} + R'_{h})}}{1 - \overline{\gamma}_{{\scaleto{\widetilde{W}}{5pt}}}} + \frac{2\sigma}{\sqrt{1 - \overline{\gamma}_{{\scaleto{\widetilde{W}}{5pt}}}^2}}, 
\qquad
D_2 := 2\sqrt{\frac{2d}{1 - \overline{\gamma}_{{\scaleto{\widetilde{W}}{5pt}}}^2}}.\label{defn:D:1:2}
\end{align}

\subsubsection{$L^2$ distance between $\overline{x}^{(k)}$ and $x_{k}$}

Next, we upper bound the second term in \eqref{decompose}, 
which, according to \eqref{decompose:00}, can be bounded as
\begin{align}
\mathcal{W}_2\left(\mathcal{L}\left(\overline{x}^{(k)}\right)\,,\,\pi \right)\leq \mathcal{W}_2\left(\mathcal{L}\left(\overline{x}^{(k)}\right),\mathcal{L}(x_{k}) \right)
+\mathcal{W}_{2}\left(\mathcal{L}(x_{k}),\pi\right),\label{decompose:two:terms}
\end{align}
where $x_{k}$ given in \eqref{overdamped:iterates:0} is the Euler-Maruyama discretization of the overdamped Langevin SDE \eqref{overdamped:SDE}. 
First, we bound the first term in \eqref{decompose:two:terms}
by providing the an upper bound on the $L^{2}$ distance
between the average iterate $\bar{x}^{(k)}$ and $x_{k}$.

Since $W$ is doubly stochastic, we can compute $\mathcal{W} \overline{\mathbf x}^{(k)} = \overline{\mathbf x}^{(k)}$, that is $\overline{\mathbf x}^{(k)}$ is consensual, similarly, we also have $\frac{1}{N}(1_N1_N^T)^T \overline{\mathbf x}^{(k)} = \overline{\mathbf x}^{(k)}$ for $k = 1,2,\ldots$. The following mean iterates can be found by taking average of~\eqref{alg}.
\begin{equation}
\label{avg:dynamics}
\overline{x}^{(k+1)} = \overline{x}^{(k)} - \eta\frac{1}{N}\sum_{i=1}^N\nabla f_i\left(x_i^{(k)}\right) - \eta\overline{\xi}^{(k)}  + \sqrt{2\eta}\overline{w}^{(k+1)},
\end{equation}
we can find the mean iterates have the same format as the one of decentralized stochastic gradient Langevin dynamics in~\cite{gurbuzbalaban2021decentralized}. Then, we can get
\begin{equation}\label{average:eqn}
\overline{x}^{(k+1)} = \overline{x}^{(k)} - \frac{\eta}{N} \nabla f\left(\overline{x}^{(k)}\right)  + \eta \hat{\mathcal{E}}_{k+1} - \eta\overline{\xi}^{(k)} + \sqrt{2\eta}\overline{w}^{(k+1)},
\end{equation} 
where the error term is 
\begin{equation}
\hat{\mathcal{E}}_{k+1} := \frac{1}{N} \sum_{i=1}^N\left(\nabla f_i\left(\overline{x}^{(k)}\right) - \nabla f_i\left(x_i^{(k)}\right) \right).
\end{equation}
On the other hand, we recall from \eqref{overdamped:iterates:0} that $x_{k}$ is the Euler-Maruyama discretization of overdamped Langevin diffusion \eqref{overdamped:SDE} with the iterates:
\begin{equation}\label{overdamped:iterates}
x_{k+1} = x_{k} - \frac{\eta}{N}\nabla f\left(x_k\right) + \sqrt{2\eta}\overline{w}^{(k+1)}.
\end{equation}
Hence, we get
\begin{equation}
\overline{x}^{(k+1)} - x_{k+1} = \overline{x}^{(k)} - x_{k} - \frac{\eta}{N} \left(\nabla f\left(\overline{x}^{(k)}\right)  - \nabla f\left(x_k\right) \right) + \eta \hat{\mathcal{E}}_{k+1} - \eta\overline{\xi}^{(k)}.
\end{equation}
Now, we are ready to state the next corollary to bound $L^2$ distance between the mean $\overline{x}^{(k)}$ in \eqref{average:eqn} and discretized overdamped Langevin iterate $x_{k}$ in \eqref{overdamped:iterates}.

\begin{corollary}
\label{corollary:avg2EulerOverdamped}
Suppose $\eta < \frac{2}{L} \wedge 1$, under assumptions in Corollary~\ref{corollary:x2avg}, then there holds
\begin{align}
\mathbb{E}\left[\left\Vert \overline{x}^{(k)} - x_k \right\Vert^2\right] 
& \leq \frac{\eta\left(\eta + \frac{\left(1 + \eta L\right)^2}{\mu\left(1 - \frac{\eta L}{2}\right)}\right)
\left(\eta^2 \cdot \frac{4L^2}{N}\left(\frac{2\left(R_{h} + R'_{h}\right)}{\left(1 - \overline{\gamma}_{{\scaleto{\widetilde{W}}{5pt}}}\right)^2} + \frac{\sigma^2N}{1 - \overline{\gamma}_{{\scaleto{\widetilde{W}}{5pt}}}^2}\right) 
+ \eta\cdot\frac{8L^2d}{1 - \overline{\gamma}_{{\scaleto{\widetilde{W}}{5pt}}}^2}
\right) + \eta^2\frac{\sigma^2}{N}}{\eta\mu\left(1 - \frac{\eta L}{2} \right)}
\nonumber 
\\
& \qquad\qquad +\frac{\overline{\gamma}_{{\scaleto{\widetilde{W}}{5pt}}}^{2 k}-\left(1-\eta \mu\left(1-\frac{\eta L}{2}\right)\right)^k}{(\overline{\gamma}_{{\scaleto{\widetilde{W}}{5pt}}})^{2}-1+\eta \mu\left(1-\frac{\eta L}{2}\right)} \frac{4 L^2(\overline{\gamma}_{{\scaleto{\widetilde{W}}{5pt}}})^{2}}{N} \mathbb{E}\left\|x^{(0)}\right\|^2,
\end{align}
where the constants $R_{h},R'_{h}$ and $\overline{\gamma}_{{\scaleto{\widetilde{W}}{5pt}}}$ are defined in Lemma~\ref{lemma:bound:grad:tilde:v} and Corollary~\ref{corollary:x2avg}.
\end{corollary}

\subsubsection{$\mathcal{W}_{2}$ distance between the law of $x_{k}$ and the Gibbs distribution $\pi$}

Next, we provide the 2-Wassestein distance between
the law of $x_{k}$ in \eqref{overdamped:iterates}, 
which is the Euler-Maruyama discretization of \eqref{overdamped:SDE}
and the Gibbs distribution $\pi$. 
We note that the function $\frac{1}{N}f$ is $\mu$-strongly convex and $L$-smooth.
We simply quote an existing result, that is, Theorem~4 from~\cite{dalalyan2019user} restated in the next lemma.

\begin{lemma}[Theorem~4 in \cite{dalalyan2019user}]
\label{lemma:Euler2Gibbs}
For any $\eta \in\left(0, \frac{2 N}{L+\mu}\right]$, we have
\begin{equation}
\mathcal{W}_2\left(\mathcal{L}\left(x_K\right), \pi\right) \leq (1-\mu \eta)^{K} \mathcal{W}_2\left(\mathcal{L}\left(x_0\right), \pi\right)+\frac{1.65 L}{\mu} \sqrt{d N^{-1}}\sqrt{\eta}. 
\end{equation}
\end{lemma}

Now, we are finally ready to complete the proof of Theorem~\ref{thm:main}.


\subsubsection{Completing the proof of Theorem~\ref{thm:main}}

Under our assumptions, the conditions
in Theorem~\ref{theorem:norm:tilde} are satisfied, 
so that one can apply Corollary~\ref{corollary:x2avg}, Corollary~\ref{corollary:avg2EulerOverdamped} and Lemma~\ref{lemma:Euler2Gibbs}. 
Note that Corollary~\ref{corollary:avg2EulerOverdamped} and Lemma~\ref{lemma:Euler2Gibbs} give the bound on the second term in decomposition~\eqref{decompose} such that it follows that:
\begin{align}
\label{decompose:t2}
& \mathcal{W}_2\left(\mathcal{L}\left(\overline{x}^{(k)}\right)\,,\,\pi \right) 
\nonumber 
\\
& \quad \leq \left(\mathbb{E}\left[\left\Vert \overline{x}^{(k)} - x_{k} \right\Vert^2\right]\right)^{1/2}  + \mathcal{W}_2\left(\mathcal{L}\left(x_{k}\right), \pi\right)
\nonumber 
\\
& \quad \leq 
\left(\frac{\overline{\gamma}_{{\scaleto{\widetilde{W}}{5pt}}}^{2 k}-\left(1-\eta \mu\left(1-\frac{\eta L}{2}\right)\right)^k}{\overline{\gamma}_{{\scaleto{\widetilde{W}}{5pt}}}^2-1+\eta \mu\left(1-\frac{\eta L}{2}\right)}\right)^{1 / 2} \frac{2 L \overline{\gamma}_{{\scaleto{\widetilde{W}}{5pt}}}}{\sqrt{N}}\left(\mathbb{E}\left\|x^{(0)}\right\|^2\right)^{1 / 2}
 + (1-\mu \eta)^{K} \mathcal{W}_2\left(\mathcal{L}\left(x_0\right), \pi\right) + \sqrt{\eta}\mathcal{E}_1,
\end{align}
where we have
\begin{align}
\mathcal{E}_1:=& \left(\frac{\eta}{\mu\left(1-\frac{\eta L}{2}\right)}+\frac{(1+\eta L)^2}{\mu^2\left(1-\frac{\eta L}{2}\right)^2}\right)^{1 / 2} \cdot\left(\frac{4 L^2 \left(R_h + R_h'\right) \eta}{N(1-\overline{\gamma}_{{\scaleto{\widetilde{W}}{5pt}}})^2}+\frac{4 L^2 \sigma^2 \eta}{1-\overline{\gamma}_{{\scaleto{\widetilde{W}}{5pt}}}^2}+\frac{8 L^2 d}{1-\overline{\gamma}_{{\scaleto{\widetilde{W}}{5pt}}}^2}\right)^{1 / 2} 
\nonumber 
\\
& \qquad +\frac{\sigma}{\sqrt{\mu\left(1-\frac{\eta L}{2}\right) N}}+\frac{1.65 L}{\mu} \sqrt{d N^{-1}},
\end{align}
and this proves \eqref{decompose:t2:main}. Finally, by~\eqref{decompose},~\eqref{decompose:t1} (which follows from Corollary~\ref{corollary:x2avg}) and~\eqref{decompose:t2}, 
we can derive that
\begin{align}
\label{eqn:temp}
& \frac{1}{N}\sum_{i=1}^N\mathcal{W}_2\left(\mathcal{L}\left(x_i^{(K)}\right)\,,\, \pi \right) 
\nonumber 
\\
& \, \leq \eta \cdot \frac{D_1}{\sqrt{N}} + \sqrt{\eta} \cdot \left(D_2 + \mathcal{E}_1\right) + \left(\frac{\overline{\gamma}_{{\scaleto{\widetilde{W}}{5pt}}}^{2 K}-\left(1-\eta \mu\left(1-\frac{\eta L}{2}\right)\right)^K}{\overline{\gamma}_{{\scaleto{\widetilde{W}}{5pt}}}^2-1+\eta \mu\left(1-\frac{\eta L}{2}\right)}\right)^{1 / 2} \frac{2 L \overline{\gamma}_{{\scaleto{\widetilde{W}}{5pt}}}}{\sqrt{N}}\left(\mathbb{E}\left\|x^{(0)}\right\|^2\right)^{1 / 2}
\nonumber 
\\
& \qquad\qquad\qquad\qquad\qquad + (1-\mu \eta)^{K} \mathcal{W}_2\left(\mathcal{L}\left(x_0\right), \pi\right)
+ \frac{2\left(\overline{\gamma}_{{\scaleto{\widetilde{W}}{5pt}}}\right)^{K} }{\sqrt{N}}\sqrt{\mathbb{E}\left[ \left\Vert x^{(0)}\right\Vert^2 \right]},
\end{align}
and this completes the proof.

\subsection{Proof of Proposition~\ref{prop:comparison}}

First, we consider generalized EXTRA SGLD.
We recall that one can take $1 > \delta^{2}=1 - \frac{\eta\mu}{2}\left(1 - \frac{\eta L}{2} \right)\geq 1-\mu\eta\geq\overline{\gamma}_{{\scaleto{\widetilde{W}}{5pt}}}$ where $\delta^{2}$ 
satisfies the constraint in \eqref{delta:choice}. 
Therefore, for any sufficiently large $K$, the term $\left(1-\eta \mu\left(1-\frac{\eta L}{2}\right)\right)^K$ dominates both terms $(1-\mu\eta)^K$ and $\left(\overline{\gamma}_{{\scaleto{\widetilde{W}}{5pt}}}\right)^{2K}$. Hence, we can get the order of the last three terms in~\eqref{eqn:temp} in Theorem~\ref{thm:main} is $\mathcal{O}\left(\left(1 - \eta\mu\left(1 - \eta L/2 \right)\right)^{K/2}\right)$.

Since $1-x\leq e^{-x}$ for any $0\leq x\leq 1$, we conclude from Theorem~\ref{thm:main},
for any $K\geq K_{0}$, with $K_{0}$ given in \eqref{const:K:0},
 \begin{equation}
 \label{compare:extra}
\frac{1}{N}\sum_{i=1}^N\mathcal{W}^{\mathrm{extra-sgld}}_2\left(\mathcal{L}\left(x_i^{(K)}\right)\,,\, \pi \right) 
\leq 
\mathcal{O}\left(\sqrt{\eta}\left(\sqrt{d} + \mathcal{E}_1\right)\right) 
+ \mathcal{O}\left(e^{-\frac{\eta\mu}{2}\left(1 - \frac{\eta L}{2} \right)K}\right),
\end{equation}
where $\mathcal{E}_{1}$ is defined in \eqref{defn:mathcal:E:1}.
Hence, for any $\varepsilon\rightarrow 0$, we have
\begin{equation}
\frac{1}{N}\sum_{i=1}^N\mathcal{W}^{\mathrm{extra-sgld}}_2\left(\mathcal{L}\left(x_i^{(K)}\right)\,,\, \pi \right) 
\leq\mathcal{O}(\varepsilon),
\end{equation}
provided that
\begin{equation}
\eta\leq\mathcal{O}\left(\frac{\varepsilon^{2}}{(\sqrt{d}+\mathcal{E}_{1})^{2}}\right),
\end{equation}
and
\begin{equation}\label{K:eqn}
K \geq K^{\mathrm{extra-sgld}}:=K_{0}\vee K_{1}:=K_{0}\vee\mathcal{O}\left(\frac{\log(1/\varepsilon)}{\eta\mu}\right)
=K_{0}\vee\mathcal{O}\left(\frac{\log(1/\varepsilon)(\sqrt{d}+\mathcal{E}_{1})^{2}}{\varepsilon^{2}\mu}\right),
\end{equation}
where $K_{0}$ is given in Table~\ref{table_constants}. We recall the definition of $\mathcal{E}_1$  from \eqref{defn:mathcal:E:1}:
\begin{align*}
\mathcal{E}_1=& \left(\frac{\eta}{\mu\left(1-\frac{\eta L}{2}\right)}+\frac{(1+\eta L)^2}{\mu^2\left(1-\frac{\eta L}{2}\right)^2}\right)^{1 / 2} \cdot\left(\frac{4 L^2 \left(R_h + R_h'\right) \eta}{N(1-\overline{\gamma}_{{\scaleto{\widetilde{W}}{5pt}}})^2}+\frac{4 L^2 \sigma^2 \eta}{1-\overline{\gamma}_{{\scaleto{\widetilde{W}}{5pt}}}^2}+\frac{8 L^2 d}{1-\overline{\gamma}_{{\scaleto{\widetilde{W}}{5pt}}}^2}\right)^{1 / 2} 
\nonumber 
\\
& \qquad +\frac{\sigma}{\sqrt{\mu\left(1-\frac{\eta L}{2}\right) N}}+\frac{1.65 L}{\mu} \sqrt{d N^{-1}},
\end{align*}
where the constants are given in Table~\ref{table_constants}.

Under our assumption $\eta L=\mathcal{O}(1)$ such that $\mathcal{O}\left(\left(\frac{\eta}{\mu\left(1-\frac{\eta L}{2}\right)}+\frac{(1+\eta L)^2}{\mu^2\left(1-\frac{\eta L}{2}\right)^2}\right)^{1 / 2}\right)=\mathcal{O}(1/\mu)$.
Thus
\begin{equation}
\mathcal{E}_1 = \mathcal{O}\left(\frac{1}{\mu}\left(L\sqrt{\eta}\sqrt{R_h + R_h'}+L\sqrt{d}\right) \right).
\end{equation}
Therefore, $\frac{1}{N}\sum_{i=1}^N\mathcal{W}^{\mathrm{extra-sgld}}_2\left(\mathcal{L}\left(x_i^{(K)}\right)\,,\, \pi \right) 
\leq\mathcal{O}(\varepsilon)$ provided that
\begin{equation}
\label{eqn:complx}
K\geq K^{\mathrm{extra-sgld}}=K_{0}\vee K_{1}= K_{0}\vee\tilde{\mathcal{O}}\left(\frac{L^{2}(\eta (R_{h}+R'_{h})+d)}{\varepsilon^{2}\mu^{3}}\right),
\end{equation}
where $\tilde{\mathcal{O}}$ hides the logarithmic dependence on $\varepsilon$.

We will show that under our assumptions $K_{0}=\mathcal{O}\left(\frac{1}{\eta\mu}\vee\frac{1}{h}\right)$
and under the assumption $h \leq \frac{1}{(L/\mu)^4(L + \Vert B \Vert^2)}$, we will have $\eta(R_{h}+R'_{h})\leq\mathcal{O}(d)$
and under the assumption $h\geq\Omega(\eta\mu)$, 
we have $K_{0}=\mathcal{O}\left(\frac{1}{\eta\mu}\right)$
such that $K\geq K^{\mathrm{extra-sgld}}=K_{0}\vee K_{1}=\tilde{\mathcal{O}}\left(\frac{1}{\eta\mu}\vee\frac{L^2d}{\varepsilon^{2}\mu^{3}}\right)=\tilde{\mathcal{O}}\left(\frac{L^2d}{\varepsilon^{2}\mu^{3}}\right)$.

As a first step of the proof, we spell out the dependence of the constants in Table~\ref{table_constants} on $L,\mu,d,h,\eta,\Vert B \Vert^2$, 
and we summarize the results in Table~\ref{table_constants_2}. 
Next, we compute from \eqref{R:h}-\eqref{R:h:prime}:
\begin{align}
& \eta R_h=  h\eta\delta^2\left(\frac{C_1\gamma_2}{2L^2} +  \frac{C_0\gamma_1\gamma_2}{2L^2} \right) + h\delta^{2}\left( \gamma_2D_0 + \frac{w_2}{N}\left(\eta\sigma^2 + 2d\right) \right) + \eta\left\Vert \nabla F(\mathbf{x}_*) \right\Vert^2,
\label{eta:R:h}
\\
& \eta R_h'= \eta^2\delta^2\left(C_1 + C_3 + \gamma_1C_0 + D_0C_2\right) + \delta^2\eta^3\left(\frac{C_1C_2}{2L^2} + \frac{\gamma_1C_0C_2}{2L^2}\right) 
 + 3\eta\left\Vert \nabla F(\mathbf{x}_*)\right\Vert^2.
 \label{eta:R:h:prime}
\end{align}
Under the assumption $h \geq\Omega(\eta\mu)$, in particular, for 
$h/(\eta\mu) \geq \frac{1}{L + \Vert B \Vert^2}$, we get
\begin{equation}
C_1 = \mathcal{O}\left(d(L + \Vert B \Vert^2)(L^{3}/\mu^{2})(h/\eta) + (L^2d)(L/\mu) \right)\leq \mathcal{O}\left(d(L + \Vert B \Vert^2)L(L/\mu)^2(h/\eta)\right).
\end{equation}
Moreover, from Table~\ref{table_constants_2}, we have:
\begin{align}
\frac{\gamma_2}{2L^2}\left(C_1 + C_0\gamma_1\right) & = \mathcal{O}\left(dL(L + \Vert B \Vert^2)^2(L/\mu)^4(h/\eta)\right),
\label{eta:R:1}\\
\gamma_2D_0 + \frac{w_2}{N}\left(\eta\sigma^2 + 2d\right) 
& = \mathcal{O}\left((L/\mu)L^2(L + \Vert  B \Vert^2)((L/\mu)^2 \vee d)\right),
\label{eta:R:2}\\
C_1 + C_3 + \gamma_1C_0 + D_0C_2 & = \mathcal{O}\left((h/\eta)dL(L/\mu)^2(L + \Vert B \Vert^2)\right) + \mathcal{O}\left(Ld(L/\mu)(1/\eta) \right) 
\nonumber 
\\
& \qquad\qquad + \mathcal{O}\left((L/\mu)^2 L(h/\eta)(L + \Vert B \Vert^2) \right) + \mathcal{O}\left(L^2(L/\mu)^3(1/\eta)\right)
\nonumber 
\\
& = \mathcal{O}\left((L/\mu)(h/\eta)(L + \Vert B \Vert^2)Ld \vee L^2d(L/\mu)^3(1/\eta) \right)
\label{eta:R:prime:1}\\
\frac{C_2}{2L^2}\left(C_1 + C_0\gamma_1\right) & = \mathcal{O}\left(L(L/\mu)^4(L + \Vert B \Vert^2)(h/\eta)(d/\eta) \right).
\label{eta:R:prime:2}
\end{align}
Now we can compute~\eqref{eta:R:h} and~\eqref{eta:R:h:prime} as follows. We first use~\eqref{eta:R:1} and~\eqref{eta:R:2} to get
\begin{align}
\eta R_h & = \mathcal{O}\left(h\eta \cdot dL(L + \Vert B \Vert^2)^2(L/\mu)^4(h/\eta) + h \cdot (L/\mu)L^2(L + \Vert  B \Vert^2)((L/\mu)^2 \vee d) \right)
\nonumber 
\\
& = \mathcal{O}\left(h^2 \cdot dL(L + \Vert B \Vert^2)^2(L/\mu)^4 + h \cdot (L/\mu)L^2(L + \Vert  B \Vert^2)((L/\mu)^2 \vee d) \right)
\nonumber 
\\
& \leq \mathcal{O}\left(hdL(L + \Vert B \Vert^2)\left(1 + (L/\mu)\left(\frac{(L/\mu)^2}{d} \vee 1 \right) \right)\right)
\label{r:h:1}
\\
& \leq \mathcal{O}\left(hdL(L/\mu)(L + \Vert B \Vert^2)\left(1 + \left(\frac{(L/\mu)^2 + d}{d} \right) \right)\right)
\nonumber 
\\
& \leq \mathcal{O}\left(hdL(L/\mu)(L + \Vert B \Vert^2)\cdot \frac{(L/\mu)^2d}{d}\right)
= \mathcal{O}\left(hdL(L/\mu)^3(L + \Vert B \Vert^2)\right)\,,
\end{align}
where we used the assumption
\begin{equation}
\label{h:bound:1}
h \leq \frac{1}{(L/\mu)^4(L + \Vert B \Vert^2)},
\end{equation}
to get $h^2 dL(L + \Vert B \Vert^2)^2(L/\mu)^4 \leq hdL(L + \Vert B \Vert^2)$ in~\eqref{r:h:1}. 
Next, we use~\eqref{eta:R:prime:1} and~\eqref{eta:R:prime:2} to compute that
\begin{align}
\eta R_h' & = \mathcal{O}\left( \eta^2 \cdot \left((L/\mu)(h/\eta)(L + \Vert B \Vert^2)Ld \vee dL^2(L/\mu)^3(1/\eta)\right)  + \eta^3 \cdot L(L/\mu)^4(L + \Vert B \Vert^2)(h/\eta)(d/\eta)\right) 
\nonumber 
\\
& = \mathcal{O}\left( \eta \cdot \left((L/\mu)h(L + \Vert B \Vert^2)Ld \vee dL^2(L/\mu)^3\right) + (L/\mu)^4(L + \Vert B \Vert^2)hd\right) 
\label{r:h:prime:1} 
\\
&\leq \mathcal{O}\left(\eta dL^2(L/\mu^3) + (L/\mu)^4(L + \Vert B \Vert^2)hd\right) 
\label{r:h:prime:2} 
\\
&\leq \mathcal{O}\left(h(d/\mu)(L/\mu)^3 + (L/\mu)^4(L + \Vert B \Vert^2)hd\right) 
\label{r:h:prime:3}
\\
& = \mathcal{O}\left(hd(L/\mu) \cdot (L/\mu)^3(L + \Vert B \Vert^2)\right),
\end{align}
where we used $\eta L \leq 1$ to get~\eqref{r:h:prime:1}, and moreover, we used the assumption~\eqref{h:bound:1} to get~\eqref{r:h:prime:2} and then used the assumption $h \geq \Omega(\eta\mu)$ again to get~\eqref{r:h:prime:3}.
As a consequence, we can compute that
\begin{equation}
\eta(R_h + R_h') = \mathcal{O}\left(hd(L/\mu)^4(L + \Vert B \Vert^2)\right) \leq \mathcal{O}\left(d\right),
\end{equation}
where we use~\eqref{h:bound:1} again in the last inequality. Now we can compute the term in~\eqref{eqn:complx} such that 
\begin{equation}\label{final:complex}
\tilde{\mathcal{O}}\left(\frac{L^{2}(\eta (R_{h}+R'_{h})+d)}{\varepsilon^{2}\mu^{3}}\right) = \tilde{\mathcal{O}}\left(\frac{L^2d}{\ep^2\mu^3}\right).
\end{equation}
To compute the term $K_0$ in~\eqref{eqn:complx}, we use Table~\ref{table_constants_2} to get
\begin{align}
D_0 + C_4 & = \mathcal{O}(L(L/\mu)(1/\eta)), \quad C_0 = \mathcal{O}(L^2(L/\mu)(h/\eta)(L + \Vert B \Vert^2)),
\\
\delta^2 & = \mathcal{O}(1), \quad \frac{1}{1 - \delta^2} = \mathcal{O}\left(\frac{1}{\eta\mu} \vee \frac{1}{h}\right).
\end{align}
Under the setting in~\eqref{h:bound:1}, we have $h \leq \frac{1}{L(L + \Vert B \Vert^2)}$, then we get
\begin{align}
\left(1 - \frac{\left\Vert \nabla F(\mathbf{x}_*) \right\Vert^2}{D_0 + C_4}\right)  \vee \left(1 - \frac{\left\Vert \nabla F(\mathbf{x}_*) \right\Vert^2}{C_0}\right) & = \mathcal{O}\left(1 - \frac{1}{L(L/\mu)(1/\eta)}\right) = \mathcal{O}(1).
\end{align}
Hence, we can compute from~\eqref{const:K:0} such that 
\begin{equation}
\label{eqn:asy:1}
K_0 =\frac{\delta^2}{1 - \delta^2}\left[\left(1 - \frac{\left\Vert \nabla F(\mathbf{x}_*) \right\Vert^2}{D_0 + C_4}\right)  \vee \left(1 - \frac{\left\Vert \nabla F(\mathbf{x}_*) \right\Vert^2}{C_0}\right)  \right] \vee 0
= \mathcal{O}\left(\frac{1}{\eta\mu} \vee \frac{1}{h}\right)
= \mathcal{O}\left(\frac{1}{\eta\mu}\right),
\end{equation}
where we used the assumption $h \geq \Omega(\eta\mu)$. 
We recall from~\eqref{compare:extra} and~\eqref{eqn:complx} that for generalized EXTRA SGLD:
\begin{equation}
\frac{1}{N}\sum_{i=1}^N\mathcal{W}^{\mathrm{extra-sgld}}_2\left(\mathcal{L}\left(x_i^{(K)}\right)\,,\, \pi \right) \leq 
\mathcal{O}\left(\sqrt{\eta}\left(\sqrt{d} + \mathcal{E}_1\right)\right) 
+ \mathcal{O}\left(e^{-\frac{\eta\mu}{2}\left(1 - \frac{\eta L}{2} \right)K}\right).
\end{equation}
Hence, we conclude that for any $\varepsilon\rightarrow 0$,
\begin{equation}
\frac{1}{N}\sum_{i=1}^N\mathcal{W}^{\mathrm{extra-sgld}}_2\left(\mathcal{L}\left(x_i^{(K)}\right)\,,\, \pi \right) 
\leq\mathcal{O}(\varepsilon),
\end{equation}
provided that
\begin{equation}
\eta\leq\mathcal{O}\left(\frac{\varepsilon^{2}}{(\sqrt{d}+\mathcal{E}_{1})^{2}}\right),
\qquad\text{and}
\qquad
K\geq K^{\mathrm{extra-sgld}}=K_0 \vee K_{1}= \tilde{\mathcal{O}}\left(\frac{L^2d}{\varepsilon^{2}\mu^{3}}\right),
\end{equation}
where we used \eqref{eqn:complx}, \eqref{final:complex}, \eqref{eqn:asy:1} and $K_{1}=\mathcal{O}\left(\frac{\log(1/\varepsilon)}{\eta\mu}\right)=\tilde{\mathcal{O}}\left(\frac{\log(1/\varepsilon)}{\eta\mu}\right)$ from \eqref{K:eqn}.

Next, we consider DE-SGLD. For DE-SGLD, it follows from Theorem~1 in~\cite{gurbuzbalaban2021decentralized} that: 
\begin{equation}\label{compare:sgld}
\frac{1}{N} \sum_{i=1}^N \mathcal{W}^{\mathrm{de-sgld}}_2\left(\mathcal{L}\left(x_i^{(K)}\right), \pi\right) 
\leq\mathcal{O}\left(e^{-\frac{\eta\mu}{2}\left(1 - \frac{\eta L}{2} \right)K}\right)
+\mathcal{O}\left(\sqrt{\eta}\left(\sqrt{d} + \mathcal{E}_1'\right)\right),
\end{equation}
where
\begin{align*}
\mathcal{E}_1' : & = \frac{1.65 L}{\mu} \sqrt{d N^{-1}}+\frac{\sigma}{\sqrt{\mu\left(1-\frac{\eta L}{2}\right) N}} 
\\
& \qquad +\left(\frac{\eta}{\mu\left(1-\frac{\eta L}{2}\right)}+\frac{(1+\eta L)^2}{\mu^2\left(1-\frac{\eta L}{2}\right)^2}\right)^{1 / 2} \cdot\left(\frac{4 L^2 D^2 \eta}{N(1-\overline{\gamma}_{{\scaleto{\widetilde{W}}{5pt}}})^2}+\frac{4 L^2 \sigma^2 \eta}{1-\overline{\gamma}_{{\scaleto{\widetilde{W}}{5pt}}}^2}+\frac{8 L^2 d}{1-\overline{\gamma}_{{\scaleto{\widetilde{W}}{5pt}}}^2}\right)^{1 / 2},
\end{align*}
with $\overline{\gamma}_{{\scaleto{\widetilde{W}}{5pt}}}:=\max \left\{\left|\lambda_2^{{\scaleto{\widetilde{W}}{5pt}}}\right|,\left|\lambda_N^{{\scaleto{\widetilde{W}}{5pt}}}\right|\right\} \in[0,1) $ and the constant (see Lemma~6 in~\cite{gurbuzbalaban2021decentralized})
\begin{equation}
\label{sgld:D}
D^2:=4 L^2 \mathbb{E}\left\Vert x^{(0)}-x_*\right\Vert ^2+8 L^2 \frac{\hat{C}_1^2 \eta^2 N}{(1-\overline{\gamma}_{{\scaleto{\widetilde{W}}{5pt}}})^2}+\frac{2 L^2\left(\eta \sigma^2 N+2 d N\right)}{\mu\left(1+\lambda_N^{{\scaleto{\widetilde{W}}{5pt}}}-\eta L\right)}+4\left\Vert \nabla F\left(\mathbf{x}_*\right)\right\Vert ^2,
\end{equation}
where
$$
\hat{C}_1:=\bar{C}_1 \cdot\left(1+\frac{2(L+\mu)}{\mu}\right), \quad\text{with}\quad \bar{C}_1:=\sqrt{2 L \sum_{i=1}^N\left(f_i(0)-f_i^*\right)}, \quad f_i^*:=\min _{x \in \mathbb{R}^d} f_i(x).
$$
Hence, by $\eta L = \mathcal{O}(1)$ and $\hat{C}_1^2 = \mathcal{O}(L^3/\mu^2)$, we can compute that 
\begin{equation}
D^2 =\mathcal{O}\left(L^2 +L^3/\mu^2+Ld(L/\mu)\right)=\mathcal{O}(L^{3}d/\mu).
\end{equation}
Therefore, we have
\begin{align}
\mathcal{E}'_1 & = \mathcal{O}\left(\frac{1}{\mu}\left(L\sqrt{\eta}D+L\sqrt{d}\right) \right) = \mathcal{O}\left(\frac{1}{\mu}\left(D\sqrt{L}+L\sqrt{d}\right) \right) 
\nonumber 
\\
& = \mathcal{O}\left( \sqrt{Ld}(L/\mu)\sqrt{L/\mu}+ (L/\mu)\sqrt{d}\right)
= \mathcal{O}\left(\sqrt{Ld}(L/\mu)\sqrt{L/\mu}\right),
\end{align}
where we use $\eta \leq 1/L$ to get the second equality in the first line. By following ~\eqref{compare:extra} and~\eqref{eqn:complx}, we conclude that 
 \begin{align}
\frac{1}{N}\sum_{i=1}^N\mathcal{W}^{\mathrm{de-sgld}}_2\left(\mathcal{L}\left(x_i^{(K)}\right)\,,\, \pi \right) 
&\leq 
\mathcal{O}\left(\sqrt{\eta}\left(\sqrt{d} + \mathcal{E}'_1\right)\right) 
+ \mathcal{O}\left(e^{-\frac{\eta\mu}{2}\left(1 - \frac{\eta L}{2} \right)K}\right)
\nonumber
\\
&\leq 
\mathcal{O}\left(\sqrt{\eta}\left(\sqrt{d} + \sqrt{Ld}(L/\mu)\sqrt{L/\mu}\right)\right) 
+ \mathcal{O}\left(e^{-\frac{\eta\mu}{2}\left(1 - \frac{\eta L}{2} \right)K}\right)\,.
\end{align}
Hence, we conclude that for any $\varepsilon\rightarrow 0$, we have
\begin{equation}
\frac{1}{N}\sum_{i=1}^N\mathcal{W}^{\mathrm{de-sgld}}_2\left(\mathcal{L}\left(x_i^{(K)}\right)\,,\, \pi \right) 
\leq\mathcal{O}(\varepsilon),
\end{equation}
provided that
\begin{equation}
\eta\leq\mathcal{O}\left(\frac{\varepsilon^{2}}{(\sqrt{d} + \sqrt{Ld}(L/\mu)\sqrt{L/\mu})^{2}}\right)
\qquad\text{and}
\qquad
K \geq K^{\mathrm{de-sgld}}:= \tilde{\mathcal{O}}\left(\frac{L^4 d}{\varepsilon^{2}\mu^{3}}\right).
\end{equation}
The proof is complete.

\begin{table}{\emph{Constants} \hfill \emph{Source}}

\centering

\rule{\textwidth}{\heavyrulewidth}

\vspace{-1.6\baselineskip}
\begin{flalign*}
&
\gamma_1 = \frac{1}{\gamma_{{\scaleto{\widetilde{W}}{5pt}}}}\left(\frac{1}{L} + 2 + \frac{1}{L\mu}\right) = \mathcal{O}(1/\mu)
\end{flalign*}

\vspace{-1.6\baselineskip}
\begin{flalign*}
&
\gamma_2 = \frac{12\left(L^2 + L\left\Vert B \right\Vert^2 \right)}{(1-\overline{\gamma}_{{\scaleto{W}{3pt}}})\left(1 - \overline{\gamma}_{{\scaleto{I_{N}-W}{5pt}}}^2\right)}\left(1 + \frac{4L^2\left(1 + \frac{2 + 2L}{\mu} \right)}{N^2\mu}\right) = \mathcal{O}\left(L^2(L/\mu)^2(L + \Vert B \Vert^2) \right)
\end{flalign*}

\vspace{-1.6\baselineskip}
\begin{flalign*}
&
w_1 = 2\left(\frac{N^2 + 1}{\gamma_{{\scaleto{\widetilde{W}}{5pt}}}} + \frac{4}{\gamma_{{\scaleto{\widetilde{W}}{5pt}}}}\cdot\left(\frac{L}{\mu} + 3\eta L - 1\right) \right) = \mathcal{O}\left(L/\mu \right)
\end{flalign*}

\vspace{-1.6\baselineskip}
\begin{flalign*}
&
w_2 =\frac
{8\left(6\left(L^2 + L\left\Vert B \right\Vert^2 \right)+N^{2}\mu\right)}{N\mu(1-\overline{\gamma}_{{\scaleto{W}{3pt}}})\left(1 - \overline{\gamma}_{{\scaleto{I_{N}-W}{5pt}}}^2\right)} = \mathcal{O}\left( (L/\mu)(L + \Vert B \Vert^2)\right)
\end{flalign*}

\vspace{-1.6\baselineskip}
\begin{flalign*}
&E_1 = \frac{8}{\gamma_{{\scaleto{\widetilde{W}}{5pt}}}}\left(L/\mu + 3\eta L - 1\right) = \mathcal{O}(L/\mu),
\quad
E_2 = \frac{2}{\gamma_{{\scaleto{\widetilde{W}}{5pt}}}} = \mathcal{O}(1)
\end{flalign*}

\vspace{-1.6\baselineskip}
\begin{flalign*}
&E_3 = \frac{12\left(L^2 + L\left\Vert B \right\Vert^2 \right)}{\mu(1-\overline{\gamma}_{{\scaleto{W}{3pt}}})\left(1 - \overline{\gamma}_{{\scaleto{I_{N}-W}{5pt}}}^2\right)} = \mathcal{O}\left((L/\mu)(L + \Vert B \Vert^2) \right), 
\quad E_4 = \frac{4}{(1-\overline{\gamma}_{{\scaleto{W}{3pt}}})\left(1 - \overline{\gamma}_{{\scaleto{I_{N}-W}{5pt}}}^2\right)} = \mathcal{O}(1)
\end{flalign*}

\vspace{-1.6\baselineskip}
\begin{flalign*}
& C_0 = \left((h/\eta)E_3\mathbb{E}\left[ \left\Vert \overline{e}_{x}^{(0)}\right\Vert^2 \right] + E_4\mathbb{E}\left[ \left\Vert \widetilde{v}^{(0)}\right\Vert^2 \right] \right)\cdot \frac{2L^2}{1 - \eta\gamma_1\gamma_2} = \mathcal{O}\left(L^2(L/\mu)(h/\eta)(L + \Vert B \Vert^2) \right)
\end{flalign*}

\vspace{-1.6\baselineskip}
\begin{flalign*}
& C_1 = \frac{2L^2\left(\eta \sigma^2 + 2d\right)}{N} \cdot \frac{w_2\gamma_1(h/\eta) + w_1}{1 - h\gamma_1\gamma_2} = 
\mathcal{O}\left(Ld(L + \Vert B \Vert^2)(L/\mu)^2(h/\eta) + (L^2d)(L/\mu) \right)
\end{flalign*}

\vspace{-1.6\baselineskip}
\begin{flalign*}
& C_2 = \frac{2L^4\left(\eta + \frac{1 + \eta L}{\mu\left(1-\frac{\eta L}{2}\right)} \right)}{N^2\left(\delta^2 +\eta\mu\left(1 - \frac{\eta L}{2}\right) - 1 \right)} = \mathcal{O}\left(L^2(L/\mu)^2(1/\eta)\right)\end{flalign*}

\vspace{-1.6\baselineskip}
\begin{flalign*}
& C_3 = \frac{2L^2}{N} \cdot \frac{\eta\sigma^2 + 2d}{\delta^2 +\eta\mu\left(1 - \frac{\eta L}{2}\right) - 1} = \mathcal{O}\left(Ld(L/\mu)(1/\eta)\right)
\end{flalign*}

\vspace{-1.6\baselineskip}
\begin{flalign*}
& C_4 =  \frac{2L^2}{\delta^2 +\eta\mu\left(1 - \frac{\eta L}{2}\right) - 1}\mathbb{E}\left[ \left\Vert \overline{e}_{x}^{(0)}\right\Vert^2 \right] = \mathcal{O}(L(L/\mu)(1/\eta))
\end{flalign*}

\vspace{-1.6\baselineskip}
\begin{flalign*}
& 
D_0 = \frac{1}{1-h\gamma_1\gamma_2}\left(E_1\mathbb{E}\left[ \left\Vert \widetilde{x}^{(0)}\right\Vert^2 \right] + E_2\mathbb{E}\left[ \left\Vert \overline{e}_{x}^{(0)}\right\Vert^2 \right]\right) = \mathcal{O}(L/\mu)
\end{flalign*}

\vspace{-1.6\baselineskip}
\begin{flalign*}
&\delta^2 \in \left[\left(1 - \frac{\eta\mu}{2}\left(1 - \frac{\eta L}{2} \right)\right) \vee \left(1 - h\frac{1-\overline{\gamma}_{{\scaleto{W}{3pt}}}}{4}\left(1 - \overline{\gamma}_{{\scaleto{I_{N}-W}{5pt}}}\right)\right) \,,\,  1\right),\,  \delta^2 = \mathcal{O}(1),\,  \frac{\delta^2}{1-\delta^2} = \mathcal{O}\left(\frac{1}{\eta\mu}\vee\frac{1}{h}\right)
\end{flalign*}

\rule{\textwidth}{\heavyrulewidth}

\caption{Summary of the constants in the proof of Proposition~\ref{prop:comparison}. }\label{table_constants_2}
\end{table}



\section{Numerical Experiments}

In this section, we present some results from the numerical experiments based on our algorithms and investigate the relative performance of DE-SGLD and EXTRA SGLD. We mainly perform Bayesian linear regression and Bayesian logistic regression by distributing the sample data evenly among the agents or nodes of different network structures. We ensure that each agent receives randomly distributed independent and identically distributed (i.i.d.) sample data. 

\begin{figure}[htbp]
    \centering
    \begin{minipage}[b]{0.22\textwidth}
        \centering
        \begin{tikzpicture}
            \foreach \x in {1,...,6} {
            \node[draw, circle, fill=red!30, minimum size=7mm, inner sep=1.5pt] (node\x) at (360/6*\x:1.5) {};
            }
            \foreach \x in {1,...,6} {
            \foreach \y in {\x,...,6} {
                \draw (node\x) -- (node\y);}
            }
        \end{tikzpicture}
        \caption*{(a) Fully connected}
    \end{minipage}
\hfill
    \begin{minipage}[b]{0.22\textwidth}
        \centering
        \begin{tikzpicture}
            \foreach \x in {1,...,6} {
            \node[draw, circle, fill=blue!30, minimum size=7mm, inner sep=1.5pt] (node\x) at (360/6*\x:1.5) {};
            }
            \foreach \x in {1,...,5} {
            \pgfmathtruncatemacro{\next}{mod(\x,6)+1}
            \draw (node\x) -- (node\next);
            }
            \draw (node6) -- (node1);
        \end{tikzpicture}
        \caption*{(b) Circular}
    \end{minipage}
\hfill
    \begin{minipage}[b]{0.22\textwidth}
        \centering
        \begin{tikzpicture}
            \node[draw, circle, fill=green!30, minimum size=7mm, inner sep=1.5pt] (center) at (0,0) {};
            \foreach \x in {1,...,5} {
            \node[draw, circle, fill=green!30, minimum size=7mm, inner sep=1.5pt] (node\x) at (360/5*\x:1.5) {};
            \draw (center) -- (node\x);
            }
        \end{tikzpicture}
        \caption*{(c) Star}
    \end{minipage}
\hfill
    \begin{minipage}[b]{0.22\textwidth}
        \centering
        \begin{tikzpicture}
            \foreach \x in {1,...,6} {
            \node[draw, circle, fill=purple!30, minimum size=7mm, inner sep=1.5pt] (node\x) at (360/6*\x:1.5) {};
            }
        \end{tikzpicture}
    \caption*{(d) Disconnected}
    \end{minipage}
\caption{Different types of network structures}
\label{fig:all_networks}
\end{figure}\hfill

Figure~\ref{fig:all_networks} represents four different types of networks: (a) fully connected network, (b) circular network, (c) star network, and (d) fully disconnected network where no agents are connected. A fully connected network is a structure in which all nodes are connected to each other. In contrast, a circular network is one in which each node is only connected to its immediate left and right neighbors. Additionally, in a star-shaped structure, the central node is connected to all other nodes, but those nodes are not connected to each other.

\subsection{Network architecture}

We follow the common approach to select the communication matrix $W=I_{N}-\delta L$ where $I_{N}$ is the $N\times N$ identity matrix, $L$ is the graph Laplacian, and $\delta>0$ is a small number \cite{chung1997spectral}. In our experiments, we select $\delta$ in the following way. First, we compute the graph Laplacian $L=(D_{deg}-A)$ from the degree matrix $D_{deg}$ and the adjacency matrix $A$. The degree $D_{deg}$ is a diagonal matrix with the entries in the main diagonal representing the degree of connections of each node and the adjacent matrix $A=(a_{ij})_{1\leq i,j\leq N}$ is the matrix with $a_{ij}=1$ if there is an edge between the nodes $i$ and $j$ otherwise $a_{ij}=0$. Then we choose $\delta$ at random so that $0<\delta<\frac{2}{\lambda_N^L}$, where $\lambda_N^L$ is the largest real eigenvalue of $L$. For example, a star-like graph with $N$ vertices is given by 
$$W=I_N-\delta L=
\begin{bmatrix}
    1-\delta(N-1) &\delta &\delta & \cdots & \delta & \delta \\
    \delta &1-\delta & 0& 0 & \cdots & 0\\
    \delta & 0 &1-\delta & 0 &\cdots & 0\\
    \vdots &\vdots & \vdots &\vdots &\vdots & \vdots \\
    \delta & 0 & \cdots &0 &1-\delta & 0 \\
    \delta & 0 & \cdots &\cdots &0&1-\delta \\
\end{bmatrix}.
$$
For the EXTRA SGLD algorithm, we compute that $\widetilde{W}=hI_N-(1-h)W$ for $h\in (0,1/2]$.

\subsection{Bayesian linear regression} 
First, we present the Bayesian linear regression with the synthetic data that we generate by simulating the following model:
\begin{equation}
    \delta_i\sim \mathcal{N}(0,\xi^2), \hspace{.5in} X_i\sim \mathcal{N}(0,I_{2}),\hspace{.5in} y_i=\beta^TX_i+\delta_i,
    \label{eq:n1}
\end{equation}
where the white noise $\delta_i$'s are i.i.d. scalars with $\xi=1, \beta\in \mathbb{R}^2$, and $I_{2}$ is the $2\times 2$ identity matrix. The prior distribution of $\beta$ follows $\mathcal{N}(0,\lambda I_{2})$, and we set $\lambda=10$ for this set of experiments. The posterior distribution can be derived from the following model
\begin{equation*}
    \pi(\beta)\sim \mathcal{N}(m,V),\hspace{4mm} m:=\left(\Sigma^{-1}+\frac{X^TX}{\xi^2}\right)^{-1}\left(\frac{X^Ty}{\xi^2}\right),\hspace{4mm} V:=\left(\frac{X^TX}{\xi^2}+\Sigma^{-1}\right)^{-1},
\end{equation*}
where $\Sigma=\lambda I_{2}$ is the covariance matrix of the prior of $\beta$, and $X=\left[X_1^T,X_2^T,\cdots\right]^T$ and $Y=[y_1,y_2,\cdots]^T$ are the input and output matrices, respectively. For this experiment, we simulate 5000 data points using the model \eqref{eq:n1} and then we distribute these data points randomly among the $N=20$ agents. All agents have an equal amount of data exclusively, and share only the parameter estimates. The posterior distribution $\pi(\beta)\propto e^{-f(\beta)}$ where $f(\beta)=\sum_{i=1}^{N}f_i(\beta)$ with
\begin{equation*}
    f_i(\beta):=-\sum_{j=1}^{n_i}\log{ p\left(y_j^i|\beta, X_j^i\right)}-\frac{1}{N}\log{p (\beta)}=\sum_{j=1}^{n_i}\left(y_j^i-\beta^TX_j^i\right)^2+\frac{1}{2\lambda N}\|\beta\|^2,
\end{equation*}
where
\begin{equation*}
    p\left(y_j^i|\beta, X_j^i\right)=\frac{1}{\sqrt{2\pi \xi^2}}e^{-\frac{1}{2\xi^2}(y_j^i-\beta^TX_j^i)^2},\hspace{4mm} p(\beta)\propto e^{-\frac{1}{2\lambda}\|\beta\|^2},
\end{equation*}
and each agent $i$ has an equal number of $n_i=50$ data points $\{(X_j^i,y_j^i)\}_{j=1}^{n_i}$.

\begin{figure}[th]
    \centering
    \includegraphics[scale=0.18]{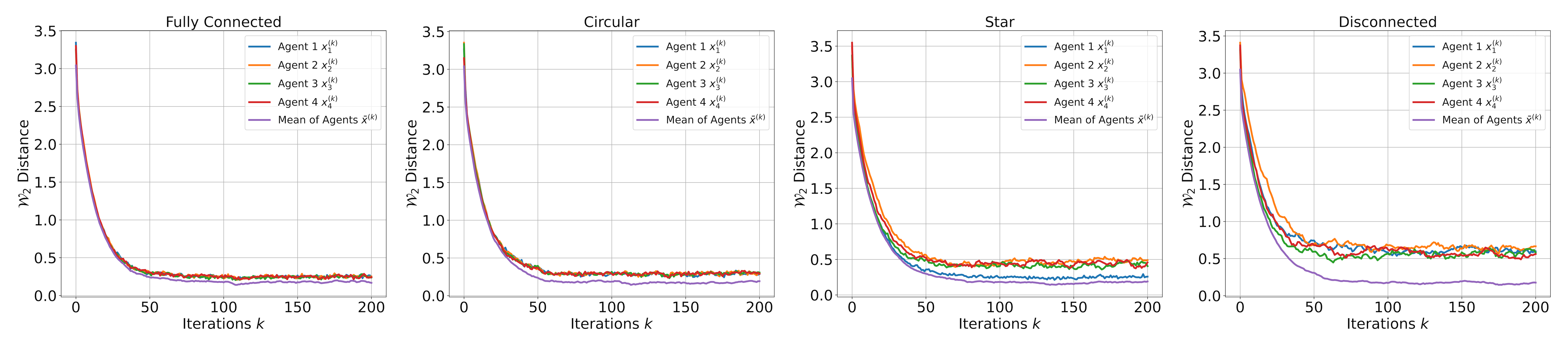}
    \caption{Performance of the EXTRA SGLD for Bayesian linear regression on four different network structures. Out of $20$ agents, we report only the first $4$ agents and the mean of the nodes $\bar{\beta}^{(k)}=\frac{1}{N}\sum_{i=1}^{N}\beta_i^{(k)}$.}
    \label{fig:fig2}
\end{figure}

We report the results of the EXTRA SGLD algorithm as follows.  Figure~\ref{fig:fig2} presents the results of the four networks. We restrict the experiments with a deterministic gradient, i.e. $\sigma=0$ and a fixed step size $\eta=0.009$. The doubly stochastic mixing matrix $\widetilde{W}=hI_N-(1-h)W$ is calculated for different values of the parameter $h$. We consider 5 linearly spaced $h$ values with the minimum being $0.001$ and the maximum being $0.5$ and we tune up the parameter $h$ to the network. For the fully connected network $h=0.50$, circular network $h=0.38$, star network $h=0.13$, and for the disconnected network $h=0.38$. In this setup, the iterations $\beta_i^{(k)}\sim \mathcal{N}\left(m_i^{(k)},\Sigma_i^{(k)}\right)$ for some mean vector $m_i^{(k)}$ and covariance matrix $\Sigma_i^{(k)}$, by using the formula from \cite{givens1984class}, we can  compute the 2-Wasserstein distance, $\mathcal{W}_2$ with the posterior distribution $\pi(\beta)\sim \mathcal{N}(m,V)$. From 200 independent runs, we can estimate $m_i^{(k)}$ and $\Sigma_i^{(k)}$ and then plot the $\mathcal{W}_2$ distance of the stationary distribution for each agent and the distribution of the average $\bar{\beta}^{(k)}=\frac{1}{N}\sum_{i}^{N}\beta_i^{(k)}$. From the plot, we see that for the first three network types, all the agents converge to the posterior distribution up to some error level. However, the convergence for the star-type network is not as good as compared to a fully connected and circular-type network. In the case of a disconnected network, individual agents perform relatively worse compared to other scenarios where the network is connected, as they are unable to leverage information from their neighbors' data points.
We also notice that the convergence is better for strongly connected networks, i.e. as the agent in a network loses its connectivity, the convergence becomes 
slower.
\hfill\\

\begin{figure}[th]
    \centering
    \includegraphics[scale=0.18]{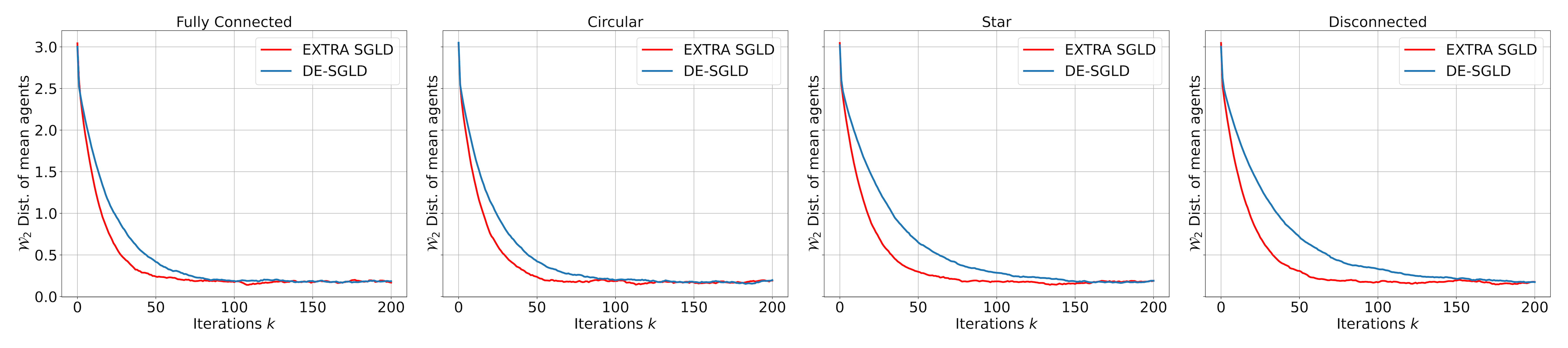}
    \caption{Comparative performance of the DE-SGLD and EXTRA SGLD for Bayesian linear regression on four different network structures in terms of the $\mathcal{W}_2$ distance of mean agents} 
    \label{fig:fig3}
\end{figure}

Next, we present the comparative analysis of the performances of the DE-SGLD and EXTRA SGLD algorithms. In this case, instead of computing the $\mathcal{W}_2$ distances for each agent, we compute the $\mathcal{W}_2$ distances of the mean of the agents from the posterior distribution $\pi(\beta)\sim \mathcal{N}(m,V)$ for all four networks. Then we compute the minimum of these distances for each network and plot them against the mean of the nodes from the DE-SGLD algorithm in the same plot which is represented in Figure~\ref{fig:fig3}.

\hfill\\
\begin{figure}[th]
    \centering
    \includegraphics[scale=0.18]{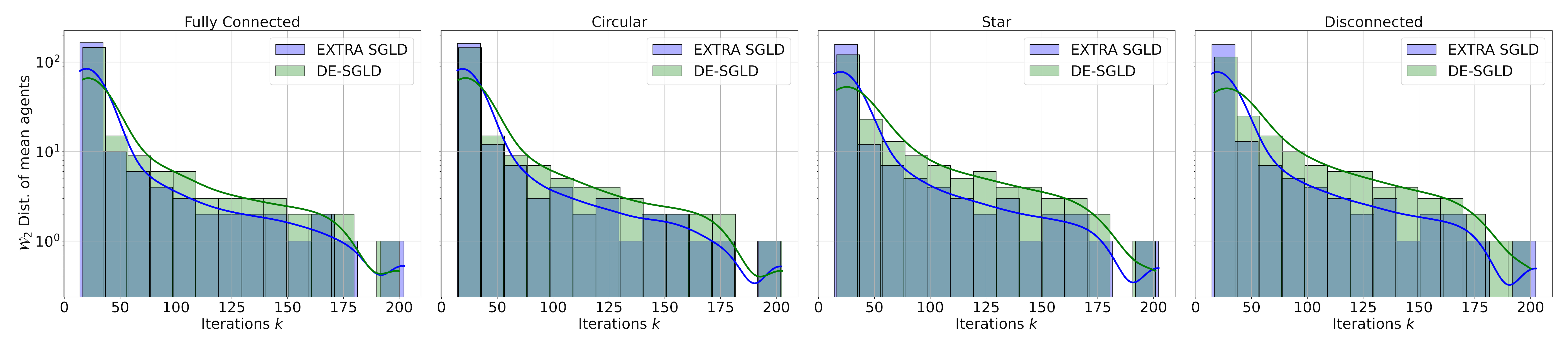}
    \caption{Histogram of the comparative performances of the DE-SGLD and EXTRA SGLD for Bayesian linear regression on four different network structures.} 
    \label{fig:fig4}
\end{figure}\hfill\\

In the comparative analysis of EXTRA SGLD and DE-SGLD, the performance is evaluated across various network structures, as shown in Figures~\ref{fig:fig3} and \ref{fig:fig4}. Figure~\ref{fig:fig3} illustrates the $\mathcal{W}_2$ distance of the mean agents over $200$ iterations for fully connected, circular, star, and disconnected networks, revealing that EXTRA SGLD consistently achieves faster and more stable convergence than DE-SGLD for any $h\in (0,1/2]$. Figure~\ref{fig:fig4}, which depicts histograms of distance distributions on a logarithmic scale at specific iterations, shows that EXTRA SGLD achieves a more concentrated distribution near zero, indicating better convergence. Asymptotically, both algorithms stabilize, but EXTRA SGLD reaches a smaller quantity in 
$\mathcal{W}_2$ distance, highlighting its efficiency in attaining consensus, especially in the more challenging disconnected setting.

\subsection{Bayesian logistic regression with synthetic data}
To test the performance of our algorithm, we first implement the Bayesian logistic regression on synthetic data. Ideally, we have a dataset $Z=\{z_j\}_{j=1}^n$ where $z_j=(X_j,y_j), X_j\in \mathbb{R}^d$ are the features and $y_j\in \{0,1\}$ are the labels with the assumption that $X_j$ are independent and the probability distribution of $y_j$ given $X_j$ and regression coefficients $\beta\in \mathbb R^d$ is given by 
\begin{equation}
    \mathbb{P}(y_j=1|X_j, \beta)=\frac{1}{1+e^{-\beta^TX_j}}.
    \label{eq:n2}
\end{equation}
The prior distribution $p(\beta)\sim \mathcal{N}(0,\lambda I_{3})$ for some $\lambda>0$, where $I_3$ is the $3\times 3$ identity matrix \cite{chatterji2018theory,dubey2016variance,zou2018subsampled}. In a distributed network system, if each agent $i$ contains a subset $Z_i$ of data, then the goal of the Bayesian logistic regression is to sample from $\pi(\beta)\propto e^{-f(\beta)}$ with $f(\beta)=\sum_{i=1}^{N}f_i(\beta)$ where
\begin{equation}
    f_i(\beta):=-\sum_{j=1}^{n_i}\log{p \left(y_j^i=1 |X_j^i,\beta\right)}-\frac{1}{N}\log{p(\beta)}=\sum_{j=1}^{n_i}\log{\left(1+e^{-\beta^TX_j}\right)}+\frac{1}{2N\lambda}\|\beta\|^2
    \label{eq:n3}
\end{equation}
is strongly convex and smooth. We generate the synthetic data from the following model
\begin{equation*}
    X_j\sim \mathcal{N}(0,20I_{3}),\hspace{4mm} p_j\sim \mathcal{U}(0,1), \hspace{4mm} y_j=\begin{cases}
        1 &\text{ if } p_j\le \frac{1}{1+e^{-\beta^TX_j}},\\
        0 & \text{ otherwise}, 
    \end{cases}
\end{equation*}
where $\mathbf{\mathcal{U}}(0,1)$ is the uniform distribution on $[0,1]$, $\beta=[\beta_1,\beta_2,\beta_3]^T\in \mathbb{R}^3$ and the prior distribution $\beta\sim \mathcal{N}(0,\lambda I_{3})$, where $I_3$ is the $3\times 3$ identity matrix. For this experiment, we take $\lambda=10$ just like the linear regression, but we limit the number of nodes to $N=6$ and distribute the data points equally to each node. For Bayesian logistic regression, we consider 10 linearly spaced $h$ values and tune the parameter to the network style. For the fully connected network, we take $h=0.111$, circular network $h=0.056$, star network $h=0.001$ 
and for the disconnected network $h=0.445$. For each node $i$, we calculate their accuracy over $n=1000$ data points and 20 runs with batch size $b=32$ and step size $\eta=0.005$. However, unlike Bayesian linear regression, Bayesian logistic regression does not have a closed-form solution for the posterior distribution $\pi(\beta)$. Therefore, in order to compute $\mathcal{W}_2$ distance between the prior and posterior distributions for Bayesian logistic regression, one may need to run the algorithm over many iterations which is not practical. For these shortcomings, we apply a different technique to measure the performance of the algorithm which is the distribution of the accuracy over the whole data set. This accuracy measure is defined as the ratio of the correctly predicted labels over the whole data set. Since our experiments are identical except for the $h$ parameters to that of the DE-SGLD. 
\begin{figure}[th]
    \centering
    \includegraphics[scale=0.18]{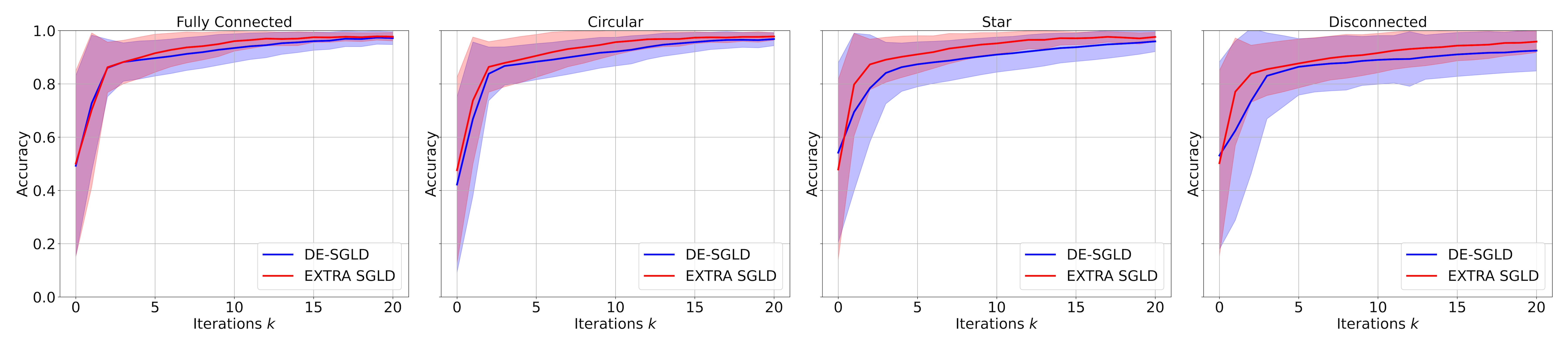}
    \caption{Accuracy distribution of the EXTRA SGLD method across different network structures at a randomly selected node.} 
    \label{fig:fig5}
\end{figure}\hfill\\

Figure~\ref{fig:fig5} shows the mean and standard deviation of the accuracy distribution of the agent $i$ using the DE-SGLD and EXTRA SGLD algorithms. It is clearly noticeable from Figure~\ref{fig:fig5} that, from left to right, all three networks provide somewhat better accuracy than the disconnected one. We also see that for any $h\in (0,1/2]$, EXTRA SGLD performs slightly better than DE-SGLD in general in terms of the accuracy distribution irrespective of network structures. 


\subsection{Bayesian logistic regression with real data}
At this point, we implement our algorithms for real data. In this case, we consider the UCI ML Breast Cancer Wisconsin (Diagnostic) data set \cite{misc_breast_cancer_wisconsin_(diagnostic)_17}. The data set contains 569 instances with 30 features which are computed from a digitized image of a fine needle aspirate (FNA) of a breast mass. For this experiment, we keep the other parameters same as the logistic regression with synthetic data except for the EXTRA parameter $h$. In this case, we take $h=0.278,0.389, 0.167,$ and $0.278$ for fully connected network, circular network, star network, and disconnected network, respectively.
\begin{figure}[th]
    \centering
    \includegraphics[scale=0.18]{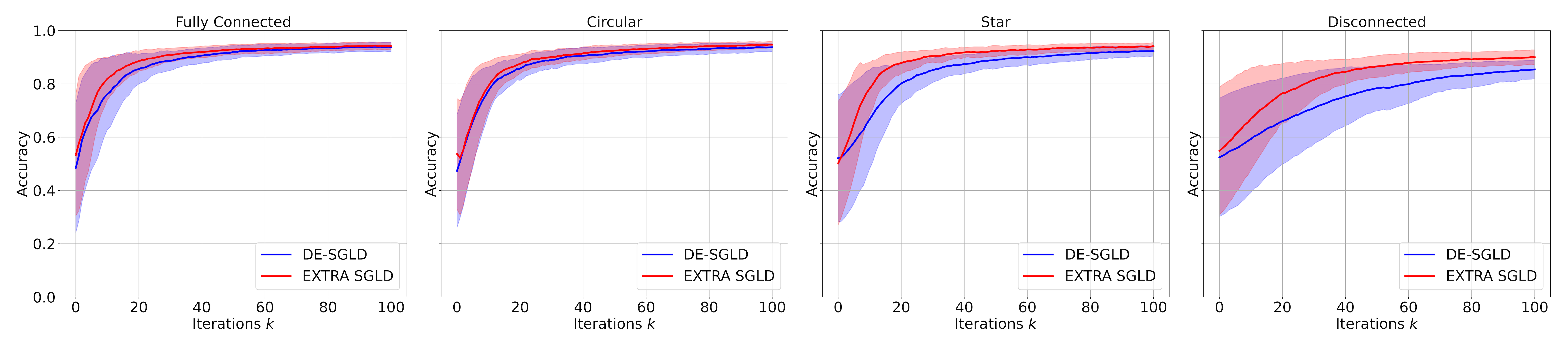}
    \caption{Comparative accuracy distribution of the DE-SGLD and EXTRA SGLD method across different network structures on Breast Cancer data set. The plots are from a randomly selected node.} 
    \label{fig:fig6}
\end{figure}\hfill\\

Figure~\ref{fig:fig6} represents the comparative accuracy distribution of the DE-SGLD and EXTRA SGLD algorithms for Bayesian logistic regression type problems.  Both algorithms exhibit a rapid increase in accuracy during the initial iterations. EXTRA SGLD and DE-SGLD behave similarly in the early stages; but when the number of iterations increases, EXTRA SGLD achieves a higher accuracy.
The shaded regions around the accuracy curves indicate the variance, with EXTRA SGLD demonstrating less variability in general, which implies a more stable performance. As iterations progress, both algorithms converge towards their maximum accuracy. Nonetheless, EXTRA SGLD maintains a slight edge, reaching a higher asymptotic accuracy and showcasing a reliable 
 convergence behavior across fully connected, circular, star, and disconnected network structures.

\section{Conclusion}

Langevin algorithms are widely used Markov Chain Monte Carlo methods in Bayesian learning, particularly for sampling from a parametric model's posterior distribution based on input data and prior parameter distributions. Their stochastic variants, such as stochastic gradient Langevin dynamics (SGLD), facilitate iterative learning using mini-batches from large datasets, making them scalable. However, in scenarios where data are decentralized across a network with communication and privacy restrictions, standard SGLD approaches are unsuitable. To address this, we utilize decentralized SGLD (DE-SGLD) algorithms, which enable collaborative Bayesian learning across a network of agents without sharing individual data points. Despite their advantages, existing DE-SGLD algorithms introduce a bias at each agent that can degrade performance. This bias persists even with full-batch processing and stems from network-related effects. Inspired by the EXTRA algorithm and its generalizations for decentralized optimization, we introduce a generalized EXTRA SGLD that eliminates this bias in full-batch scenarios. Additionally, we demonstrate that, in the mini-batch context, our algorithm offers performance bounds that significantly surpass those of conventional DE-SGLD algorithms. Our empirical results further validate the effectiveness of our approach.

\section*{Acknowledgments}
Mert G\"{u}rb\"{u}zbalaban's research is supported in part by the grants Office of Naval Research Award Numbers
N00014-21-1-2244 and N00014-24-1-2628, National Science Foundation (NSF)
CCF-1814888, NSF DMS-2053485.  
Mohammad Rafiqul Islam is partially supported by the grant NSF DMS-2053454.
Xiaoyu Wang is supported by the Guangzhou-HKUST(GZ) Joint Funding Program (No.2024A03J0630), Guangzhou Municipal Key Laboratory of Financial Technology Cutting-Edge Research.
Lingjiong Zhu is partially supported by the grants NSF DMS-2053454 and DMS-2208303.

\bibliographystyle{alpha}
\bibliography{extra}



\newpage

\appendix

\section{Proofs of the Key Technical Results}\label{sec:proofs:key}

\subsection{Proof of Theorem~\ref{theorem:norm:tilde}}
By substituting the upper bound for $\Vert \widetilde{\mathbf{v}} \Vert_2^{\delta,K}$ to the upper bound for $\Vert \widetilde{\mathbf{x}} \Vert_2^{\delta,K}$ in~\eqref{const:twist}, we get
\begin{align}
\left\Vert \widetilde{\mathbf{x}}\right\Vert_2^{\delta,K} & \leq \eta\gamma_1\left((h/\eta)\gamma_2\left\Vert \widetilde{\mathbf{x}}\right\Vert_2^{\delta,K} + (h/\eta)\frac{w_2(\eta\sigma^2 + 2d)}{N\delta^{2K-2}} \right.
\nonumber 
\\
& \qquad\qquad\qquad \left. + \delta^2(h/\eta)(E_3/\eta)\mathbb{E}\left[ \left\Vert \overline{e}_{x}^{(0)}\right\Vert^2 \right] + \delta^2(E_4/h)\mathbb{E}\left[ \left\Vert \widetilde{v}^{(0)}\right\Vert^2 \right]\right) 
\nonumber 
\\
& \qquad\qquad\qquad\qquad + \eta\frac{w_1(\eta\sigma^2+2d)}{N\delta^{2K-2}} + \delta^2E_1\mathbb{E}\left[ \left\Vert \widetilde{x}^{(0)}\right\Vert^2 \right] + \delta^2E_2\mathbb{E}\left[ \left\Vert \overline{e}_{x}^{(0)}\right\Vert^2 \right].
\end{align}
Under the assumption $h < \frac{1}{\gamma_1\gamma_2}$, where $h$ is defined in \eqref{choice:tilde:W}, we have $h \gamma_1\gamma_2<1$ and the constants $\gamma_1, \gamma_2$ are constants independent of $\eta$ and $\delta$ in~\eqref{defn:gamma:1:2}. We can compute that
\begin{align}
\left\Vert \widetilde{\mathbf{x}}\right\Vert_2^{\delta,K} 
& \leq \frac{1}{1 - h\gamma_1\gamma_2}\Bigg\{\eta^2 \cdot \frac{\left(w_2\gamma_1(h/\eta) + w_1\right)\sigma^2}{N\delta^{2K-2}} + \eta \cdot \left[\frac{2d\left(w_2\gamma_1(h/\eta) + w_1\right)}{N\delta^{2K-2}} \right.
\nonumber 
\\
& \qquad\qquad\qquad\qquad\qquad\qquad \left. + \gamma_1\delta^2\left((h/\eta)(E_3/\eta)\mathbb{E}\left[ \left\Vert \overline{e}_{x}^{(0)}\right\Vert^2 \right] + (E_4/h)\mathbb{E}\left[ \left\Vert \widetilde{v}^{(0)}\right\Vert^2 \right] \right) \right]
\nonumber 
\\
& \qquad\qquad\qquad\qquad + \delta^2\left(E_1\mathbb{E}\left[ \left\Vert \widetilde{x}^{(0)}\right\Vert^2 \right] + E_2\mathbb{E}\left[ \left\Vert \overline{e}_{x}^{(0)}\right\Vert^2 \right] \right)
\Bigg\}
\nonumber 
\\
& \quad := \frac{\eta^2}{\delta^{2K-2}} \cdot \frac{\left(w_2\gamma_1(h/\eta) + w_1\right)\sigma^2/N}{1 - h\gamma_1\gamma_2} + \frac{\eta}{\delta^{2K-2}} \cdot  \Bigg[\frac{2d\left(w_2\gamma_1(h/\eta) + w_1\right)/N}{1 - h\gamma_1\gamma_2} 
\nonumber 
\\
& \qquad\qquad\qquad + \frac{\gamma_1}{1-h\gamma_1\gamma_2}\delta^{2K}\left((h/\eta)(E_3/\eta)\mathbb{E}\left[ \left\Vert \overline{e}_{x}^{(0)}\right\Vert^2 \right] + (E_4/h)\mathbb{E}\left[ \left\Vert \widetilde{v}^{(0)}\right\Vert^2 \right] \right)\Bigg] + \delta^2D_0,
\label{eqn:tilde:x:delta}
\end{align}
with 
$D_0 := \frac{1}{1-h\gamma_1\gamma_2}\left(E_1\mathbb{E}\left[ \left\Vert \widetilde{x}^{(0)}\right\Vert^2 \right] + E_2\mathbb{E}\left[ \left\Vert \overline{e}_{x}^{(0)}\right\Vert^2 \right]\right)$.

Finally, by Lemma~\ref{lemma:tilde:vx} (or equivalently \eqref{const:twist}) for the bound of $\left\Vert \widetilde{\mathbf{v}}\right\Vert_2^{\delta,K}$ and the bound of $\left\Vert \widetilde{\mathbf{x}}\right\Vert_2^{\delta,K}$ in~\eqref{eqn:tilde:x:delta}, we get
\begin{align}
\left\Vert \widetilde{\mathbf{v}}\right\Vert_2^{\delta,K} 
& \leq (h/\eta)\gamma_2\left\Vert \widetilde{\mathbf{x}}\right\Vert_2^{\delta,K} + (h/\eta)\frac{w_2(\eta\sigma^2 + 2d)}{N\delta^{2K-2}} + \delta^2(h/\eta)(E_3/\eta)\mathbb{E}\left[ \left\Vert \overline{e}_{x}^{(0)}\right\Vert^2 \right] + \frac{\delta^2}{h}E_4\mathbb{E}\left[ \left\Vert \widetilde{v}^{(0)}\right\Vert^2 \right]
\nonumber 
\\
&\leq (h/\eta) \gamma_2 \Bigg[\frac{\eta^2}{\delta^{2K-2}} \cdot \frac{\left(w_2\gamma_1(h/\eta) + w_1\right)\sigma^2/N}{1 - h\gamma_1\gamma_2} + \frac{\eta}{\delta^{2K-2}} \cdot  \Bigg[\frac{2d\left(w_2\gamma_1(h/\eta) + w_1\right)/N}{1 - h\gamma_1\gamma_2} 
\nonumber 
\\
& \qquad\qquad\quad + \frac{\gamma_1}{1-h\gamma_1\gamma_2}\delta^{2K}\left((h/\eta)(E_3/\eta)\mathbb{E}\left[ \left\Vert \overline{e}_{x}^{(0)}\right\Vert^2 \right] + (E_4/h)\mathbb{E}\left[ \left\Vert \widetilde{v}^{(0)}\right\Vert^2 \right] \right)\Bigg] + \delta^2D_0
 \Bigg]
 \nonumber 
 \\
 & \qquad\qquad + (h/\eta)\frac{w_2(\eta\sigma^2 + 2d)}{N\delta^{2K-2}} + \delta^2(h/\eta)(E_3/\eta)\mathbb{E}\left[ \left\Vert \overline{e}_{x}^{(0)}\right\Vert^2 \right] + \delta^2(E_4/h)\mathbb{E}\left[ \left\Vert \widetilde{v}^{(0)}\right\Vert^2 \right]
 \nonumber 
 \\
 &= \frac{h\eta}{\delta^{2K-2}} \cdot \frac{\gamma_2\left(w_2\gamma_1(h/\eta) + w_1\right)\sigma^2/N}{1 - h\gamma_1\gamma_2} + \frac{h}{\delta^{2K-2}}\cdot \Bigg[\frac{2\gamma_2d\left(w_2\gamma_1(h/\eta) + w_1\right)/N}{1 - h\gamma_1\gamma_2} + \frac{w_2\sigma^2}{N}
\nonumber 
\\
& \qquad+ \frac{\gamma_1\gamma_2}{1-h\gamma_1\gamma_2}\delta^{2K}\left((h/\eta)(E_3/\eta)\mathbb{E}\left[ \left\Vert \overline{e}_{x}^{(0)}\right\Vert^2 \right] + (E_4/h)\mathbb{E}\left[ \left\Vert \widetilde{v}^{(0)}\right\Vert^2 \right] \right)\Bigg] + (h/\eta)\delta^2\gamma_2D_0
\nonumber 
\\
& \qquad\qquad + (h/\eta)\frac{2dw_2}{N\delta^{2K-2}} + \delta^2(h/\eta)(E_3/\eta)\mathbb{E}\left[ \left\Vert \overline{e}_{x}^{(0)}\right\Vert^2 \right] +\delta^2(E_4/h)\mathbb{E}\left[ \left\Vert \widetilde{v}^{(0)}\right\Vert^2 \right].
\end{align}
The proof is complete. 


\subsection{Proof of Corollary~\ref{corollary:x2avg}}
By~\eqref{iter1}, we can compute that
\begin{align}
x^{(k+1)} & = \left(\widetilde{W}^{k+1} \otimes I_d\right)x^{(0)}  - \eta \sum_{s=0}^{k}\left(\widetilde{W}^{k-s} \otimes I_d\right)\left(\nabla F\left(x^{(s)}\right)  + v^{(s)}\right) - \eta\xi^{(k)} + \sqrt{2\eta}w^{(k+1)}
\nonumber 
\\
& \qquad\qquad -\eta \sum_{s=0}^{k}\left(\widetilde{W}^{k-s} \otimes I_d\right) \xi^{(s+1)}+\sqrt{2 \eta} \sum_{s=0}^{k}\left(\widetilde{W}^{k-s} \otimes I_d\right) w^{(s+1)}.
\end{align}
Then, by using the definition of $\overline{\mathbf{x}}^{(k)}$ in~\eqref{def:avg:x:v}, we can compute that
\begin{align}
x^{(k+1)}-\overline{\mathbf{x}}^{(k+1)} & = x^{(k+1)}-\frac{1}{N}\left(\left(1_N 1_N^T\right) \otimes I_d\right) x^{(k+1)} \\
& =\left(\widetilde{W}^{k+1} \otimes I_d\right) x^{(0)}-\frac{1}{N}\left(\left(1_N 1_N^T \widetilde{W}^{k+1}\right) \otimes I_d\right) x^{(0)} 
\nonumber 
\\
& \qquad -\eta \sum_{s=0}^{k}\left(\widetilde{W}^{k-s} \otimes I_d\right)\left( \nabla F\left(x^{(s)}\right) + v^{(s)}\right) 
\nonumber 
\\
& \qquad +\eta \sum_{s=0}^{k} \frac{1}{N}\left(\left(1_N 1_N^T \widetilde{W}^{k-s}\right) \otimes I_d\right) \left(\nabla F\left(x^{(s)}\right) + v^{(s)}\right)  
\nonumber 
\\
& \qquad -\eta \sum_{s=0}^{k}\left(\widetilde{W}^{k-s} \otimes I_d\right) \xi^{(s+1)}+\eta \sum_{s=0}^{k} \frac{1}{N}\left(\left(1_N 1_N^T \widetilde{W}^{k-s}\right) \otimes I_d\right) \xi^{(s+1)} 
\nonumber 
\\
& \qquad + \sqrt{2 \eta} \sum_{s=0}^{k}\left(\widetilde{W}^{k-s} \otimes I_d\right) w^{(s+1)}-\sqrt{2 \eta} \sum_{s=0}^{k} \frac{1}{N}\left(\left(1_N 1_N^T \widetilde{W}^{k-s}\right) \otimes I_d\right) w^{(s+1)} .
\end{align}
It follows that
\begin{align}
\left\Vert x^{(k+1)}-\frac{1}{N}\left(\left(1_N 1_N^T\right) \otimes I_d\right) x^{(k+1)}\right\Vert^2 
&\leq 4\left\Vert \left(\left(\widetilde{W}^{k+1}-\frac{1}{N} 1_N 1_N^T\right) \otimes I_d\right) x^{(0)}\right\Vert^2 
\nonumber 
\\
& + 4 \eta^2\left\Vert \sum_{s=0}^{k}\left(\left(\widetilde{W}^{k-s}-\frac{1}{N} 1_N 1_N^T\right) \otimes I_d\right) \left(\nabla F\left(x^{(s)}\right) + v^{(s)}\right)\right\Vert^2 
\nonumber 
\\
& \qquad + 4 \eta^2\left\Vert \sum_{s=0}^{k}\left(\left(\widetilde{W}^{k-s}-\frac{1}{N} 1_N 1_N^T\right) \otimes I_d\right) \xi^{(s+1)}\right\Vert^2 
\nonumber 
\\
& \qquad\qquad + 8 \eta\left\Vert \sum_{s=0}^{k}\left(\left(\widetilde{W}^{k-s}-\frac{1}{N} 1_N 1_N^T\right) \otimes I_d\right) w^{(s+1)}\right\Vert^2, 
\end{align}
where we can further compute 
\begin{align}
4\left\Vert \left(\left(\widetilde{W}^{k+1}-\frac{1}{N} 1_N 1_N^T\right) \otimes I_d\right) x^{(0)}\right\Vert^2
& \leq 4\left\Vert \left(\left(\widetilde{W}^{k+1}-\frac{1}{N} 1_N 1_N^T\right) \otimes I_d\right)\right\Vert^2 \mathbb{E}\left[ \left\Vert x^{(0)}\right\Vert^2 \right]
\nonumber 
\\
& \leq 4\overline{\gamma}_{{\scaleto{\widetilde{W}}{5pt}}}^{2k+2} \mathbb{E}\left[ \left\Vert x^{(0)}\right\Vert^2 \right].
\end{align} 
Then, it follows that
\begin{align}
& 4 \eta^2 \left\Vert\sum_{s=0}^{k}\left(\left(\widetilde{W}^{k-s}-\frac{1}{N} 1_N 1_N^T\right) \otimes I_d\right) \left(\nabla F\left(x^{(s)}\right) + v^{(s)} \right)\right\Vert^2 
\nonumber 
\\
& \quad \leq 4 \eta^2\left(\sum_{s=0}^{k}\left\Vert \widetilde{W}^{k-s}-\frac{1}{N} 1_N 1_N^T\right\Vert \cdot\left\Vert \nabla F\left(x^{(s)}\right) + v^{(s)}\right\Vert\right)^2 
\nonumber 
\\
& \quad = 4 \eta^2\left(\sum_{s=0}^{k} \overline{\gamma}_{{\scaleto{\widetilde{W}}{5pt}}}^{k-s} \cdot\left(\left\Vert\nabla F\left(x^{(s)}\right)\right\Vert + \left\Vert v^{(s)}\right\Vert \right)\right)^2 
\nonumber 
\\
& \quad = 4 \eta^2\left(\sum_{s=0}^{k} \overline{\gamma}_{{\scaleto{\widetilde{W}}{5pt}}}^{k-s}\right)^2\left(\frac{\sum_{s=0}^{k} \overline{\gamma}_{{\scaleto{\widetilde{W}}{5pt}}}^{k-s} \cdot\left(\left\Vert\nabla F\left(x^{(s)}\right)\right\Vert + \left\Vert v^{(s)}\right\Vert \right)}{\sum_{s=0}^{k} \overline{\gamma}_{{\scaleto{\widetilde{W}}{5pt}}}^{k-s}}\right)^2 
\nonumber 
\\
& \quad \leq 8 \eta^2\left(\sum_{s=0}^{k} \overline{\gamma}_{{\scaleto{\widetilde{W}}{5pt}}}^{k-s}\right)^2 \sum_{s=0}^{k} \frac{\overline{\gamma}_{{\scaleto{\widetilde{W}}{5pt}}}^{k-s}}{\sum_{s=0}^{k} \overline{\gamma}_{{\scaleto{\widetilde{W}}{5pt}}}^{k-s}}\left(\left\Vert\nabla F\left(x^{(s)}\right)\right\Vert^2 + \left\Vert v^{(s)} \right\Vert^2 \right).
\end{align}
Therefore, we can obtain from Lemma~\ref{lemma:bound:grad:tilde:v} that
\begin{equation}
4 \eta^2\mathbb{E}\left[\left\Vert\sum_{s=0}^{k}\left(\left(\widetilde{W}^{k-s}-\frac{1}{N} 1_N 1_N^T\right) \otimes I_d\right) \left(\nabla F\left(x^{(s)}\right) + v^{(s)} \right)\right\Vert^2\right] \leq 8\eta^2 \cdot \frac{R_{h} + R'_{h}}{\left(1 - \overline{\gamma}_{{\scaleto{\widetilde{W}}{5pt}}}\right)^2}.
\end{equation}
Finally, we can also compute that:
\begin{align}
& 4 \eta^2 \mathbb{E}\left[\left\Vert \sum_{s=0}^{k}\left(\left(\widetilde{W}^{k-s}-\frac{1}{N} 1_N 1_N^T\right) \otimes I_d\right) \xi^{(s+1)}\right\Vert^2\right] 
\nonumber 
\\
& \qquad\qquad\qquad\qquad + 8 \eta\mathbb{E}\left[\left\Vert \sum_{s=0}^{k}\left(\left(\widetilde{W}^{k-s}-\frac{1}{N} 1_N 1_N^T\right) \otimes I_d\right) w^{(s+1)}\right\Vert^2\right]
\nonumber 
\\
& \quad \leq 4\eta^2\sum_{s=0}^{k}\overline{\gamma}_{{\scaleto{\widetilde{W}}{5pt}}}^{2(k-s)}\mathbb{E}\left\Vert \xi^{(s+1)} \right\Vert^2 + 8 \eta\sum_{s=0}^{k}\overline{\gamma}_{{\scaleto{\widetilde{W}}{5pt}}}^{2(k-s)}\mathbb{E}\left\Vert w^{(s+1)} \right\Vert^2
\nonumber 
\\
& \quad \leq 4\eta^2\sigma^2N\sum_{s=0}^{k}\overline{\gamma}_{{\scaleto{\widetilde{W}}{5pt}}}^{2(k-s)} + 8 \eta dN\sum_{s=0}^{k}\overline{\gamma}_{{\scaleto{\widetilde{W}}{5pt}}}^{2(k-s)}
\nonumber 
\\
& \quad \leq \frac{4\eta^2\sigma^2N}{1 - \overline{\gamma}_{{\scaleto{\widetilde{W}}{5pt}}}^2} + \frac{8 \eta dN}{1 - \overline{\gamma}_{{\scaleto{\widetilde{W}}{5pt}}}^2}.
\end{align}
As a result, for every $k = 1,2,3,\ldots$, we have
\begin{align}
\mathbb{E}\left[\left\Vert x^{(k)}-\frac{1}{N}\left(\left(1_N 1_N^T\right) \otimes I_d\right) x^{(k)}\right\Vert^2\right] 
\leq 4\left(\overline{\gamma}_{{\scaleto{\widetilde{W}}{5pt}}}\right)^{2k} \mathbb{E}\left[ \left\Vert x^{(0)}\right\Vert^2 \right] +  8\eta^2 \frac{R_{h} + R'_{h}}{\left(1 - \overline{\gamma}_{{\scaleto{\widetilde{W}}{5pt}}}\right)^2} + \frac{4\eta^2\sigma^2N}{1 - \overline{\gamma}_{{\scaleto{\widetilde{W}}{5pt}}}^2} + \frac{8 \eta dN}{1 - \overline{\gamma}_{{\scaleto{\widetilde{W}}{5pt}}}^2}.
\end{align}
The proof is complete.


\subsection{Proof of Corollary~\ref{corollary:avg2EulerOverdamped}}

For any given $k \geq 1$, we can compute that 
\begin{align}
\label{error:bound}
\mathbb{E}\left[\left\Vert \hat{\mathcal{E}}_{k+1} \right\Vert^2\right] 
& = \mathbb{E}\left[\left\Vert \frac{1}{N}\left( \sum_{i=1}^N\nabla f_i\left(\overline{x}^{(k)}\right) - \nabla f_i\left(x_i^{(k)}\right) \right) \right\Vert^2\right]
\nonumber 
\\
& \leq \frac{1}{N^2}N L^2 \sum_{i=1}^N \mathbb{E}\left[\left\Vert x_i^{(k)} - \overline{x}^{(k)} \right\Vert^2\right]
\nonumber 
\\ 
& = \frac{L^2}{N}\left(4\left(\overline{\gamma}_{{\scaleto{\widetilde{W}}{5pt}}}\right)^{2k}\mathbb{E}\left[ \left\Vert x^{(0)}\right\Vert^2 \right] +  8\eta^2 \cdot \frac{R_{h} + R'_{h}}{\left(1 - \overline{\gamma}_{{\scaleto{\widetilde{W}}{5pt}}}\right)^2} + \frac{4\eta^2\sigma^2N}{1 - \overline{\gamma}_{{\scaleto{\widetilde{W}}{5pt}}}^2} + \frac{8 \eta dN}{1 - \overline{\gamma}_{{\scaleto{\widetilde{W}}{5pt}}}^2}\right)
\nonumber 
\\
& = \frac{4L^2\left(\overline{\gamma}_{{\scaleto{\widetilde{W}}{5pt}}}\right)^{2k}}{N} \mathbb{E}\left[ \left\Vert x^{(0)}\right\Vert^2 \right] + \eta^2 \cdot \frac{4L^2}{N}\left(\frac{2\left(R_{h} + R'_{h}\right)}{\left(1 - \overline{\gamma}_{{\scaleto{\widetilde{W}}{5pt}}}\right)^2} + \frac{\sigma^2N}{1 - \overline{\gamma}_{{\scaleto{\widetilde{W}}{5pt}}}^2}\right) + \eta\cdot\frac{8L^2d}{1 - \overline{\gamma}_{{\scaleto{\widetilde{W}}{5pt}}}^2},
\end{align}
where the last equality is due to Corollary~\ref{corollary:x2avg}. 
Next, we recall the dynamics:
\begin{equation}
\overline{x}^{(k+1)} - x_{k+1} = \overline{x}^{(k)} - x_{k} - \frac{\eta}{N} \left(\nabla f\left(\overline{x}^{(k)}\right)  - \nabla f\left(x_k\right) \right) + \eta \hat{\mathcal{E}}_{k+1} - \eta\overline{\xi}^{(k)}.
\end{equation}
Under the assumption $\eta < 2/L$ and $L$-smoothness and the $\mu$-convexity of $\frac{1}{N}f$, we can compute that
\begin{align}
\left\Vert \overline{x}^{(k+1)}-x_{k+1}\right\Vert ^2 
&\leq\left(1-2 \eta \mu\left(1-\frac{\eta L}{2}\right)\right)\left\Vert \overline{x}^{(k)}-x_{k}\right\Vert ^2+\eta^2\left\Vert  \hat{\mathcal{E}}_{k+1} -\overline{\xi}^{(k)}\right\Vert ^2 
\nonumber 
\\
& \qquad\qquad +2\left\langle\overline{x}^{(k)}-x_{k}-\eta \frac{1}{N}\left(\nabla f\left(\overline{x}^{(k)}\right)-\nabla f\left(x_{k}\right)\right), \eta  \hat{\mathcal{E}}_{k+1} -\eta \overline{\xi}^{(k)}\right\rangle.\label{take:expectations:in}
\end{align}
Next, we take expectations in \eqref{take:expectations:in} to get
\begin{align}
& \mathbb{E}\left[\left\Vert \overline{x}^{(k+1)} - x_{k+1} \right\Vert^2\right]
\nonumber 
\\
& \quad \leq\left(1-2 \eta \mu\left(1-\frac{\eta L}{2}\right)\right)\mathbb{E}\left[\left\Vert \overline{x}^{(k)}-x_{k}\right\Vert^2\right]+\eta^2\mathbb{E}\left[\left\Vert \hat{\mathcal{E}}_{k+1}\right\Vert^2\right] + \eta^2\mathbb{E}\left[\left\Vert \overline{\xi}^{(k)} \right\Vert^2\right]
\nonumber 
\\
& \qquad\qquad +2\mathbb{E}\left\langle\overline{x}^{(k)}-x_{k}-\eta \frac{1}{N}\left[\nabla f\left(\overline{x}^{(k)}\right)-\nabla f\left(x_{k}\right)\right], \eta \hat{\mathcal{E}}_{k+1}\right\rangle
\nonumber 
\\
& \quad \leq\left(1-2 \eta \mu\left(1-\frac{\eta L}{2}\right)\right)\mathbb{E}\left[\left\Vert \overline{x}^{(k)}-x_{k}\right\Vert^2\right]+\eta^2\mathbb{E}\left[\left\Vert \hat{\mathcal{E}}_{k+1}\right\Vert^2\right] + \eta^2\mathbb{E}\left[\left\Vert \overline{\xi}^{(k)} \right\Vert^2\right]
\nonumber 
\\
& \qquad\qquad +2\mathbb{E}\left[\left(\left\Vert \overline{x}^{(k)}-x_{k} \right\Vert + \eta \frac{1}{N}\left\Vert \nabla f\left(\overline{x}^{(k)}\right)-\nabla f\left(x_{k}\right)\right\Vert \right) \cdot \eta \left\Vert \hat{\mathcal{E}}_{k+1} \right\Vert\right]
\nonumber 
\\
& \quad \leq \left(1-2 \eta \mu\left(1-\frac{\eta L}{2}\right)\right)\mathbb{E}\left[\left\Vert \overline{x}^{(k)}-x_{k}\right\Vert^2\right] + \eta^2\mathbb{E}\left[\left\Vert \hat{\mathcal{E}}_{k+1}\right\Vert^2\right] + \eta^2\mathbb{E}\left[\left\Vert \overline{\xi}^{(k)} \right\Vert^2\right]
\nonumber 
\\
& \qquad\qquad +2\left(1 + \eta\frac{L}{N} \right)\eta\mathbb{E}\left[\left\Vert \overline{x}^{(k)}-x_{k} \right\Vert \cdot \left\Vert \hat{\mathcal{E}}_{k+1} \right\Vert\right]
\nonumber
\\
& \quad \leq  \left(1-2 \eta \mu\left(1-\frac{\eta L}{2}\right)\right)\mathbb{E}\left[\left\Vert \overline{x}^{(k)}-x_{k}\right\Vert^2\right] + \eta^2\mathbb{E}\left[\left\Vert \hat{\mathcal{E}}_{k+1}\right\Vert^2\right] + \eta^2\mathbb{E}\left[\left\Vert \overline{\xi}^{(k)} \right\Vert^2\right]
\nonumber 
\\
& \qquad\qquad +2\left(1 + \eta L \right)\eta\mathbb{E}\left[\left\Vert \overline{x}^{(k)}-x_{k} \right\Vert \cdot \left\Vert \hat{\mathcal{E}}_{k+1} \right\Vert\right]
\nonumber 
\\
& \quad \leq \left(1- \eta \mu\left(1-\frac{\eta L}{2}\right)\right)\mathbb{E}\left[\left\Vert \overline{x}^{(k)}-x_{k}\right\Vert^2\right] + \eta\left(\eta + \frac{\left(1 + \eta L\right)^2}{\mu\left(1 - \frac{\eta L}{2} \right)}\right)\mathbb{E}\left[\left\Vert \hat{\mathcal{E}}_{k+1}\right\Vert^2\right] + \eta^2\frac{\sigma^2}{N},
\end{align}
where we used the inequality $2xy \leq c'x^2 + \frac{y^2}{c'}$ for any $c'>0$ and $x,y\in\mathbb{R}$ where we took $c' = \frac{\mu\left( 1 - \frac{\eta L}{2}\right)}{1 + \eta L}$, and also the fact that since $\eta < 2/L$, we have $\eta^2 \frac{L}{2} < \eta \mu$ so that $\eta\mu\left( 1 - \frac{\eta L}{2}\right) \in (0,1)$. Now by using the bound in~\eqref{error:bound}, we get
\begin{align}
& \mathbb{E}\left[\left\Vert \overline{x}^{(k+1)} - x_{k+1} \right\Vert^2\right]
\nonumber 
\\
&\leq \left(1- \eta \mu\left(1-\frac{\eta L}{2}\right)\right)\mathbb{E}\left[\left\Vert \overline{x}^{(k)}-x_{k}\right\Vert^2\right] + \eta^2\left(1 + \frac{\left(1 + \eta L\right)^2}{\eta\mu\left(1 - \frac{\eta L}{2} \right)}\right)\mathbb{E}\left[\left\Vert \hat{\mathcal{E}}_{k+1}\right\Vert^2\right] + \eta^2\frac{\sigma^2}{N}
\nonumber 
\\
& = \left(1- \eta \mu\left(1-\frac{\eta L}{2}\right)\right)\mathbb{E}\left[\left\Vert \overline{x}^{(k)}-x_{k}\right\Vert^2\right] 
\nonumber
\\
&\qquad
+ \eta \left(\eta + \frac{\left(1 + \eta L\right)^2}{\mu\left(1 - \frac{\eta L}{2} \right)}\right)\Bigg[\frac{4L^2\left(\overline{\gamma}_{{\scaleto{\widetilde{W}}{5pt}}}\right)^{2k}}{N} \mathbb{E}\left[ \left\Vert x^{(0)}\right\Vert^2 \right] + \eta^2 \cdot \frac{4L^2}{N}\left(\frac{2\left(R_{h} + R'_{h}\right)}{\left(1 - \overline{\gamma}_{{\scaleto{\widetilde{W}}{5pt}}}\right)^2} + \frac{\sigma^2N}{1 - \overline{\gamma}_{{\scaleto{\widetilde{W}}{5pt}}}^2}\right) 
\nonumber 
\\
& \qquad\qquad\qquad\qquad\qquad\qquad\qquad\qquad\qquad\qquad   
+ \eta\cdot\frac{8L^2d}{1 - \overline{\gamma}_{{\scaleto{\widetilde{W}}{5pt}}}^2}
\Bigg] + \eta^2\frac{\sigma^2}{N}.
\end{align}
Since $x^{(0)} = x_0$, we finally get
\begin{align}
& \mathbb{E}\left[\left\Vert \overline{x}^{(k)} - x_{k} \right\Vert^2\right]
\nonumber 
\\
& \quad \leq  \sum_{i=0}^{k-1}\left(1- \eta \mu\left(1-\frac{\eta L}{2}\right)\right)^{i}
\nonumber 
\\
& \qquad\qquad \cdot \left(\eta\left(\eta + \frac{\left(1 + \eta L\right)^2}{\mu\left(1 - \frac{\eta L}{2}\right)}\right)
\left(\eta^2 \cdot \frac{4L^2}{N}\left(\frac{2\left(R_{h} + R'_{h}\right)}{\left(1 - \overline{\gamma}_{{\scaleto{\widetilde{W}}{5pt}}}\right)^2} + \frac{\sigma^2N}{1 - \overline{\gamma}_{{\scaleto{\widetilde{W}}{5pt}}}^2}\right) 
+ \eta\cdot\frac{8L^2d}{1 - \overline{\gamma}_{{\scaleto{\widetilde{W}}{5pt}}}^2}
\right) + \eta^2\frac{\sigma^2}{N} \right)
\nonumber 
\\
& \qquad + \sum_{i=0}^{k-1}\left(1- \eta \mu\left(1-\frac{\eta L}{2}\right)\right)^{i}\eta\left(\eta + \frac{\left(1 + \eta L\right)^2}{\mu\left(1 - \frac{\eta L}{2}\right)}\right)\frac{4L^2\left(\overline{\gamma}_{{\scaleto{\widetilde{W}}{5pt}}}\right)^{2(k-i)}}{N} \mathbb{E}\left[ \left\Vert x^{(0)}\right\Vert^2 \right]
\nonumber 
\\
& \quad =\frac{1-\left(1-\eta \mu\left(1-\frac{\eta L}{2}\right)\right)^k}{1-\left(1-\eta \mu\left(1-\frac{\eta L}{2}\right)\right)} 
\nonumber 
\\
& \qquad\qquad \cdot\left(\eta\left(\eta + \frac{\left(1 + \eta L\right)^2}{\mu\left(1 - \frac{\eta L}{2}\right)}\right)
\left(\eta^2 \cdot \frac{4L^2}{N}\left(\frac{2\left(R_{h} + R'_{h}\right)}{\left(1 - \overline{\gamma}_{{\scaleto{\widetilde{W}}{5pt}}}\right)^2} + \frac{\sigma^2N}{1 - \overline{\gamma}_{{\scaleto{\widetilde{W}}{5pt}}}^2}\right) 
+ \eta\cdot\frac{8L^2d}{1 - \overline{\gamma}_{{\scaleto{\widetilde{W}}{5pt}}}^2}
\right) + \eta^2\frac{\sigma^2}{N} \right)
\nonumber 
 \\
& \qquad+\frac{\overline{\gamma}_{{\scaleto{\widetilde{W}}{5pt}}}^{2 k}-\left(1-\eta \mu\left(1-\frac{\eta L}{2}\right)\right)^k}{1-\left(1-\eta \mu\left(1-\frac{\eta L}{2}\right)\right)(\overline{\gamma}_{{\scaleto{\widetilde{W}}{5pt}}})^{-2}} \frac{4 L^2}{N} \mathbb{E}\left\|x^{(0)}\right\|^2
\nonumber 
\\
& \quad \leq \frac{\eta\left(\eta + \frac{\left(1 + \eta L\right)^2}{\mu\left(1 - \frac{\eta L}{2}\right)}\right)
\left(\eta^2 \cdot \frac{4L^2}{N}\left(\frac{2\left(R_{h} + R'_{h}\right)}{\left(1 - \overline{\gamma}_{{\scaleto{\widetilde{W}}{5pt}}}\right)^2} + \frac{\sigma^2N}{1 - \overline{\gamma}_{{\scaleto{\widetilde{W}}{5pt}}}^2}\right) 
+ \eta\cdot\frac{8L^2d}{1 - \overline{\gamma}_{{\scaleto{\widetilde{W}}{5pt}}}^2}
\right) + \eta^2\frac{\sigma^2}{N}}{\eta\mu\left(1 - \frac{\eta L}{2} \right)}
\nonumber 
\\
& \qquad\qquad +\frac{\overline{\gamma}_{{\scaleto{\widetilde{W}}{5pt}}}^{2 k}-\left(1-\eta \mu\left(1-\frac{\eta L}{2}\right)\right)^k}{(\overline{\gamma}_{{\scaleto{\widetilde{W}}{5pt}}})^{2}-1+\eta \mu\left(1-\frac{\eta L}{2}\right)} \frac{4 L^2(\overline{\gamma}_{{\scaleto{\widetilde{W}}{5pt}}})^{2}}{N} \mathbb{E}\left\|x^{(0)}\right\|^2.
\end{align}
The proof is complete.

\section{Proofs of Technical Lemmas} 
\label{sec:proofs:lemmas}

\subsection{Proof of Lemma~\ref{lemma:inf:seq}}
We can compute that
\begin{align}
& \max_{k = 0,\ldots,K-1}\mathbb{E}\left[\left(\frac{1}{\delta^{k+1}}\left\Vert a^{(k+1)} \right\Vert\right)^2\right]
\nonumber 
\\ 
& \qquad = \frac{1}{\delta^2}\max_{k = 0,\ldots,K-1}\mathbb{E}\left[\left(\frac{1}{\delta^{k}}\left\Vert a^{(k+1)} \right\Vert\right)^2\right]
\nonumber 
\\
& \qquad \leq \frac{c_1}{\delta^2}\max_{k = 0,\ldots,K-1}\mathbb{E}\left[\left(\frac{1}{\delta^{k}}\left\Vert a^{(k)} \right\Vert\right)^2\right] + \frac{c_2}{\delta^2}\max_{k = 0,\ldots,K-1}\mathbb{E}\left[\left(\frac{1}{\delta^{k}}\left\Vert b^{(k)} \right\Vert\right)^2\right] + \frac{c_0}{\delta^{2K}}
\label{max:ineq:t1}
\\
& \qquad = \frac{c_1}{\delta^2} \Vert \mathbf{a} \Vert_2^{\delta, K} + \frac{c_2}{\delta^2}\Vert \mathbf{b} \Vert_2^{\delta, K} + \frac{c_0}{\delta^{2K}},
\nonumber 
\end{align}
where we used the simple inequality for maximum, that is, $\max_{k}(x_{k}+y_{k}) \leq \max_{k}(x_{k}) + \max_{k}(y_{k})$
for any real sequences $(x_{k}),(y_{k})$ to get~\eqref{max:ineq:t1} by using \eqref{lemma:6:assumption}. Therefore, for any $\delta \in (0,1)$, we have
\begin{align}
\max_{k = 0,\ldots,K}\mathbb{E}\left[\frac{1}{\delta^{k}}\left\Vert a^{(k)} \right\Vert^2\right] 
& = \max_{k = -1,\ldots,K-1}\mathbb{E}\left[\frac{1}{\delta^{k+1}}\left\Vert a^{(k+1)} \right\Vert^2\right] 
\nonumber 
\\
& \leq \max_{k = 0,\ldots,K-1}\mathbb{E}\left[\frac{1}{\delta^{k+1}}\left\Vert a^{(k+1)} \right\Vert^2\right] + \mathbb{E}\left[\left\Vert a^{(0)} \right\Vert^2\right]
\nonumber 
\\
& \leq \frac{c_1}{\delta^2} \Vert \mathbf{a} \Vert_2^{\delta, K} + \frac{c_2}{\delta^2}\Vert \mathbf{b} \Vert_2^{\delta, K} + \frac{c_0}{\delta^{2K}} + \mathbb{E}\left[\left\Vert a^{(0)} \right\Vert^2\right].
\end{align}
The proof is complete.

    
\subsection{Proof of Lemma~\ref{lemma:seq}}

The proof follows from an adaption of the proof of 
Lemma~\ref{lemma:inf:seq}
by using the assumption that $\mathbb{E}\left\Vert a^{(k+1)} \right\Vert^2$ is bounded by the sum of finite components and applying the inequality that $\max_k(x_{1_k} + x_{2_k} + \ldots + x_{n_k} ) \leq \max_k(x_{1_k}) + \max_k(x_{2_k}) + \ldots + \max_k(x_{n_k})$ for any real sequences $(x_{i_k})$, $i = 1,2,\ldots,n$.


\subsection{Proof of Lemma~\ref{lemma:err}}

Before proving the lemma, we give a preliminary result. With the fact in~\eqref{eqn:opt:null}, Lemma~3.1 in~\cite{shi2015extra} shows first-order optimality condition of EXTRA algorithm in decentralized optimization. Given mixing matrices $W$ and $\widetilde{W}$, define $U:=\widetilde{W}-W$ by letting $U^{1/2} := P D^{1 / 2} P^{\mathrm{T}} \in \mathbb{R}^{N \times N}$. Under Assumptions~\ref{assumption:f} and~\ref{assumption:mixing}, then $\mathbf{x}_*$ is consensual if and only if there exists $\mathbf{q}_*=\mathcal{U} \mathbf{p}$ for some $\mathbf{p} \in \mathbb{R}^{Nd}$ where $\mathcal{U} := U \otimes I_d$ such that
\begin{equation}
\label{eqn:opt:cond}
\begin{cases}
\mathcal{U}^{1/2} \mathbf{q}_*+\eta \nabla F\left(\mathbf{x}_*\right)=\mathbf{0},
\\
\mathcal{U}^{1/2} \mathbf{x}_*=\mathbf{0}.
\end{cases}
\end{equation}
According to Assumption~\ref{assumption:mixing} and by decomposing $U^{1/2} = PD^{1/2}P^T$, we get
\begin{equation}
\label{eqn:opt:null}
\nul\left\{U^{1/2}\right\} = \nul\left\{P^T\right\} = \nul\left\{\widetilde W - W\right\} = \operatorname{span}\{1_{N}\},
\end{equation}
where $\operatorname{span}\{1_N\}$ is the span of the vector space supported by all-one vectors $\left[1^T_N,1^T_N,\ldots,1^T_N\right]$,
and it implies $U^{1/2}$ is symmetric and $1_{N}^TU^{1/2} = 0$. 

Now we deliver the proof as the following. We have the error such that $e_x^{(k)} = x^{(k)} - \mathbf{x}_*$, where $\mathbf{x}_* = \left[x_*^T,\ldots x_*^T\right]^T$ is consensual.
Thus, we can compute that
\begin{equation}
e_x^{(k)} = x^{(k)} - \overline{\mathbf{x}}^{(k)} + 1_N \otimes\left(\overline{x}^{(k)} - x_* \right) = \widetilde{x}^{(k)} + 1_N \otimes \overline{e}_x^{(k)},
\end{equation}
where $\overline{x}^{(k)}$ is the mean of $x^{(k)}$ in~\eqref{def:avg:x:v}, and we used the definition of $\widetilde{x}^{(k)}$ in~\eqref{def:tildex:tildev}, and the definition of $\overline{e}_x^{(k)}$ in~\eqref{def:avi:err} to obtain the last equality above.

Next, we notice that $e_v^{(k)} = v^{(k)} + \nabla F\left(\mathbf{x}_*\right)$. 
By Lemma~\ref{eqn:opt:cond}, we have
\begin{equation}
\mathbf{v}_* = \frac{1}{\eta}\mathcal{U}^{1/2}\mathbf{q}_* = -\nabla F(\mathbf{x}_*),
\end{equation}
where we have $\mathbf{v}_* = -\nabla F(\mathbf{x}_*) = \left(\nabla f_1\left(x_*\right), \ldots, \nabla f_N\left(x_*\right)\right)^T$, which is also consensual, and similarly, we can compute 
\begin{equation}\label{e:v:k}
e_v^{(k)} = v^{(k)} - \overline{\mathbf{v}}^{(k)} + 1_N \otimes\left(\overline{v}^{(k)} - v_* \right) = \widetilde{v}^{(k)} + 1_N \otimes \overline{e}_x^{(k)},
\end{equation}
where $\overline{v}^{(k)}$ is the mean of $v^{(k)}$ in~\eqref{def:avg:x:v} and we used the definition of $\widetilde{v}^{(k)}$ in~\eqref{def:tildex:tildev}, and the definition of $\overline{e}_v^{(k)}$ in~\eqref{def:avi:err} to obtain the last equality in \eqref{e:v:k}. Moreover, by~\eqref{iter2}, we can compute
\begin{equation}
v_i^{(k+1)} =  v_i^{(k)} - \sum_{j \in \Omega_i}U_{ij}\left(v^{(k)} + \nabla F\left(x^{(k)}\right) - \mathcal{B}x^{(k)} \right)_{j} - \sum_{j \in \Omega_i}U_{ij}\xi_j^{(k)} + \sum_{j \in \Omega_i}U_{ij}\sqrt{\frac{2}{\eta}}w_j^{(k+1)}.
\end{equation}
Then by the definition, $U = \widetilde W - W$ with doubly stochastic matrices $\widetilde{W},W$ such that $\nul \left\{W-\widetilde{W}\right\}=\operatorname{span}\{1_N\}$ under Assumption~\ref{assumption:mixing}. Therefore, we have
\begin{equation}
\overline{v}^{(k+1)} = \overline{v}^{(k)} = \cdots = \overline{v}^{(0)} = 0,
\end{equation}
where $v^{(0)} = \frac{1}{\eta}\mathcal{U}^{1/2}q^{(0)} = 0$ from~\eqref{def:v}. Noticing $\frac{1}{N}\sum_{i=1}^N \nabla f_i\left(x_*\right) = 0$ under optimality condition, we can get from~\eqref{def:avi:err} that 
\begin{equation}
\overline{e}_v^{(k)} = \overline{v}^{(k)} = 0, \quad e_v^{(k)} = \widetilde{v}^{(k)}.
\end{equation}
The proof is complete.


\subsection{Proof of Lemma~\ref{lemma:avi:ex}}
From Lemma~\ref{lemma:err}, $\overline{v}^{(k)} = 0$, such that we can compute the average iterates $\overline{x}^{(k)}$ from \eqref{iter1} as follows
\begin{equation}
\overline{x}^{k+1} - x_* = \overline{x}^{(k)} - x_* - \frac{\eta}{N}\left(\sum_{i=1}^N\nabla f_i\left(\overline{x}^{(k)}\right) - \mathcal{E}^{(k)}\right) - \eta\overline{\xi}^{(k)} + \sqrt{2\eta}\overline{w}^{(k+1)},
\end{equation}
where we recall that
\begin{equation}
\label{grad:err}
\mathcal{E}^{(k)} := \sum_{i=1}^N\left(\nabla f_i\left(\overline{x}^{(k)}\right)-\nabla f_i\left(x_i^{(k)}\right)\right).
\end{equation}
Now by the optimality, we have $1^T \nabla f(\mathbf{x}_*) = \sum_{i=1}^N \nabla f_i\left(x_*\right) = 0$ and $\overline{e}_x^{(k)} = \frac{1}{N}\sum_{i=1}^N\left(x_i^{(k)} - x_*\right) = \overline{x}^{(k)} - x_* \in \mathbb{R}^d$, we can compute 
\begin{align}
\left\Vert \overline{e}_x^{(k+1)} \right\Vert^2 & = \left\Vert \overline{e}_x^{(k)} - \frac{\eta}{N} \sum_{i=1}^N\left(\nabla f_i\left(\overline{x}^{(k)}\right) - \nabla f_i\left(x_*\right)\right)\right\Vert^2 + \left\Vert\frac{\eta}{N} \mathcal{E}^{(k)} - \eta \overline{\xi}^{(k)} + \sqrt{2\eta} \overline{w}^{(k+1)} \right\Vert^2
\nonumber 
\\
& \qquad\qquad + 2\left\langle \overline{e}_x^{(k)} - \frac{\eta}{N}\sum_{i=1}^N\left(\nabla f_i\left(\overline{x}^{(k)}\right) - \nabla f_i\left(x_*\right)\right) \,,\, \frac{\eta}{N} \mathcal{E}^{(k)} - \eta \overline{\xi}^{(k)} + \sqrt{2\eta} \overline{w}^{(k+1)}  \right\rangle.
\label{three:terms:to:compute}
\end{align}
Next, we compute the first term on the right hand side of \eqref{three:terms:to:compute} as follows
\begin{align}
& \left\Vert \overline{e}_x^{(k)} - \frac{\eta}{N}\sum_{i=1}^N\left(\nabla f_i\left(\overline{x}^{(k)}\right) - \nabla f_i\left(x_*\right)\right) \right\Vert^2 
\nonumber 
\\
& \quad = \left\Vert \overline{e}_x^{(k)} \right\Vert^2 + \eta^2\sum_{i=1}^N\left\Vert \frac{1}{N}\left(\nabla f_i\left(\overline{x}^{(k)}\right) - \nabla f_i\left(x_*\right)\right)  \right\Vert^2 
\nonumber
\\
&\qquad\qquad\qquad\qquad\qquad\qquad
- 2\eta \left\langle \overline{e}_x^{(k)} \,,\, \frac{1}{N}\sum_{i=1}^N\left(\nabla f_i\left(\overline{x}^{(k)}\right) - \nabla f_i\left(x_*\right)\right) \right\rangle
\nonumber 
\\
& \quad \leq \left\Vert \overline{e}_x^{(k)} \right\Vert^2 - \left(2\eta - \eta^2L\right)\left\langle \overline{e}_x^{(k)} \,,\, \frac{1}{N}\sum_{i=1}^N\left(\nabla f_i\left(\overline{x}^{(k)}\right) - \nabla f_i\left(x_*\right)\right)  \right\rangle 
\nonumber 
\\
& \quad \leq \left(1 - 2\eta\mu\left(1 - \frac{\eta L}{2} \right) \right)\left\Vert \overline{e}_x^{(k)} \right\Vert^2,
\end{align}
where we used the condition $\eta < 2/L$, and used the assumption on $L$-smoothness of $f_i$ to obtain the first term in the inequality of the second line above, and the assumption on $\mu$-strongly convexity of $f_i$ to get the inequality of the last line. 

By taking the expectations in \eqref{three:terms:to:compute}, since $\overline{\xi}^{(k)}$ and $\overline{w}^{(k)}$ have mean zero conditional on the natural filtration, and they are independent to $\mathcal{E}^{(k)}$, we can compute that
\begin{align}
& \mathbb{E}\left[\left\Vert \overline{e}_x^{(k+1)} \right\Vert^2\right] 
\nonumber 
\\
& \leq \left(1 - 2\eta\mu\left(1 - \frac{\eta L}{2} \right) \right)\mathbb{E}\left[\left\Vert \overline{e}_x^{(k)} \right\Vert^2\right] + \frac{\eta^2}{N^2}\mathbb{E}\left[\left\Vert \mathcal{E}^{(k)} \right\Vert^2\right] + \eta^2 \mathbb{E}\left[\left\Vert \overline{\xi}^{(k)}\right\Vert^2\right] + 2\eta\mathbb{E}\left[\left\Vert \overline{w}^{(k+1)}\right\Vert^2\right]  
\nonumber 
\\
& \qquad + \frac{2\eta}{N}\mathbb{E}\left\langle \overline{e}_x^{(k)} \,,\,  \mathcal{E}^{(k)} \right\rangle +  \frac{2\eta^2}{N^2}\mathbb{E}\left\langle \sum_{i=1}^N\left(\nabla f_i\left(x_*\right) - \nabla f_i\left(\overline{x}^{(k)}\right) \right) \,,\,  \mathcal{E}^{(k)} \right\rangle
\nonumber 
\\
&  \leq \left(1 - 2\eta\mu\left(1 - \frac{\eta L}{2} \right) \right)\mathbb{E}\left[\left\Vert \overline{e}_x^{(k)} \right\Vert^2\right] + \frac{\eta^2}{N^2}\mathbb{E}\left[\left\Vert \mathcal{E}^{(k)} \right\Vert^2\right] + \eta^2\frac{\sigma^2}{N} + 2\eta \frac{d}{N} 
\nonumber 
\\
& \qquad\qquad + \frac{2\eta}{N}\mathbb{E}\left[\left\Vert \overline{e}_x^{(k)} \right\Vert \cdot \left\Vert \mathcal{E}^{(k)} \right\Vert \right] + \frac{2\eta^2}{N^2}\mathbb{E}\left[ \left\Vert\sum_{i=1}^N\left(\nabla f_i\left(x_*\right) - \nabla f_i\left(\overline{x}^{(k)}\right) \right) \right\Vert \cdot \left\Vert\mathcal{E}^{(k)}\right\Vert \right]
\label{avgx:t1}
\\
& \leq \left(1 - 2\eta\mu\left(1 - \frac{\eta L}{2} \right) \right)\mathbb{E}\left[\left\Vert \overline{e}_x^{(k)} \right\Vert^2\right] + \frac{\eta^2}{N^2}\mathbb{E}\left[\left\Vert \mathcal{E}^{(k)} \right\Vert^2\right] + \eta^2\frac{\sigma^2}{N} + 2\eta \frac{d}{N} 
\nonumber 
\\
& \qquad\qquad + \frac{2\eta}{N}\left(1 + \eta L\right)\mathbb{E}\left[\left\Vert \overline{e}_x^{(k)} \right\Vert \cdot \left\Vert \mathcal{E}^{(k)} \right\Vert \right] 
\nonumber 
\\
& \leq \left(1 - \eta\mu\left(1 - \frac{\eta L}{2} \right) \right)\mathbb{E}\left[\left\Vert \overline{e}_x^{(k)} \right\Vert^2\right] 
 + \frac{\eta}{N}\left(\frac{\eta}{N} + \frac{1+\eta L}{N\mu\left(1 - \frac{\eta L}{2} \right)} \right)\mathbb{E}\left[\left\Vert \mathcal{E}^{(k)} \right\Vert^2\right] + \eta^2\frac{\sigma^2}{N} + 2\eta \frac{d}{N} ,
\label{avgx:t2}
\end{align}
where we used $1 - \eta\mu\left(1 - \frac{\eta L}{2} \right) \in (0,1)$. Note that we used Cauchy-Schwarz inequality to get~\eqref{avgx:t1} and used the inequality $2xy \leq \frac{x^2}{c'} + c'y^2$ for $c>0$ to get~\eqref{avgx:t2} by taking $c' = \eta\mu\left(1 - \frac{\eta L}{2} \right)$. Moreover, we can compute from~\eqref{grad:err} to get
\begin{equation}
\frac{\eta}{N}\left\Vert \mathcal{E}^{(k)} \right\Vert^2 \leq \frac{\eta}{N}\sum_{i=1}^N\left\Vert \nabla f_i\left(\overline{x}^{(k)}\right)-\nabla f_i\left(x_i^{(k)}\right) \right\Vert^2 \leq \frac{\eta L^2}{N} \left\Vert \widetilde{x}^{(k)}\right\Vert^2,
\end{equation}
where we used $\sum_{i=1}^N\left\Vert \overline{x}^{(k)} - x_i^{(k)} \right\Vert^2 = \sum_{i=1}^N \left\Vert \widetilde{x}^{(k)}_i \right\Vert^2 = \left\Vert \widetilde{x} \right\Vert^2$. Hence, we get
\begin{align}
\label{avgx:t3}
\mathbb{E}\left[\left\Vert \overline{e}_x^{(k+1)} \right\Vert^2\right] & \leq \left(1 - \eta\mu\left(1 - \frac{\eta L}{2} \right) \right)\mathbb{E}\left[\left\Vert \overline{e}_x^{(k)} \right\Vert^2 \right]
\nonumber 
\\
& \qquad\qquad + \frac{\eta L^2}{N}\left(\frac{\eta}{N} + \frac{1+\eta L}{N\mu\left(1 - \frac{\eta L}{2} \right)} \right)\mathbb{E}\left[\left\Vert \widetilde{x}^{(k)}\right\Vert^2\right] + \eta^2\frac{\sigma^2}{N} + 2\eta \frac{d}{N}.
\end{align}
By Lemma~\ref{lemma:inf:seq}, we get
\begin{align}
\label{avg:ex:t1}
 \left\Vert \overline{\mathbf{e}}_x \right\Vert_2^{\delta, K} & \leq \frac{1}{\delta^2}\left(1 - \eta\mu\left(1 - \frac{\eta L}{2} \right) \right) \left\Vert \overline{\mathbf{e}}_x \right\Vert_2^{\delta, K} 
\nonumber 
\\
& \qquad + \frac{\eta L^2}{\delta^2 N}\left(\frac{\eta}{N} + \frac{1+\eta L}{N\mu\left(1 - \frac{\eta L}{2} \right)} \right)\left\Vert \widetilde{\mathbf{x}}\right\Vert_2^{\delta, K} + \frac{\eta}{N\delta^{2K}}\left(\eta\sigma^2 + 2d \right) + \mathbb{E}\left[ \left\Vert \overline{e}_{x}^{(0)}\right\Vert^2 \right].
\end{align}
By taking $\delta \in \left(\sqrt{1 - \eta\mu\left(1-\frac{\eta L}{2}\right)}\,,\,1\right)$, we can compute from \eqref{avg:ex:t1}, it follows
\begin{align}
& \left(\delta^2 +\eta\mu\left(1 - \frac{\eta L}{2}\right) - 1 \right)\left\Vert \overline{\mathbf{e}}_x \right\Vert_2^{\delta, K} 
\nonumber 
\\
& \qquad \leq \frac{\eta L^2}{N}\left(\frac{\eta}{N} + \frac{1+\eta L}{N\mu\left(1 - \frac{\eta L}{2} \right)} \right)\left\Vert \widetilde{\mathbf{x}}\right\Vert_2^{\delta, K} + \frac{\eta\left(\eta\sigma^2 + 2d \right)}{N\delta^{2K-2}}
+ \delta^2\mathbb{E}\left[ \left\Vert \overline{e}_{x}^{(0)}\right\Vert^2 \right].
\end{align}
Therefore, we conclude that for every $K\geq 0$,
\begin{align}
\left\Vert \overline{\mathbf{e}}_x \right\Vert_2^{\delta, K}  
& \leq \eta\cdot\frac{L^2}{N^2\left(\delta^2 +\eta\mu\left(1 - \frac{\eta L}{2}\right) - 1 \right)}\left(\eta + \frac{1 + \eta L}{\mu\left(1-\frac{\eta L}{2}\right)} \right)\left\Vert \widetilde{\mathbf{x}}\right\Vert_2^{\delta, K} 
\nonumber 
\\
& \qquad\qquad + \frac{\eta}{N\delta^{2K-2}} \cdot \frac{\eta\sigma^2 + 2d}{\delta^2 +\eta\mu\left(1 - \frac{\eta L}{2}\right) - 1}
+ \frac{\delta^2}{\delta^2 +\eta\mu\left(1 - \frac{\eta L}{2}\right) - 1}\mathbb{E}\left[ \left\Vert \overline{e}_{x}^{(0)}\right\Vert^2 \right].
\end{align}
This completes the proof. 


\subsection{Proof of Lemma~\ref{lemma:tilde:x}}
Following~\eqref{iter1}, we can compute that
\begin{equation}
x^{(k+1)} - \mathbf{x}_* = \widetilde{\mathcal{W}}x^{(k)} - \mathbf{x}_*  - \eta\left(\nabla F\left(x^{(k)}\right)  + v^{(k)} + \nabla F(\mathbf{x}_*) - \nabla F(\mathbf{x}_*)\right) - \eta\xi^{(k)} + \sqrt{2\eta}w^{(k+1)}.
\end{equation}
Next, by using $\widetilde{\mathcal{W}}\mathbf{x}_* = \mathbf{x}_*$ and~\eqref{tilde:v} in Lemma~\ref{lemma:err}, we get
\begin{equation}
\label{eqn:errx}
e_x^{(k+1)} = \widetilde{\mathcal{W}}e_x^{(k)} - \eta\left(\nabla F\left(x^{(k)}\right) - \nabla F(\mathbf{x}_*)\right) - \eta \widetilde{v}^{(k)} - \eta\xi^{(k)} + \sqrt{2\eta}w^{(k+1)}.
\end{equation}
Moreover, we use the definition of $e_x^{(k)}$ in~\eqref{def:ex} from Lemma~\ref{lemma:err} and $\mathcal{J} = J \otimes I_d$ with $J = \frac{1}{N}1_N1_N^T$ to get 
\begin{equation}
\label{eqn:rel:1}
\widetilde{x}^{(k)} = e_x^{(k)} - 1_N \otimes \overline{e}_x^{(k)} = \left[\left(I_N-J\right) \otimes I_d \right] e^{(k)}_x,
\end{equation}
where the component on $(i,j)$-position of $\left[\left(I_N-J\right) \otimes I_d \right] e^{(k)}_x$ is $[e_x^{(k)}]_{ij} - \frac{1}{N}\sum_{i=1}^N[e_x^{(k)}]_{ij}$ for $i = 1,\ldots,N$. Hence, we obtained the last equality in \eqref{eqn:rel:1} which can be re-written as: 
\begin{equation}
\widetilde{x}^{(k)} = \left(I_{Nd} - \mathcal{J}\right)e_x^{(k)}.
\end{equation} 
Next, by multiplying $\left(I_{Nd} - \mathcal{J}\right)$ on both hand sides of \eqref{eqn:errx}, we get
\begin{align}
\widetilde{x}^{(k+1)} & = \left(I_{Nd}-\mathcal{J}\right)\widetilde{\mathcal{W}}\widetilde{x}^{(k)} - \eta\left(I_{Nd}-\mathcal{J}\right)\left(\nabla F\left(x^{(k)}\right) - \nabla F(\mathbf{x}_*)\right) 
\nonumber 
\\
& \qquad\qquad\qquad\qquad - \eta \left(I_{Nd}-\mathcal{J}\right)\widetilde{v}^{(k)} - \eta\left(I_{Nd}-\mathcal{J}\right)\xi^{(k)} + \sqrt{2\eta}\left(I_{Nd}-\mathcal{J}\right)w^{(k+1)}
\nonumber 
\\
& = \left(\mathcal{\widetilde{W} - J}\right)\widetilde{x}^{(k)} - \eta\left(I_{Nd}-\mathcal{J}\right)\left(\nabla F\left(x^{(k)}\right) - \nabla F(\mathbf{x}_*)\right) 
\nonumber 
\\
& \qquad\qquad\qquad\qquad - \eta\widetilde{v}^{(k)}
 - \eta\left(\xi^{(k)} - \overline{\xi}^{(k)}\right) + \sqrt{2\eta}\left(w^{(k+1)} - \overline{w}^{(k+1)}\right),
\end{align}
where we used $(I_{N} - J)W = W - JW = W - J$ and $\mathcal{J}\widetilde{v}^{(k)} = \mathcal{J}e_v^{(k)} = \overline{e}_v^{(k)} \otimes 1_N^T = \mathbf{0}$ in Lemma~\ref{lemma:err} to get the last equality. In the following, we can compute that
\begin{align}
\left\Vert \widetilde{x}^{(k+1)} \right\Vert^2 & = \left\Vert \left(\mathcal{\widetilde{W} - J}\right)\widetilde{x}^{(k)} - \eta\left(I_{Nd}-\mathcal{J}\right)\left(\nabla F\left(x^{(k)}\right) - \nabla F(\mathbf{x}_*)\right)  \right\Vert^2 
\nonumber 
\\
& \qquad\qquad + 2\bigg\langle \left(\mathcal{\widetilde{W} - J}\right)\widetilde{x}^{(k)} - \eta\left(I_{Nd}-\mathcal{J}\right)\left(\nabla F\left(x^{(k)}\right) - \nabla F(\mathbf{x}_*)\right) \,,\, 
\nonumber 
\\
& \qquad\qquad\qquad\qquad\qquad\qquad - \eta\widetilde{v}^{(k)}
 - \eta\left(\xi^{(k)} - \overline{\xi}^{(k)}\right) + \sqrt{2\eta}\left(w^{(k+1)} - \overline{w}^{(k+1)}\right) \bigg\rangle
 \nonumber 
 \\
 & \qquad\qquad + \left\Vert  - \eta\widetilde{v}^{(k)}
 - \eta\left(\xi^{(k)} - \overline{\xi}^{(k)}\right) + \sqrt{2\eta}\left(w^{(k+1)} - \overline{w}^{(k+1)}\right)  \right\Vert^2.
\label{tildex}
\end{align}
By the fact that $\mathcal{J} = J \otimes I_d$, we have $\lambda_{\max}(J) = 1/N$ and $\lambda_{\min}(J) = 0$. We can further compute
\begin{equation}
\label{norm:ineq}
\left\Vert \left(I_{Nd}-\mathcal{J}\right)\left(\nabla F\left(x^{(k)}\right) - \nabla F(\mathbf{x}_*)\right)  \right\Vert^2 \leq \left\Vert\nabla F\left(x^{(k)}\right) - \nabla F(\mathbf{x}_*)\right\Vert^2,
\end{equation}
where we used $\lambda_{\max}^2(I_{N}-J) = (1 - \lambda_{\min}(J))^2$ since $\lambda(I_{N}-A) = 1 - \lambda(A)$
for any $N\times N$ matrix $A$.

Next, we can compute the first term in~\eqref{tildex} as below.
\begin{align}
& \left\Vert \left(\mathcal{\widetilde{W} - J}\right)\widetilde{x}^{(k)} - \eta\left(I_{Nd}-\mathcal{J}\right)\left(\nabla F\left(x^{(k)}\right) - \nabla F(\mathbf{x}_*)\right)  \right\Vert^2 
\nonumber 
\\
& \quad = \left\vert\lambda_2^{{\scaleto{\widetilde{W}}{5pt}}}\right\vert^2\left\Vert \widetilde{x}^{(k)}  \right\Vert^2 + \eta^2\left\Vert \left(I_{Nd}-\mathcal{J}\right)\left(\nabla F\left(x^{(k)}\right) - \nabla F(\mathbf{x}_*)\right)  \right\Vert^2
\label{tildex:t1} 
\\
& \qquad\qquad\qquad\qquad -2\eta \left\langle \left(\mathcal{\widetilde{W} - J}\right)\widetilde{x}^{(k)} \,,\, \left(I_{Nd}-\mathcal{J}\right)\left(\nabla F\left(x^{(k)}\right) - \nabla F(\mathbf{x}_*)\right)  \right\rangle
\nonumber 
\\
& \quad \leq  \left\vert\lambda_2^{{\scaleto{\widetilde{W}}{5pt}}}\right\vert^2\left\Vert \widetilde{x}^{(k)}  \right\Vert^2 + \eta^2\left\Vert\nabla F\left(x^{(k)}\right) - \nabla F(\mathbf{x}_*)\right\Vert^2 -2\eta \left\langle e_x^{(k)} \,,\, \nabla F\left(x^{(k)}\right) - \nabla F(\mathbf{x}_*) \right\rangle
\nonumber 
\\
& \qquad\qquad\qquad\qquad - 2\eta\left\langle\left(\left(\mathcal{\widetilde{W} - J}\right)\left(I_{Nd}-\mathcal{J}\right)^2 - I_{Nd}\right)  e_x^{(k)} \,,\, \nabla F\left(x^{(k)}\right) - \nabla F(\mathbf{x}_*) \right\rangle
\nonumber 
\\
& \quad \leq  \left\vert\lambda_2^{{\scaleto{\widetilde{W}}{5pt}}}\right\vert^2\left\Vert \widetilde{x}^{(k)}  \right\Vert^2 + \eta^2\left\Vert\nabla F\left(x^{(k)}\right) - \nabla F(\mathbf{x}_*)\right\Vert^2 -2\eta \left\langle e_x^{(k)} \,,\, \nabla F\left(x^{(k)}\right) - \nabla F(\mathbf{x}_*) \right\rangle
\nonumber 
\\
& \qquad\qquad\qquad\qquad  + \eta\left(\frac{1}{L}\left\Vert  \nabla F\left(x^{(k)}\right) - \nabla F(\mathbf{x}_*)  \right\Vert^2 + L\left\Vert e_x^{(k)} \right\Vert^2 \right)
\label{tildex:t2}
\\
& \quad =  \left\vert\lambda_2^{{\scaleto{\widetilde{W}}{5pt}}}\right\vert^2\left\Vert \widetilde{x}^{(k)}  \right\Vert^2 + \left(\eta^2 + \frac{\eta }{L}\right)\left\Vert\nabla F\left(x^{(k)}\right) - \nabla F(\mathbf{x}_*)\right\Vert^2 
\nonumber 
\\
& \qquad\qquad\qquad\qquad -2\eta \left\langle e_x^{(k)} \,,\, \nabla F\left(x^{(k)}\right) - \nabla F(\mathbf{x}_*) \right\rangle + \eta L\left\Vert e_x^{(k)} \right\Vert^2
\nonumber 
\\
& \quad \leq  \left\vert\lambda_2^{{\scaleto{\widetilde{W}}{5pt}}}\right\vert^2\left\Vert \widetilde{x}^{(k)}  \right\Vert^2 -\left(\eta - \eta^2L \right)\left\langle e_x^{(k)} \,,\, \nabla F\left(x^{(k)}\right) - \nabla F(\mathbf{x}_*) \right\rangle  + \eta L\left\Vert e_x^{(k)} \right\Vert^2
\label{tildex:t3}
\\
& \quad \leq  \left\vert\lambda_2^{{\scaleto{\widetilde{W}}{5pt}}}\right\vert^2\left\Vert \widetilde{x}^{(k)}  \right\Vert^2 -\left(\eta - \eta^2L \right)\mu\left\Vert e_x^{(k)} \right\Vert^2  + \eta L\left\Vert e_x^{(k)} \right\Vert^2
\label{tildex:t4}
\\
& \quad =  \left\vert\lambda_2^{{\scaleto{\widetilde{W}}{5pt}}}\right\vert^2\left\Vert \widetilde{x}^{(k)}  \right\Vert^2 + \eta\mu\left(\eta L + (L/\mu) -1\right)\left\Vert e_x^{(k)} \right\Vert^2
\nonumber 
\\
& \quad =  \left(\left\vert\lambda_2^{{\scaleto{\widetilde{W}}{5pt}}}\right\vert^2 + \eta\mu\left(\eta L + (L/\mu) -1 \right) \right)\left\Vert \widetilde{x}^{(k)}\right\Vert^2  + \eta\mu \left(\eta L + (L/\mu) -1\right) \left\Vert \overline{e}_x^{(k)} \right\Vert^2,
\label{tildex:t5}
\end{align}
where we used the inequality that
$$
\left\Vert \left(\left(\widetilde{W} - J\right) \otimes I_d \right)\widetilde{x}^{(k)} \right\Vert^2 \leq \left\Vert \widetilde{W} - J  \right\Vert^2 \left\Vert \widetilde{x}^{(k)}  \right\Vert^2 \leq \left\vert\lambda_2^{{\scaleto{\widetilde{W}}{5pt}}}\right\vert^2\left\Vert \widetilde{x}^{(k)}  \right\Vert^2,
$$
where $1 = \lambda_1^{{\scaleto{\widetilde{W}}{5pt}}} > \lambda_2^{{\scaleto{\widetilde{W}}{5pt}}} \geq \ldots \geq \lambda_N^{{\scaleto{\widetilde{W}}{5pt}}} > 0$ to get~\eqref{tildex:t1}. To get~\eqref{tildex:t2}, we used matrix property of $J$, $J = \frac{1}{N}1_N1_N^T$ is symmetric and $(1_N1_N^T)A = A(1_N1_N^T)$, so that $(I_{N}-J)^T A = A(I_{N}-J)$. We notice $(\widetilde{W} - J)J = J$, and thus 
$$
\left(\widetilde{W} - J\right)\left(I_{N}-J\right)^2 = \left(\widetilde{W} - 2J\right)\left(I_{N}-J\right) = \widetilde{W} - J,
$$
where we used the fact that $J^2 = J$. Moreover, we can also compute
$$
\left\Vert \left(\left(\widetilde{W} - J - I_{N}\right) \otimes I_d\right) e_{x}^{(k)} \right\Vert^2 \leq \left\Vert \widetilde{W} - I_{N} - J\right\Vert^2 \left\Vert e_x^{(k)} \right\Vert^2 \leq \left\Vert e_x^{(k)} \right\Vert^2, 
$$
where we have $\lambda(\widetilde{W} - I_{N}) = \lambda(\widetilde{W})-1 \in (-1,0]$, and since $J = \frac{1}{N}1_N^T1_N$, we can decompose $\widetilde{W} - I_{N} - J$ and $\widetilde{W} - I_{N}$ in the same eigenspace, so that they have the same non-zero eigenvalues, and the largest eigenvalue $\lambda_{\max}(\widetilde{W} - I_{N} - J)= 1$ corresponds to eigenvector $1^T_N$. Hence, we obtain
\begin{align}
& - 2\eta\left\langle\left(\left(\mathcal{\widetilde{W} - J}\right)\left(I_{Nd}-\mathcal{J}\right)^2 - I_{Nd}\right)  e_x^{(k)} \,,\, \nabla F\left(x^{(k)}\right) - \nabla F(\mathbf{x}_*) \right\rangle
\nonumber 
\\
& \quad \leq 2\eta\left\Vert \left(\left(\mathcal{\widetilde{W} - J}\right)\left(I_{Nd}-\mathcal{J}\right)^2 - I_{Nd}\right)  e_x^{(k)} \right\Vert \cdot \left\Vert \nabla F\left(x^{(k)}\right) - \nabla F(\mathbf{x}_*)  \right\Vert
\nonumber 
\\
& \quad = 2\eta\left\Vert \left(\mathcal{\widetilde{W} - J} - I_{Nd}\right)  e_x^{(k)} \right\Vert \cdot \left\Vert \nabla F\left(x^{(k)}\right) - \nabla F(\mathbf{x}_*)  \right\Vert
\nonumber 
\\
& \quad \leq \eta\left(\frac{1}{L}\left\Vert  \nabla F\left(x^{(k)}\right) - \nabla F(\mathbf{x}_*)  \right\Vert^2 + L\left\Vert e_x^{(k)} \right\Vert^2 \right),
\end{align}
where we used Cauchy-Schwarz inequality for the inner product and the inequality $2xy \leq cx^2 + \frac{y^2}{c}$ for any $x,y\in\mathbb{R}$ by taking $c = L > 0$. Next, we used $L$-smoothness of $F$ to get~\eqref{tildex:t3} and $\mu$-convexity of $F$ to get~\eqref{tildex:t4} where we also used  $\eta \leq 1/L$. Finally,  we used triangle inequality to $\widetilde{x}^{(k)} + 1_N \otimes \overline{e}_x^{(k)} = e_x^{(k)}$ and get 
$
\left\Vert e_x^{(k)} \right\Vert^2 \leq \left\Vert \widetilde{x}^{(k)} \right\Vert^2 + \left\Vert\overline{e}_x^{(k)}\right\Vert^2,
$ 
for~\eqref{tildex:t5}. 

Now by using the fact that the expectation of random noise terms and their average terms are zero conditioning on the natural filtration to compute the expectation of the inner product term in~\eqref{tildex}, we get
\begin{align}
& \mathbb{E}\bigg\langle \left(\mathcal{\widetilde{W} - J}\right)\widetilde{x}^{(k)} - \eta\left(I_{Nd}-\mathcal{J}\right)\left(\nabla F\left(x^{(k)}\right) - \nabla F(\mathbf{x}_*)\right) \,,\,  \eta\widetilde{v}^{(k)}\bigg\rangle
\nonumber 
\\
& \quad = \eta \mathbb{E}\left\langle \mathcal{\widetilde{W}} \widetilde{x}^{(k)}\,,\, \widetilde{v}^{(k)} \right\rangle + \eta^2\mathbb{E}\left\langle \left(I_{Nd}-\mathcal{J}\right)\left(\nabla F(\mathbf{x}_*) - \nabla F\left(x^{(k)}\right)\right)\,,\, \widetilde{v}^{(k)} \right\rangle
\label{tildex:inner:t1} 
\\
& \quad \leq \eta\mathbb{E}\left[\left\Vert \mathcal{\widetilde{W}}\widetilde{x}^{(k)} \right\Vert \cdot \left\Vert \widetilde{v}^{(k)} \right\Vert \right] + \eta^2 \mathbb{E}\left[\left\Vert \left(I_{Nd}-\mathcal{J}\right)\left(\nabla F(\mathbf{x}_*) - \nabla F\left(x^{(k)}\right)\right)\right\Vert \cdot \left\Vert \widetilde{v}^{(k)}  \right\Vert \right]
\nonumber 
\\
& \quad \leq \eta\mathbb{E}\left[\left\Vert \widetilde{x}^{(k)} \right\Vert \cdot \left\Vert \widetilde{v}^{(k)} \right\Vert \right] + \eta^2 \mathbb{E}\left[\left\Vert \nabla F(\mathbf{x}_*) - \nabla F\left(x^{(k)}\right)\right\Vert \cdot \left\Vert \widetilde{v}^{(k)}  \right\Vert \right]
\label{tildex:inner:t5} 
\\
& \quad \leq \eta L \mathbb{E} \left\Vert \widetilde{x}^{(k)} \right\Vert^2 + \frac{\eta}{4L}\mathbb{E}\left[\left\Vert \widetilde{v}^{(k)} \right\Vert^2\right] 
\label{tildex:inner:t2} 
\\
& \qquad\qquad\qquad\qquad + \eta^2 L\mu \mathbb{E}\left[\left\Vert e_x^{(k)}\right\Vert^2\right] + \frac{\eta^2}{4L\mu}\mathbb{E}\left[\left\Vert \widetilde{v}^{(k)}  \right\Vert^2\right]
\label{tildex:inner:t3}
\\
& \quad \leq (\eta L + \eta^2L\mu) \mathbb{E} \left\Vert \widetilde{x}^{(k)} \right\Vert^2 + \eta^2 \mu L\mathbb{E}\left[\left\Vert \overline{e}_x^{(k)}\right\Vert^2\right] + \left(\frac{\eta}{4L} + \frac{\eta^2}{4L\mu}\right)\mathbb{E}\left[\left\Vert \widetilde{v}^{(k)}  \right\Vert^2\right].
\label{tildex:inner:t4}
\end{align}
To get~\eqref{tildex:inner:t1}, we used Cauchy-Schwarz inequality and the fact that $\left\langle \mathcal{J} \widetilde{x}^{(k)}\,,\, \widetilde{v}^{(k)} \right\rangle = \left\langle \widetilde{x}^{(k)}\,,\, \mathcal{J}^T\widetilde{v}^{(k)} \right\rangle $, where $\mathcal{J}^T\widetilde{v}^{(k)} = \mathcal{J}\widetilde{v}^{(k)} = \mathbf{0}$ by Lemma~\ref{lemma:err}. Recall we have the eigenvalues of $\widetilde{W}$ are $1 = \lambda_1^{{\scaleto{\widetilde{W}}{5pt}}} > \lambda_2^{{\scaleto{\widetilde{W}}{5pt}}} \geq \cdots \geq \lambda_{N}^{{\scaleto{\widetilde{W}}{5pt}}} > 0$ to have
$
\left\Vert\left(\widetilde{W} \otimes I_d\right) \widetilde{x}^{(k)}\right\Vert^2 \leq \left\Vert \widetilde{x}^{(k)} \right\Vert^2
$, which yields~\eqref{tildex:inner:t5}. Then we used the inequality $2xy \leq c'x^2 + \frac{y^2}{c'}$ for $c' > 0$ and took $c' = 2L$ to get~\eqref{tildex:inner:t2}. Next, we used the same inequality by taking $c' = \frac{2}{L}$ and using $L$-smoothness of $F$ to get~\eqref{tildex:inner:t3}. Finally, we used triangle inequality to $\widetilde{x}^{(k)} + 1_N \otimes \overline{e}_x^{(k)} = e_x^{(k)}$ and get 
$
\left\Vert e_x^{(k)} \right\Vert^2 \leq \left\Vert \widetilde{x}^{(k)} \right\Vert^2 + \left\Vert\overline{e}_x^{(k)}\right\Vert^2,
$ 
and choose $\eta \leq 1 / L$ to obtain the last inequality~\eqref{tildex:inner:t4}.

Accordingly, we can get the following inequality from~\eqref{tildex} and~\eqref{tildex:t5},
\begin{align}
\mathbb{E}\left[\left\Vert \widetilde{x}^{(k+1)} \right\Vert^2\right] & \leq \left(\left\vert\lambda_2^{{\scaleto{\widetilde{W}}{5pt}}}\right\vert^2 + \eta\mu\left(3(L/\mu) + 3\eta L - 1\right) \right) \mathbb{E} \left\Vert \widetilde{x}^{(k)} \right\Vert^2 
\nonumber 
\\
& \qquad + \eta\mu\left(L/\mu + 3\eta L - 1\right)\mathbb{E}\left[\left\Vert \overline{e}_x^{(k)} \right\Vert^2\right]
+ \eta\left(\frac{1}{2L} + \eta + \frac{\eta}{2L\mu }\right)\mathbb{E}\left[\left\Vert \widetilde{v}^{(k)} \right\Vert^2\right]
\nonumber 
\\
 & \qquad\qquad\qquad\qquad + \eta^2\mathbb{E}\left[\left\Vert \xi^{(k)} - \overline{\xi}^{(k)}\right\Vert^2\right] + 2\eta\mathbb{E}\left[\left\Vert w^{(k+1)} - \overline{w}^{(k+1)}\right\Vert^2\right],
 \label{tildex:t6}
\end{align}
where we can further compute
\begin{align}
& \eta^2\mathbb{E}\left[\left\Vert \xi^{(k)} - \overline{\xi}^{(k)}\right\Vert^2\right] + 2\eta\mathbb{E}\left[\left\Vert w^{(k+1)} - \overline{w}^{(k+1)}\right\Vert^2\right]
\nonumber 
\\
& \quad = \eta^2\mathbb{E}\left[\left\Vert \xi^{(k)}\right\Vert^2\right] + \eta^2 \mathbb{E}\left[\left\Vert \overline{\xi}^{(k)}\right\Vert^2\right] + 2\eta\mathbb{E}\left[\left\Vert w^{(k+1)}\right\Vert^2\right] + 2\eta \mathbb{E}\left[\left\Vert \overline{w}^{(k+1)}\right\Vert^2\right]
\nonumber 
\\
& \quad = \eta\left(N + \frac{1}{N}\right)\left(\eta\sigma^2 + 2d \right).
\end{align}
Under the assumption of the stepsize in Lemma~\ref{lemma:avi:ex}, Lemma~\ref{lemma:inf:seq}, in particular, Lemma~\ref{lemma:seq} and the inequality~\eqref{tildex:t6}, it follows that
\begin{align}
\delta^2\left\Vert \widetilde{\mathbf{x}} \right\Vert_2^{\delta, K} 
& \leq \left(\left\vert\lambda_2^{{\scaleto{\widetilde{W}}{5pt}}}\right\vert^2 +  \eta\mu\left(3(L/\mu) + 3\eta L - 1\right) \right) \left\Vert \widetilde{\mathbf{x}} \right\Vert_2^{\delta, K}  
\nonumber 
\\
& \qquad\qquad + \eta\mu \left(L/\mu + 3\eta L - 1\right)\left\Vert \overline{\mathbf{e}}_x \right\Vert_2^{\delta, K} 
 + \eta\left(\frac{1}{2L} + \eta + \frac{\eta}{2L\mu }\right)\left\Vert \widetilde{\mathbf{v}} \right\Vert_2^{\delta, K} 
\nonumber 
\\
 & \qquad\qquad\qquad\qquad + \frac{\eta}{\delta^{2K-2}}\left(N + \frac{1}{N} \right)(\eta\sigma^2 + 2d) + \delta^2\mathbb{E}\left[ \left\Vert \widetilde{x}^{(0)}\right\Vert^2 \right].
 \label{tildex:t7}
\end{align}
Next, we consider the following three scenarios.

(1). If $\left\vert \lambda_2^{{\scaleto{\widetilde{W}}{5pt}}} \right\vert^2 < \frac{1}{2}$, then for all $1 > \delta^2 \geq 2\left\vert \lambda_2^{{\scaleto{\widetilde{W}}{5pt}}} \right\vert^2$, we compute that 
\begin{align}
\delta^2 - \left\vert \lambda_2^{{\scaleto{\widetilde{W}}{5pt}}} \right\vert^2 - \eta\mu\left(3(L/\mu) + 3\eta L  - 1\right)
&\geq \left\vert \lambda_2^{{\scaleto{\widetilde{W}}{5pt}}} \right\vert^2 - 3L\eta - 3\eta^2 \mu L  + \eta\mu
\nonumber 
\\
&\geq  \left\vert \lambda_2^{{\scaleto{\widetilde{W}}{5pt}}} \right\vert^2 - 3(L+\mu)\eta,\label{case:1}
\end{align}
where we used the assumption $\eta L \leq 1$ to get $3\eta^2\mu L \leq 3\mu\eta$ and obtain the last inequality. Thus, by the assumption that 
\begin{equation}
\label{tildex:eta:1}
\eta \leq \frac{\left\vert \lambda_2^{{\scaleto{\widetilde{W}}{5pt}}} \right\vert^2}{6(L+\mu)},
\end{equation}
we obtain
\begin{equation}
\delta^2 - \left\vert \lambda_2^{{\scaleto{\widetilde{W}}{5pt}}} \right\vert^2 - \eta\mu\left(3(L/\mu) + 3\eta L  - 1 \right) > \frac{\left\vert \lambda_2^{{\scaleto{\widetilde{W}}{5pt}}} \right\vert^2}{2} > 0,
\end{equation}

(2). If $\frac{2}{3} \geq \left\vert \lambda_2^{{\scaleto{\widetilde{W}}{5pt}}} \right\vert^2 \geq \frac{1}{2}$, for all $1 > \delta^2 \geq \frac{\left\vert \lambda_2^{{\scaleto{\widetilde{W}}{5pt}}} \right\vert^2}{2\left(1 - \left\vert \lambda_2^{{\scaleto{\widetilde{W}}{5pt}}} \right\vert^2\right)} \geq \frac{1}{2}$. 
We can compute that 
\begin{align}
\label{case:2}
\delta^2 - \left\vert \lambda_2^{{\scaleto{\widetilde{W}}{5pt}}} \right\vert^2 
& \geq \frac{\left\vert \lambda_2^{{\scaleto{\widetilde{W}}{5pt}}} \right\vert^2}{2\left(1 - \left\vert \lambda_2^{{\scaleto{\widetilde{W}}{5pt}}} \right\vert^2\right)} - \left\vert \lambda_2^{{\scaleto{\widetilde{W}}{5pt}}} \right\vert^2 = \frac{\left\vert \lambda_2^{{\scaleto{\widetilde{W}}{5pt}}} \right\vert^2\left(\left\vert \lambda_2^{{\scaleto{\widetilde{W}}{5pt}}} \right\vert^2 - \frac{1}{2}\right)}{1 - \left\vert \lambda_2^{{\scaleto{\widetilde{W}}{5pt}}} \right\vert^2} > 0,
\end{align}
and under the assumption that
\begin{equation}
\label{tildex:eta:2}
\eta \leq \frac{\left\vert \lambda_2^{{\scaleto{\widetilde{W}}{5pt}}} \right\vert^2\left(\left\vert \lambda_2^{{\scaleto{\widetilde{W}}{5pt}}} \right\vert^2 - \frac{1}{2}\right)}{6(L + \mu)\left(1 - \left\vert \lambda_2^{{\scaleto{\widetilde{W}}{5pt}}} \right\vert^2\right)},
\end{equation} 
we can compute that
\begin{align}
\delta^2 - \left\vert \lambda_2^{{\scaleto{\widetilde{W}}{5pt}}} \right\vert^2 - \eta\mu\left(3(L/\mu) + 3\eta L  - 1 \right)
& \geq \delta^2 - \left\vert \lambda_2^{{\scaleto{\widetilde{W}}{5pt}}} \right\vert^2 -3(L+\mu)\eta
\nonumber 
\\
& \geq \frac{\left\vert \lambda_2^{{\scaleto{\widetilde{W}}{5pt}}} \right\vert^2\left(\left\vert \lambda_2^{{\scaleto{\widetilde{W}}{5pt}}} \right\vert^2 - \frac{1}{2}\right)}{1 - \left\vert \lambda_2^{{\scaleto{\widetilde{W}}{5pt}}} \right\vert^2} - 3(L + \mu)\eta
\nonumber 
\\
& \geq \frac{\left\vert \lambda_2^{{\scaleto{\widetilde{W}}{5pt}}} \right\vert^2\left(\left\vert \lambda_2^{{\scaleto{\widetilde{W}}{5pt}}} \right\vert^2 - \frac{1}{2}\right)}{2\left(1 - \left\vert \lambda_2^{{\scaleto{\widetilde{W}}{5pt}}} \right\vert^2\right)} > 0.
\end{align}

(3). If $1 > \left\vert \lambda_2^{{\scaleto{\widetilde{W}}{5pt}}} \right\vert^2 > \frac{2}{3}$, we can easily find the quantity relation $1 > \frac{4\left\vert \lambda_2^{{\scaleto{\widetilde{W}}{5pt}}} \right\vert^2-2}{3\left\vert \lambda_2^{{\scaleto{\widetilde{W}}{5pt}}} \right\vert^2-1} > \frac{1}{2}$, then for all $1 > \delta^2 \geq \frac{4\left\vert \lambda_2^{{\scaleto{\widetilde{W}}{5pt}}} \right\vert^2-2}{3\left\vert \lambda_2^{{\scaleto{\widetilde{W}}{5pt}}} \right\vert^2-1} > \frac{1}{2}$, we can compute that 
\begin{align}
\label{case:3}
\delta^2 - \left\vert \lambda_2^{{\scaleto{\widetilde{W}}{5pt}}} \right\vert^2 & \geq \frac{4\left\vert \lambda_2^{{\scaleto{\widetilde{W}}{5pt}}} \right\vert^2-2}{3\left\vert \lambda_2^{{\scaleto{\widetilde{W}}{5pt}}} \right\vert^2-1} - \left\vert \lambda_2^{{\scaleto{\widetilde{W}}{5pt}}} \right\vert^2 = \frac{5\left\vert \lambda_2^{{\scaleto{\widetilde{W}}{5pt}}} \right\vert^2 - 3\left\vert \lambda_2^{{\scaleto{\widetilde{W}}{5pt}}} \right\vert^4-2}{3\left\vert \lambda_2^{{\scaleto{\widetilde{W}}{5pt}}} \right\vert^2-1} > 0,
\end{align}
where we can find $2/3,1$ are two roots for $5x - 3x^2 - 2 = 0$, such that $\frac{5x - 3x^2 - 1}{3x - 1} > 0$ for any $1 > x > \frac{2}{3}$ which implies \eqref{case:3}.
Under the assumption on $\eta$ such that 
\begin{equation}
\label{tildex:eta:3}
\eta \leq \frac{5\left\vert \lambda_2^{{\scaleto{\widetilde{W}}{5pt}}} \right\vert^2 - 3\left\vert \lambda_2^{{\scaleto{\widetilde{W}}{5pt}}} \right\vert^4-2}{6(L + \mu)\left(3\left\vert \lambda_2^{{\scaleto{\widetilde{W}}{5pt}}} \right\vert^2-1\right)},
\end{equation} 
we can further compute
\begin{align}
\delta^2 - \left\vert \lambda_2^{{\scaleto{\widetilde{W}}{5pt}}} \right\vert^2 - \eta\mu\left(3(L/\mu) + 3\eta L  - 1 \right)
& \geq \delta^2 - \left\vert \lambda_2^{{\scaleto{\widetilde{W}}{5pt}}} \right\vert^2 -3(L+\mu)\eta
\nonumber 
\\
& \geq \frac{5\left\vert \lambda_2^{{\scaleto{\widetilde{W}}{5pt}}} \right\vert^2 - 3\left\vert \lambda_2^{{\scaleto{\widetilde{W}}{5pt}}} \right\vert^4-2}{3\left\vert \lambda_2^{{\scaleto{\widetilde{W}}{5pt}}} \right\vert^2-1} - 3(L + \mu)\eta
\nonumber 
\\
& \geq \frac{1}{2} \cdot \frac{5\left\vert \lambda_2^{{\scaleto{\widetilde{W}}{5pt}}} \right\vert^2 - 3\left\vert \lambda_2^{{\scaleto{\widetilde{W}}{5pt}}} \right\vert^4-2}{3\left\vert \lambda_2^{{\scaleto{\widetilde{W}}{5pt}}} \right\vert^2-1} > 0.
\end{align}
Moreover, we can find from~\eqref{tildex:eta:1},~\eqref{tildex:eta:2} and~\eqref{tildex:eta:3} such that 
\begin{equation}
\eta \leq \frac{\gamma_{{\scaleto{\widetilde{W}}{5pt}}}}{6(L + \mu)},
\end{equation}
where we define the constant
\begin{equation}
\label{tildex:def:const:lambda}
\gamma_{{\scaleto{\widetilde{W}}{5pt}}}:= 
\begin{cases}
\left\vert \lambda_2^{{\scaleto{\widetilde{W}}{5pt}}} \right\vert^2 &\text{if $0 < \left\vert \lambda_2^{{\scaleto{\widetilde{W}}{5pt}}} \right\vert^2 < \frac{1}{2}$},
\\
\frac{\left\vert \lambda_2^{{\scaleto{\widetilde{W}}{5pt}}} \right\vert^2\left(\left\vert \lambda_2^{{\scaleto{\widetilde{W}}{5pt}}} \right\vert^2 - \frac{1}{2}\right)}{1 - \left\vert \lambda_2^{{\scaleto{\widetilde{W}}{5pt}}} \right\vert^2} &\text{if $\frac{1}{2} \leq \left\vert \lambda_2^{{\scaleto{\widetilde{W}}{5pt}}} \right\vert^2 \leq \frac{2}{3}$},
\\
\frac{5\left\vert \lambda_2^{{\scaleto{\widetilde{W}}{5pt}}} \right\vert^2 - 3\left\vert \lambda_2^{{\scaleto{\widetilde{W}}{5pt}}} \right\vert^4-2}{3\left\vert \lambda_2^{{\scaleto{\widetilde{W}}{5pt}}} \right\vert^2-1} &\text{if $ \frac{2}{3} \leq \left\vert \lambda_2^{{\scaleto{\widetilde{W}}{5pt}}} \right\vert^2 < 1$}.
\end{cases}
\end{equation}
We note that the quantities $\left\vert \lambda_2^{{\scaleto{\widetilde{W}}{5pt}}} \right\vert^2\,,\, \frac{\left\vert \lambda_2^{{\scaleto{\widetilde{W}}{5pt}}} \right\vert^2\left(\left\vert \lambda_2^{{\scaleto{\widetilde{W}}{5pt}}} \right\vert^2 - \frac{1}{2}\right)}{1 - \left\vert \lambda_2^{{\scaleto{\widetilde{W}}{5pt}}} \right\vert^2} \,,\, \frac{5\left\vert \lambda_2^{{\scaleto{\widetilde{W}}{5pt}}} \right\vert^2 - 3\left\vert \lambda_2^{{\scaleto{\widetilde{W}}{5pt}}} \right\vert^4-2}{3\left\vert \lambda_2^{{\scaleto{\widetilde{W}}{5pt}}} \right\vert^2-1}$ are positive over the regimes $0 < \left\vert \lambda_2^{{\scaleto{\widetilde{W}}{5pt}}} \right\vert^2 < \frac{1}{2}$, $\frac{1}{2} \leq \left\vert \lambda_2^{{\scaleto{\widetilde{W}}{5pt}}} \right\vert^2 \leq \frac{2}{3}$ and $ \frac{2}{3} < \left\vert \lambda_2^{{\scaleto{\widetilde{W}}{5pt}}} \right\vert^2 < 1$, respectively. 
Therefore, the constant $\gamma_{{\scaleto{\widetilde{W}}{5pt}}}>0$.
Thus, we can show from~\eqref{tildex:t7} that for every $K\geq 0$ it holds that:
\begin{align}
\label{tildex:t8}
\gamma_{{\scaleto{\widetilde{W}}{5pt}}}\left\Vert \widetilde{\mathbf{x}} \right\Vert_2^{\delta, K} 
&\leq \left(\delta^2 - \left\vert\lambda_2^{{\scaleto{\widetilde{W}}{5pt}}}\right\vert^2 - \eta\mu\left(3(L/\mu) + 3\eta L - 1\right) \right)  \left\Vert \widetilde{\mathbf{x}} \right\Vert_2^{\delta, K} 
\nonumber 
\\
& \leq \eta\mu \left(L/\mu + 3\eta L - 1\right)\left\Vert \overline{\mathbf{e}}_x \right\Vert_2^{\delta, K} 
+ \eta\left(\frac{1}{2L} + \eta + \frac{\eta}{2L\mu}\right)\left\Vert \widetilde{\mathbf{v}} \right\Vert_2^{\delta, K} 
\nonumber 
\\
 & \qquad\qquad + \frac{\eta}{\delta^{2K-2}}\left(N + \frac{1}{N} \right)(\eta\sigma^2 + 2d) + \delta^2\mathbb{E}\left[ \left\Vert \widetilde{x}^{(0)}\right\Vert^2 \right].
\end{align}
We can obtain the desired result by dividing $\gamma_{{\scaleto{\widetilde{W}}{5pt}}}$ on 
both hand sides of \eqref{tildex:t8}. The proof is complete.


\subsection{Proof of Lemma~\ref{lemma:tilde:xv}}

First of all, Lemma~\ref{lemma:avi:ex} implies that
\begin{align}
\label{avgx:bd}
\left\Vert \overline{\mathbf{e}}_x \right\Vert_2^{\delta, K}  
& \leq \eta\cdot\frac{L^2\left(\eta + \frac{1 + \eta L}{\mu\left(1-\frac{\eta L}{2}\right)} \right)}{N^2\left(\delta^2 +\eta\mu\left(1 - \frac{\eta L}{2}\right) - 1 \right)}\left\Vert \widetilde{\mathbf{x}}\right\Vert_2^{\delta, K} 
\nonumber 
\\
& \qquad\qquad + \frac{\eta}{N\delta^{2K-2}} \cdot \frac{\eta\sigma^2 + 2d}{\delta^2 +\eta\mu\left(1 - \frac{\eta L}{2}\right) - 1}
+ \frac{\delta^2}{\delta^2 +\eta\mu\left(1 - \frac{\eta L}{2}\right) - 1}\mathbb{E}\left[ \left\Vert \overline{e}_{x}^{(0)}\right\Vert^2 \right]
\nonumber 
\\
& \leq \frac{4L^2\left(1 + \frac{2 + 2L}{\mu} \right)}{N^2\mu}\left\Vert \widetilde{\mathbf{x}}\right\Vert_2^{\delta, K}  + \frac{4}{N\delta^{2K-2}} \cdot \frac{\eta\sigma^2 + 2d}{\mu}
+ \frac{4\delta^2}{\eta\mu}\mathbb{E}\left[ \left\Vert \overline{e}_{x}^{(0)}\right\Vert^2 \right],
\end{align}
where we used the assumption that $\eta \leq 1/L$ to get $\frac{1+\eta L}{\mu\left(1 - \frac{\eta L}{2}\right)} \leq \frac{1 + L}{\mu/2}$ in the first term, and then we chose $\delta^2$ such that $\delta^2 \geq 1 - \frac{\eta \mu}{2}\left(1 - \frac{\eta L}{2}\right)$ to get $\delta^2 - 1 + \eta\mu\left(1 - \frac{\eta L}{2} \right) \geq \frac{\eta \mu}{2}\left(1 - \frac{\eta L}{2}\right) \geq \frac{\eta\mu}{4}$ where we used $\eta \leq 1/L$ again. By substituting the upper bound of $\left\Vert \overline{\mathbf{e}}_x \right\Vert_2^{\delta,K}$ in Lemma~\ref{lemma:avi:ex} to Lemma~\ref{lemma:tilde:x}, we can compute that 
\begin{align}
\label{tilde:xv:t1}
\left\Vert \widetilde{\mathbf{x}} \right\Vert_2^{\delta, K} 
& \leq \frac{\eta}{\gamma_{{\scaleto{\widetilde{W}}{5pt}}}} \left(L/\mu + 3\eta L - 1 \right) \frac{4L^2\left(1 + \frac{2 + 2L}{\mu} \right)}{N^2} \left\Vert \widetilde{\mathbf{x}} \right\Vert_2^{\delta, K}  
\nonumber 
\\
& \quad +  \frac{\eta}{\gamma_{{\scaleto{\widetilde{W}}{5pt}}}}\left(\frac{1}{2L} + \eta + \frac{\eta}{2L\mu}\right)\left\Vert \widetilde{\mathbf{v}} \right\Vert_2^{\delta, K}
+ \frac{\eta}{\delta^{2K-2}}(\eta\sigma^2 + 2d)\left(\frac{N + \frac{1}{N}}{\gamma_{{\scaleto{\widetilde{W}}{5pt}}}} + \frac{4}{N\gamma_{{\scaleto{\widetilde{W}}{5pt}}}}\cdot \left(L/\mu + 3\eta L - 1\right) \right) 
\nonumber 
\\
& \quad\quad + \frac{4\delta^2}{\gamma_{{\scaleto{\widetilde{W}}{5pt}}}} \left(L/\mu + 3\eta L - 1\right)\mathbb{E}\left[ \left\Vert \overline{e}_{x}^{(0)}\right\Vert^2 \right]  + \frac{\delta^2}{\gamma_{{\scaleto{\widetilde{W}}{5pt}}}}\mathbb{E}\left[ \left\Vert \widetilde{x}^{(0)}\right\Vert^2 \right].
\end{align}
Recall the definition of $A$ in \eqref{const:A} such that
\begin{align}
\label{tildex:const:A}
A  = \left(L/\mu - 1 + \frac{\gamma_{{\scaleto{\widetilde{W}}{5pt}}}}{2(1+\mu/L)}\right)\cdot \frac{4L^2}{N^2}\left(1 + \frac{2 + 2L}{\mu} \right)
\geq \left(L/\mu + 3\eta L-1\right) \cdot \frac{4L^2}{N^2}\left(1 + \frac{2 + 2L}{\mu} \right),
\end{align}
where we used $\eta \leq \frac{\gamma_{{\scaleto{\widetilde{W}}{5pt}}}}{6(L + \mu)}$ under the assumption in Lemma~\ref{lemma:tilde:x}. Furthermore, by using the assumption that $
\eta \leq \frac{\gamma_{{\scaleto{\widetilde{W}}{5pt}}}}{2A}$, we can further compute $1 - \frac{\eta A}{\gamma_{{\scaleto{\widetilde{W}}{5pt}}}} \geq \frac{1}{2}$. Hence, we can compute from~\eqref{tilde:xv:t1} that  
\begin{align}
\frac{\left\Vert \widetilde{\mathbf{x}} \right\Vert_2^{\delta, K} }{2} 
& \leq \left(1 - \frac{\eta A}{\gamma_{{\scaleto{\widetilde{W}}{5pt}}}} 
 \right)\left\Vert \widetilde{\mathbf{x}} \right\Vert_2^{\delta, K} 
\nonumber
\\
&\leq \frac{\eta}{\gamma_{{\scaleto{\widetilde{W}}{5pt}}}}\left(\frac{1}{2L} + \eta + \frac{\eta}{2L\mu}\right)\left\Vert \widetilde{\mathbf{v}} \right\Vert_2^{\delta, K}
+ \frac{\eta}{\delta^{2K-2}}(\eta\sigma^2 + 2d)\left(\frac{N + \frac{1}{N}}{\gamma_{{\scaleto{\widetilde{W}}{5pt}}}} + \frac{4}{N\gamma_{{\scaleto{\widetilde{W}}{5pt}}}}\cdot \left(L/\mu + 3\eta L - 1\right) \right) 
\nonumber 
\\
& \qquad\qquad + \frac{4\delta^2}{\gamma_{{\scaleto{\widetilde{W}}{5pt}}}} \left(L/\mu + 3\eta L - 1\right)\mathbb{E}\left[ \left\Vert \overline{e}_{x}^{(0)}\right\Vert^2 \right]  + \frac{\delta^2}{\gamma_{{\scaleto{\widetilde{W}}{5pt}}}}\mathbb{E}\left[ \left\Vert \widetilde{x}^{(0)}\right\Vert^2 \right].\label{both:hand:side:multiply}
\end{align}
The proof is complete by multiplying $2$ on both hand sides of \eqref{both:hand:side:multiply}.



\subsection{Proof of Lemma~\ref{lemma:tilde:vx}}

Considering~\eqref{iter2} and the choice of $\widetilde{W}$ in~\eqref{choice:tilde:W}, we can compute that
\begin{equation}
U = \widetilde{W} - W = h I_{N} + \left(1 - h\right)W - W = h\left(I_N - W\right), \quad h \in (0,1/2].
\end{equation}
From Lemma~\ref{lemma:err} and~\eqref{def:err}, and under our assumption that $\mathcal{B} = B \otimes I_d$ with $1^T_N B = c$, we have $\mathcal{U}\mathcal{B}\mathbf{x}_* = c\,\mathcal{U}\mathbf{x}_* = 0$, we can compute
\begin{align}
& \left\Vert \widetilde{v}^{(k+1)} \right\Vert^2 = \left\Vert e_v^{(k+1)} \right\Vert^2  = \left\Vert v^{(k+1)} + \nabla F(\mathbf{x}_*) \right\Vert^2
\nonumber 
\\
& \quad = \Bigg\Vert \widetilde{v}^{(k)} -h(I_{Nd} -\mathcal{W})\left(\widetilde{v}^{(k)} + \nabla F\left(x^{(k)}\right) - \nabla F\left(\mathbf{x}_*\right) - \mathcal{B}\left(x^{(k)}-\mathbf{x}_*\right) \right) 
\nonumber 
\\
& \qquad\qquad\qquad\qquad\qquad\qquad\qquad\qquad   - h(I_{Nd} -\mathcal{W})\xi^{(k)} + (h/\eta)(I_{Nd} -\mathcal{W})\sqrt{2\eta}w^{(k+1)} \Bigg\Vert^2
\label{tilde:v:t}
\\
& \quad = \Bigg\Vert \left((1-h)I_{Nd} + h\mathcal{W}\right)\widetilde{v}^{(k)} - \Bigg[h(I_{Nd} -\mathcal{W})\left(\nabla F\left(x^{(k)}\right) - \nabla F\left(\mathbf{x}_*\right)\right)
\nonumber 
\\
& \qquad\qquad  - h(I_{Nd} -\mathcal{W})\mathcal{B}\left(x^{(k)}-\mathbf{x}_*\right) + h\left(I_{Nd}- \mathcal{W}\right)\xi^{(k)} - (h/\eta)\left(I_{Nd}- \mathcal{W}\right)\sqrt{2\eta}w^{(k+1)}\Bigg] \Bigg\Vert^2
\nonumber 
\\
& \quad = \left\Vert\left((1-h)I_{Nd} + h\mathcal{W}\right)\widetilde{v}^{(k)}\right\Vert^2 
\label{tilde:v:t0} 
\\
& \qquad\qquad + \Bigg\Vert h(I_{Nd} -\mathcal{W})\left(\nabla F\left(x^{(k)}\right) - \nabla F\left(\mathbf{x}_*\right)\right)- h(I_{Nd} -\mathcal{W})\mathcal{B}\left(x^{(k)}-\mathbf{x} _*\right) \Bigg\Vert^2
\label{tilde:v:t1}
\\
& \qquad\qquad  + \Bigg\Vert h\left(I_{Nd}- \mathcal{W}\right)\xi^{(k)} - (h/\eta)\left(I_{Nd}- \mathcal{W}\right)\sqrt{2\eta}w^{(k+1)} \Bigg\Vert^2
\nonumber 
\\
& \qquad\qquad + 2\Bigg\langle h(I_{Nd} -\mathcal{W})\left(\nabla F\left(x^{(k)}\right) - \nabla F\left(\mathbf{x}_*\right)\right)- h(I_{Nd} -\mathcal{W})\mathcal{B}\left(x^{(k)}-\mathbf{x} _*\right) \,,\, 
\nonumber 
\\
& \qquad\qquad\qquad\qquad h\left(I_{Nd}- \mathcal{W}\right)\xi^{(k)} - (h/\eta)\left(I_{Nd}- \mathcal{W}\right)\sqrt{2\eta}w^{(k+1)}\Bigg\rangle
\nonumber 
\\
& \qquad\qquad - 2\Bigg\langle \left((1-h)I_{Nd} + h\left(\mathcal{W-J}\right)\right)\widetilde{v}^{(k)}\,,\, h(I_{Nd} -\mathcal{W})\left(\nabla F\left(x^{(k)}\right) - \nabla F\left(\mathbf{x}_*\right)\right)
\nonumber 
\\
& \qquad\qquad\qquad\qquad - h(I_{Nd} -\mathcal{W})\mathcal{B}\left(x^{(k)}-\mathbf{x} _*\right)\Bigg\rangle 
\label{tilde:v:t2} 
\\
& \qquad\qquad + 
2\Bigg\langle \left((1-h)I_{Nd} + h\left(\mathcal{W-J}\right)\right)\widetilde{v}^{(k)} \,,\,h\left(I_{Nd}- \mathcal{W}\right)\xi^{(k)} - (h/\eta)\left(I_{Nd}- \mathcal{W}\right)\sqrt{2\eta}w^{(k+1)}\Bigg\rangle.
\nonumber 
\end{align}
Since $\mathcal{J}\widetilde{v}^{(k)} = \mathcal{J}e_v^{(k)} = \overline{e}_v^{(k)} \otimes I_d = 0$ by Lemma~\ref{lemma:err}, we compute~\eqref{tilde:v:t0} such that 
\begin{align}
\left\Vert\left((1-h)I_{Nd} + h\mathcal{W}\right)\widetilde{v}^{(k)}\right\Vert^2
&= \left\Vert\left((1-h)I_{Nd} + h\left(\mathcal{W-J}\right)\right)\widetilde{v}^{(k)}\right\Vert^2
\nonumber 
\\
&\leq \left\Vert (1-h)I_{N} + h\left(W - J\right) \right\Vert^2 \cdot \left\Vert \widetilde{v}^{(k)} \right\Vert^2
\nonumber 
\\
&  \leq \left(1 - 2h + h^2 + 2(1-h)h \overline{\gamma}_{{\scaleto{W}{3pt}}} + h^2\overline{\gamma}_{{\scaleto{W}{3pt}}}^2\right)\left\Vert \widetilde{v}^{(k)} \right\Vert^2 
\nonumber 
\\
&= \left(1 - 2h\left(1 - \overline{\gamma}_{{\scaleto{W}{3pt}}}\right) + h^2\left(1 - \overline{\gamma}_{{\scaleto{W}{3pt}}}\right)^2\right)\left\Vert \widetilde{v}^{(k)} \right\Vert^2
\nonumber 
\\
&\leq \left(1 - h(1-\overline{\gamma}_{{\scaleto{W}{3pt}}})\right)\left\Vert \widetilde{v}^{(k)} \right\Vert^2,
\end{align}
where we used the assumption $h < 1 < \frac{1}{1-\overline{\gamma}_{{\scaleto{W}{3pt}}}}$ in the last inequality. Then by $L$-smoothness of $F$, we can bound the term~\eqref{tilde:v:t1} as follows:
\begin{align}
& \Bigg\Vert h(I_{Nd} -\mathcal{W})\left(\nabla F\left(x^{(k)}\right) - \nabla F\left(\mathbf{x}_*\right)\right)- h(I_{Nd} -\mathcal{W})\mathcal{B}\left(x^{(k)}-\mathbf{x} _*\right) \Bigg\Vert^2
\nonumber
\\
&\leq
2\Bigg\Vert h(I_{Nd} -\mathcal{W})\left(\nabla F\left(x^{(k)}\right) - \nabla F\left(\mathbf{x}_*\right)\right)\Bigg\Vert^2
+2 \Bigg\Vert h(I_{Nd} -\mathcal{W})\mathcal{B}\left(x^{(k)}-\mathbf{x} _*\right) \Bigg\Vert^2
\nonumber
\\
&\leq
2\overline{\gamma}_{{\scaleto{I_{N}-W}{5pt}}}^2\left(h^{2}L^{2}+\Vert hB\Vert^{2}\right)\left\Vert e_x^{(k)} \right\Vert^2.
\label{tilde:v:t4}
\end{align}

Next, we compute the inner product term in~\eqref{tilde:v:t2}.
\begin{align}
&  2\Bigg\langle \left((1-h)I_{Nd} + h\mathcal{W}\right)\widetilde{v}^{(k)}\,,\,  h(I_{Nd} -\mathcal{W})\mathcal{B}\left(x^{(k)}-\mathbf{x} _*\right)\Bigg\rangle 
\nonumber 
\\
& \qquad\qquad -2\Bigg\langle \left((1-h)I_{Nd} + h\mathcal{W}\right)\widetilde{v}^{(k)}\,,\, h(I_{Nd} -\mathcal{W})\left(\nabla F\left(x^{(k)}\right) - \nabla F\left(\mathbf{x}_*\right)\right) \Bigg\rangle
\nonumber 
\\
& \quad \leq 2\left\langle (I_{Nd} - \mathcal{W})\widetilde{v}^{(k)}\,,\, h\mathcal{B}\left(x^{(k)} - \mathbf{x}_* \right) - h\left(\nabla F\left(x^{(k)}\right) - \nabla F\left(\mathbf{x}_*\right)\right) \right\rangle
\label{tilde:v:t3}
\\
& \qquad\qquad - 2\Bigg\langle h(I_{Nd} - \mathcal{W})\widetilde{v}^{(k)}\,,\,  h(I_{Nd} -\mathcal{W})\mathcal{B}\left(x^{(k)}-\mathbf{x} _*\right)\Bigg\rangle
\nonumber 
\\
& \qquad\qquad - 2\Bigg\langle h(I_{Nd} - \mathcal{W})\widetilde{v}^{(k)}\,,\,  h(I_{Nd} -\mathcal{W})\left(\nabla F\left(x^{(k)}\right) - \nabla F\left(\mathbf{x}_*\right)\right) \Bigg\rangle.
\end{align}
We can use the $L$-smoothness of $F$ to bound the term~\eqref{tilde:v:t3}. 
It follows that
\begin{align}
& 2\left\langle (I_{Nd} - \mathcal{W})\widetilde{v}^{(k)}\,,\, h\mathcal{B}\left(x^{(k)} - \mathbf{x}_* \right) - h\left(\nabla F\left(x^{(k)}\right) - \nabla F\left(\mathbf{x}_*\right)\right) \right\rangle
\nonumber 
\\
&\leq 2\left\Vert (I_{Nd} - \mathcal{W})\widetilde{v}^{(k)} \right\Vert \cdot \left\Vert h\mathcal{B}\left(x^{(k)} - \mathbf{x}_* \right) - h\left(\nabla F\left(x^{(k)}\right) - \nabla F\left(\mathbf{x}_*\right)\right) \right\Vert
\nonumber 
\\
&\leq (1/c)\left\Vert (I_{Nd} - \mathcal{W})\widetilde{v}^{(k)} \right\Vert^{2}
+c \left\Vert h\mathcal{B}\left(x^{(k)} - \mathbf{x}_* \right) - h\left(\nabla F\left(x^{(k)}\right) - \nabla F\left(\mathbf{x}_*\right)\right) \right\Vert^{2}
\nonumber
\\
&\leq (1/c)\left\Vert (I_{Nd} - \mathcal{W})\widetilde{v}^{(k)} \right\Vert^{2}
+2c \left\Vert h\mathcal{B}\left(x^{(k)} - \mathbf{x}_* \right)\right\Vert^{2}+2c\left\Vert h\left(\nabla F\left(x^{(k)}\right) - \nabla F\left(\mathbf{x}_*\right)\right) \right\Vert^{2}
\nonumber
\\
&\leq (1/c)\overline{\gamma}_{{\scaleto{I_{N}-W}{5pt}}}^2\left\Vert \widetilde{v}^{(k)}  \right\Vert^2 
+2c\Vert hB\Vert^{2}\left\Vert e_x^{(k)} \right\Vert^2+2ch^{2}L^{2}\left\Vert e_x^{(k)} \right\Vert^2,
\end{align}
where we used the inequality $2xy \leq cx^2 + y^2/c$ for any $c > 0$ and $x,y\in\mathbb{R}$. 
Therefore, we can compute the inner product term~\eqref{tilde:v:t2} as follows.
\begin{align}
& 2\Bigg\langle \left((1-h)I_{Nd} + h\mathcal{W}\right)\widetilde{v}^{(k)}\,,\,  h(I_{Nd} -\mathcal{W})\mathcal{B}\left(x^{(k)}-\mathbf{x} _*\right)\Bigg\rangle 
\nonumber 
\\
& \qquad\qquad -2\Bigg\langle \left(1-h)I_{Nd} + h\mathcal{W}\right)\widetilde{v}^{(k)}\,,\, h(I_{Nd} -\mathcal{W})\left(\nabla F\left(x^{(k)}\right) - \nabla F\left(\mathbf{x}_*\right)\right) \Bigg\rangle
\nonumber 
\\
& \quad \leq (1/c)\overline{\gamma}_{{\scaleto{I_{N}-W}{5pt}}}^2\left\Vert \widetilde{v}^{(k)}  \right\Vert^2 
+2c\Vert hB\Vert^{2}\left\Vert e_x^{(k)} \right\Vert^2+2ch^{2}L^{2}\left\Vert e_x^{(k)} \right\Vert^2 
\nonumber 
\\
& \qquad\qquad + h^2\overline{\gamma}_{{\scaleto{I_{N}-W}{5pt}}}^2\left\Vert \widetilde{v}^{(k)} \right\Vert^2 + \left\Vert hB\right\Vert^2\overline{\gamma}_{{\scaleto{I_{N}-W}{5pt}}}^2\left\Vert e_x^{(k)} \right\Vert^2
+ h^2\overline{\gamma}_{{\scaleto{I_{N}-W}{5pt}}}^2\left\Vert \widetilde{v}^{(k)} \right\Vert^2 + h^2L^2\overline{\gamma}_{{\scaleto{I_{N}-W}{5pt}}}^2\left\Vert e_x^{(k)} \right\Vert^2
\nonumber 
\\
& \quad = (1/c)\overline{\gamma}_{{\scaleto{I_{N}-W}{5pt}}}^2\left\Vert \widetilde{v}^{(k)}  \right\Vert^2 + 2h^2\overline{\gamma}_{{\scaleto{I_{N}-W}{5pt}}}^2\left\Vert \widetilde{v}^{(k)} \right\Vert^2
\nonumber 
\\
& \qquad\qquad + 2ch^2_{\eta}L^2\left\Vert e_x^{(k)} \right\Vert^2 + h^2L^2\overline{\gamma}_{{\scaleto{I_{N}-W}{5pt}}}^2\left\Vert e_x^{(k)} \right\Vert^2
+ 2cL\left\Vert hB\right\Vert^2\left\Vert e_x^{(k)} \right\Vert^2 + \left\Vert hB \right\Vert^2\overline{\gamma}_{{\scaleto{I_{N}-W}{5pt}}}^2\left\Vert e_x^{(k)} \right\Vert^2 
\nonumber 
\\
& \quad \leq (1/c)\overline{\gamma}_{{\scaleto{I_{N}-W}{5pt}}}^2\left\Vert \widetilde{v}^{(k)}  \right\Vert^2 + 2h^2\overline{\gamma}_{{\scaleto{I_{N}-W}{5pt}}}^2\left\Vert \widetilde{v}^{(k)} \right\Vert^2 +  \left(3ch^2L^2 + 3cL\left\Vert hB\right\Vert^2 \right)\left\Vert e_x^{(k)} \right\Vert^2,
\end{align}
where we assume that $c \geq \overline{\gamma}_{I_N - W}^2$.
Now we take the expectation of~\eqref{tilde:v:t} to get
\begin{align}
\mathbb{E}\left[\left\Vert \widetilde{v}^{(k+1)} \right\Vert^2\right] 
& \leq \left(1 - h(1-\overline{\gamma}_{{\scaleto{W}{3pt}}}) + (1/c)\overline{\gamma}_{{\scaleto{I_{N}-W}{5pt}}}^2 + 2h^2\overline{\gamma}_{{\scaleto{I_{N}-W}{5pt}}}^2 \right)\mathbb{E}\left[\left\Vert \widetilde{v}^{(k)} \right\Vert^2 \right]
\nonumber 
\\
& \qquad\qquad + \left(3ch^2L^2 + 3cL\left\Vert hB\right\Vert^2\right)\mathbb{E}\left[\left\Vert e_x^{(k)} \right\Vert^2\right] + 2h^2\sigma^2N + 4\eta(h/\eta)^2dN.
\nonumber 
\\
& \leq \left(1 - \frac{h(1-\overline{\gamma}_{{\scaleto{W}{3pt}}})}{2} + (1/c)\overline{\gamma}_{{\scaleto{I_{N}-W}{5pt}}}^2 \right)\mathbb{E}\left[\left\Vert \widetilde{v}^{(k)} \right\Vert^2 \right]
\nonumber 
\\
& \qquad\qquad + \left(3ch^2L^2 + 3cL\left\Vert hB\right\Vert^2\right)\mathbb{E}\left[\left\Vert e_x^{(k)} \right\Vert^2\right] + 2h^2\sigma^2N + 4\eta(h/\eta)^2dN,
\end{align}
under the assumption such that $h \leq \frac{1-\overline{\gamma}_{{\scaleto{W}{3pt}}}}{4\overline{\gamma}_{{\scaleto{I_{N}-W}{5pt}}}^2}$. 
By taking $c = \frac{2\overline{\gamma}_{{\scaleto{I_{N}-W}{5pt}}}}{(1-\overline{\gamma}_{{\scaleto{W}{3pt}}})h}$ where $h \leq 1 < \frac{1}{\overline{\gamma}_{{\scaleto{I_{N}-W}{5pt}}}^2}$ under our assumptions, we can compute
\begin{align}
1 - \frac{h(1-\overline{\gamma}_{{\scaleto{W}{3pt}}})}{2} + (1/c)\overline{\gamma}_{{\scaleto{I_{N}-W}{5pt}}}^2 
& = 1 - \frac{h(1-\overline{\gamma}_{{\scaleto{W}{3pt}}})}{2} + \frac{(1-\overline{\gamma}_{{\scaleto{W}{3pt}}})h}{2}\overline{\gamma}_{{\scaleto{I_{N}-W}{5pt}}} 
\nonumber 
\\
& \leq 1 - h\frac{1-\overline{\gamma}_{{\scaleto{W}{3pt}}}}{2}\left(1 - \overline{\gamma}_{{\scaleto{I_{N}-W}{5pt}}}\right) > 0,
\end{align} 
where $1-\overline{\gamma}_{{\scaleto{W}{3pt}}} \leq \overline{\gamma}_{I_N - W} < 1$ by definition~\eqref{defn:bar:gamma}, then we can find the constant
\begin{equation}
\delta^2 \geq 1 - \frac{h}{2}\frac{1-\overline{\gamma}_{{\scaleto{W}{3pt}}}}{2}\left(1 - \overline{\gamma}_{{\scaleto{I_{N}-W}{5pt}}}\right) > 0.
\end{equation}
Therefore, by Lemma~\ref{lemma:inf:seq}, we get
\begin{align}
\frac{h}{2}\frac{1-\overline{\gamma}_{{\scaleto{W}{3pt}}}}{2}\left(1 - \overline{\gamma}_{{\scaleto{I_{N}-W}{5pt}}}\right)\left\Vert \widetilde{\mathbf{v}}\right\Vert_2^{\delta, K} 
&\leq \left(\delta^2 - \left(1 - h\frac{1-\overline{\gamma}_{{\scaleto{W}{3pt}}}}{2}\left(1 - \overline{\gamma}_{{\scaleto{I_{N}-W}{5pt}}}\right)\right) \right)\left\Vert \widetilde{\mathbf{v}}\right\Vert_2^{\delta, K} \label{tildevx:t1}
\\
&\leq   \left(3h(h/\eta)L^2 + 3h(h/\eta)L\left\Vert B\right\Vert^2\right)\left\Vert \mathbf{e}_x \right\Vert_2^{\delta, K}
\nonumber 
\\
& \qquad\qquad  + h^2 \cdot \frac{2\sigma^2N}{\delta^{2K-2}} + \eta\cdot (h/\eta)^2 \frac{4dN}{\delta^{2K-2}} + \delta^2\left\Vert \widetilde{v}^{(0)} \right\Vert^{2}.
\nonumber
\end{align}
Then it follows that
\begin{align}
\left\Vert \widetilde{\mathbf{v}}\right\Vert_2^{\delta, K} & \leq \frac{12(h/\eta)\left(L^2 + L\left\Vert B \right\Vert^2 \right)}{(1-\overline{\gamma}_{{\scaleto{W}{3pt}}})\left(1 - \overline{\gamma}_{{\scaleto{I_{N}-W}{5pt}}}^2\right)}\left\Vert \mathbf{e}_x \right\Vert_2^{\delta, K} + \frac{8N(h/\eta)}{(1-\overline{\gamma}_{{\scaleto{W}{3pt}}})\left(1 - \overline{\gamma}_{{\scaleto{I_{N}-W}{5pt}}}^2\right)}\frac{\eta\sigma^2 + 2d}{\delta^{2K-2}} 
\nonumber 
\\
& \qquad\qquad + \frac{4\delta^2}{h(1-\overline{\gamma}_{{\scaleto{W}{3pt}}})\left(1 - \overline{\gamma}_{{\scaleto{I_{N}-W}{5pt}}}^2\right)}\left\Vert \widetilde{v}^{(0)} \right\Vert^{2},\label{tilde:v:upper:bound}
\end{align}
where we note that $h/\eta$ is in the order of $\eta^{\alpha}$ under our assumption. 
Moreover, by Lemma~\ref{lemma:avi:ex}, see also~\eqref{avgx:bd}, we have
\begin{align}
\left\Vert \overline{\mathbf{e}}_x \right\Vert_2^{\delta, K}  
& \leq \frac{4L^2\left(1 + \frac{2 + 2L}{\mu} \right)}{N^2\mu}\left\Vert \widetilde{\mathbf{x}}\right\Vert_2^{\delta, K}  + \frac{4}{N\delta^{2K-2}} \cdot \frac{\eta\sigma^2 + 2d}{\mu}
+ \frac{4\delta^2}{\eta\mu}\mathbb{E}\left[ \left\Vert \overline{e}_{x}^{(0)}\right\Vert^2 \right],
\end{align}
Hence, by $\widetilde{x}^{(k)} + 1_N \otimes \overline{e}_x^{(k)} = e_x^{(k)}$, we can compute that
\begin{align}
\left\Vert \mathbf{e}_x \right\Vert_2^{\delta, K}  
& = \left\Vert \widetilde{\mathbf{x}} \right\Vert_2^{\delta, K} + \left\Vert \overline{\mathbf{e}}_x \right\Vert_2^{\delta, K}
\nonumber 
\\
& \leq \left(1 + \frac{4L^2\left(1 + \frac{2 + 2L}{\mu} \right)}{N^2\mu}\right)\left\Vert \widetilde{\mathbf{x}}\right\Vert_2^{\delta, K} + \frac{4}{N\mu} \cdot \frac{\eta\sigma^2 + 2d}{\delta^{2K-2}}
+ \frac{4\delta^2}{\eta\mu}\mathbb{E}\left[ \left\Vert \overline{e}_{x}^{(0)}\right\Vert^2 \right].\label{e:x:substitute}
\end{align}
Therefore, we can substitute the formula \eqref{e:x:substitute} into the upper bound of $\left\Vert \widetilde{\mathbf{v}} \right\Vert_2^{\delta,K}$ in \eqref{tilde:v:upper:bound} to get
\begin{align}
\left\Vert \widetilde{\mathbf{v}}\right\Vert_2^{\delta, K} & \leq \frac{12(h/\eta)\left(L^2 + L\left\Vert B \right\Vert^2 \right)}{(1-\overline{\gamma}_{{\scaleto{W}{3pt}}})\left(1 - \overline{\gamma}_{{\scaleto{I_{N}-W}{5pt}}}^2\right)}\left\Vert \mathbf{e}_x \right\Vert_2^{\delta, K} 
 + \frac{8N(h/\eta)}{(1-\overline{\gamma}_{{\scaleto{W}{3pt}}})\left(1 - \overline{\gamma}_{{\scaleto{I_{N}-W}{5pt}}}^2\right)}\frac{\eta\sigma^2 + 2d}{\delta^{2K-2}} 
\nonumber 
\\
& \qquad\qquad + \frac{4\delta^2}{h(1-\overline{\gamma}_{{\scaleto{W}{3pt}}})\left(1 - \overline{\gamma}_{{\scaleto{I_{N}-W}{5pt}}}^2\right)}\left\Vert \widetilde{v}^{(0)} \right\Vert^{2}
\nonumber 
\\
& \leq \frac{12(h/\eta)\left(L^2 + L\left\Vert B \right\Vert^2 \right)}{(1-\overline{\gamma}_{{\scaleto{W}{3pt}}})\left(1 - \overline{\gamma}_{{\scaleto{I_{N}-W}{5pt}}}^2\right)}\left(1 + \frac{4L^2\left(1 + \frac{2 + 2L}{\mu} \right)}{N^2\mu}\right)\left\Vert \widetilde{\mathbf{x}}\right\Vert_2^{\delta, K} 
\nonumber 
\\
& \qquad
 + \left(\frac{6\left(L^2 + L\left\Vert B \right\Vert^2 \right)}{N\mu} + N\right)\cdot \frac{8(h/\eta)}{(1-\overline{\gamma}_{{\scaleto{W}{3pt}}})\left(1 - \overline{\gamma}_{{\scaleto{I_{N}-W}{5pt}}}^2\right)} \cdot \frac{\eta\sigma^2 + 2d}{\delta^{2K-2}} 
 \nonumber
 \\
 & \qquad\qquad+ \frac{12\delta^2(h/\eta)\left(L^2 + L\left\Vert B \right\Vert^2 \right)}{\eta\mu(1-\overline{\gamma}_{{\scaleto{W}{3pt}}})\left(1 - \overline{\gamma}_{{\scaleto{I_{N}-W}{5pt}}}^2\right)} \mathbb{E}\left[ \left\Vert \overline{e}_{x}^{(0)}\right\Vert^2 \right] + \frac{4\delta^2}{h(1-\overline{\gamma}_{{\scaleto{W}{3pt}}})\left(1 - \overline{\gamma}_{{\scaleto{I_{N}-W}{5pt}}}^2\right)}\left\Vert \widetilde{v}^{(0)} \right\Vert^{2}
.
\end{align}
The proof is complete.


\subsection{Proof of Lemma~\ref{lemma:bound:grad:tilde:v}}
We first use~\eqref{def:tildex:tildev} to get the following first inequality, and then we use the uniform upper bound for $\left\Vert \widetilde{\mathbf{v}}\right\Vert_2^{\delta,k} $ from~\eqref{uniform:bound:v} in Theorem~\ref{theorem:norm:tilde} to derive as follows.
\begin{align}
& \frac{1}{\delta^{2k}}\mathbb{E}\left[\left\Vert \widetilde{v}^{(k)}\right\Vert^{2}\right]
\nonumber 
\\
& \quad \leq \left\Vert \widetilde{\mathbf{v}}\right\Vert_2^{\delta,k} 
\nonumber 
\\
& \quad \leq\frac{h\eta}{\delta^{2k-2}} \cdot \frac{\gamma_2\left(w_2\gamma_1(h/\eta) + w_1\right)\sigma^2/N}{1 - h\gamma_1\gamma_2} + \frac{h}{\delta^{2k-2}}\cdot \Bigg[\frac{2\gamma_2d\left(w_2\gamma_1(h/\eta) + w_1\right)/N}{1 - h\gamma_1\gamma_2} + \frac{w_2\sigma^2}{N}
\nonumber 
\\
& \qquad\qquad+ \frac{\gamma_1\gamma_2}{1-h\gamma_1\gamma_2}\delta^{2k}\left((h/\eta)(E_3/\eta)\mathbb{E}\left[ \left\Vert \overline{e}_{x}^{(0)}\right\Vert^2 \right] +(E_4/h)\mathbb{E}\left[ \left\Vert \widetilde{v}^{(0)}\right\Vert^2 \right] \right)\Bigg] 
\nonumber 
\\
& \qquad\qquad + (h/\eta)\delta^2\left(\frac{2dw_2}{N\delta^{2k}} +\gamma_2D_0 \right) + \delta^2(h/\eta)(E_3/\eta)\mathbb{E}\left[ \left\Vert \overline{e}_{x}^{(0)}\right\Vert^2 \right] + (E_4/h)\mathbb{E}\left[ \left\Vert \widetilde{v}^{(0)}\right\Vert^2 \right].
\end{align}
Therefore, we obtain:
\begin{align}
\mathbb{E}\left[\left\Vert \widetilde{v}^{(k)}\right\Vert^{2}\right]
& \leq h\delta^2 \cdot \left(\eta\sigma^2 + 2d\right) \cdot \frac{\gamma_2\left(w_2\gamma_1(h/\eta) + w_1\right)/N}{1 - h\gamma_1\gamma_2} + h\delta^2\cdot\frac{w_2\sigma^2}{N} + (h/\eta)\delta^2\frac{2dw_2}{N} 
\nonumber 
\\
& \qquad + \delta^{2k+2} \cdot \Bigg(\frac{h\gamma_1\gamma_2}{1 - h\gamma_1\gamma_2}\left((h/\eta)(E_3/\eta)\mathbb{E}\left[ \left\Vert \overline{e}_{x}^{(0)}\right\Vert^2 \right] + (E_4/h)\mathbb{E}\left[ \left\Vert \widetilde{v}^{(0)}\right\Vert^2 \right] \right)
\nonumber 
\\
& \qquad\qquad\qquad\qquad + (h/\eta)\gamma_2D_0 + (h/\eta)(E_3/\eta)\mathbb{E}\left[ \left\Vert \overline{e}_{x}^{(0)}\right\Vert^2 \right] +(E_4/h)\mathbb{E}\left[ \left\Vert \widetilde{v}^{(0)}\right\Vert^2 \right]\Bigg)
\nonumber 
\\
& 
=: h\delta^2 \cdot \frac{C_1\gamma_2}{2L^2} + \delta^{2k+2}h \cdot \frac{C_0\gamma_1\gamma_2}{2L^2}
\nonumber 
\\
& \qquad\qquad 
+ \delta^{2k+2} \cdot \left((h/\eta)\gamma_2D_0 + C_0\right) + (h/\eta)\delta^2 \cdot \frac{w_2}{N}\left(\eta\sigma^2 + 2d\right).
\end{align}
By uniform bound for $\left\Vert \widetilde{\mathbf{x}} \right\Vert_2^{\delta,k}$ in~\eqref{uniform:bound:x} in Theorem~\ref{theorem:norm:tilde}, we can obtain that 
\begin{align}
\label{nabla:F:bound:t1}
\delta^{2k}\left\Vert \widetilde{\mathbf{x}} \right\Vert_2^{\delta,k} & \leq \delta^2\eta \cdot \frac{\eta \sigma^2 + 2d}{N} \cdot \frac{w_2\gamma_1(h/\eta) + w_1}{1 - h\gamma_1\gamma_2} 
\nonumber 
\\
& \qquad + \delta^{2k+2}\eta \cdot \frac{\gamma_1}{1-h\gamma_1\gamma_2}\left((h/\eta)(E_3/\eta)\mathbb{E}\left[ \left\Vert \overline{e}_{x}^{(0)}\right\Vert^2 \right] + (E_4/h)\mathbb{E}\left[ \left\Vert \widetilde{v}^{(0)}\right\Vert^2 \right] \right) + \delta^{2k+2}D_0
\nonumber 
\\
& 
= \delta^2\eta \cdot \frac{C_1}{2L^2} + \delta^{2k+2}\eta \cdot \frac{\gamma_1C_0}{2L^2} + \delta^{2k+2}D_0.
\end{align}
Then we can further get the uniform bound for $\mathbb{E}\left[\left\Vert \nabla F\left(x^{(k)}\right) \right\Vert^2\right]$ in the following derivation.
\begin{align}
& \mathbb{E}\left[\left\Vert \nabla F\left(x^{(k)}\right) \right\Vert^2 \right]
\nonumber 
\\
& \quad \leq 2\mathbb{E}\left[\left\Vert \nabla F\left(x^{(k)}\right) - \nabla F\left(\mathbf{x}_*\right) \right\Vert^2\right]  + 2\left\Vert \nabla F\left(\mathbf{x}_*\right) \right\Vert^2 
\nonumber 
\\
& \quad \leq 2L^2\mathbb{E}\left[\left\Vert x^{(k)} - \mathbf{x}_* \right\Vert^2\right] + 2\left\Vert \nabla F\left(\mathbf{x}_*\right) \right\Vert^2
\nonumber 
\\
& \quad \leq 2L^2\delta^{2k}\left(\left\Vert\widetilde{\mathbf{x}} \right\Vert_2^{\delta,k} + \left\Vert \overline{\mathbf{e}}_x \right\Vert_2^{\delta,k}\right) + 2\left\Vert \nabla F(\mathbf{x}_*) \right\Vert^2
\nonumber 
\\
& \quad \leq \delta^{2k} \cdot 2L^2\Bigg( \left\Vert \widetilde{\mathbf{x}}\right\Vert_2^{\delta, k} +  \eta\cdot\frac{L^2}{N^2\left(\delta^2 +\eta\mu\left(1 - \frac{\eta L}{2}\right) - 1 \right)}\left(\eta + \frac{1 + \eta L}{\mu\left(1-\frac{\eta L}{2}\right)} \right)\left\Vert \widetilde{\mathbf{x}}\right\Vert_2^{\delta, k} 
\nonumber 
\\
& \qquad\qquad + \frac{\eta}{N\delta^{2k-2}} \cdot \frac{\eta\sigma^2 + 2d}{\delta^2 +\eta\mu\left(1 - \frac{\eta L}{2}\right) - 1}
+ \frac{\delta^2}{\delta^2 +\eta\mu\left(1 - \frac{\eta L}{2}\right) - 1}\mathbb{E}\left[ \left\Vert \overline{e}_{x}^{(0)}\right\Vert^2 \right]
\Bigg) + 2\left\Vert \nabla F(\mathbf{x}_*) \right\Vert^2
\nonumber 
\\
& \quad 
= 2L^2\left(\delta^2\eta \cdot \frac{C_1}{2L^2} + \delta^{2k+2}\eta \cdot \frac{\gamma_1C_0}{2L^2} + \delta^{2k+2}D_0\right) 
\nonumber
\\
&\qquad\qquad+ \eta \cdot \left(\delta^2\eta \cdot \frac{C_1C_2}{2L^2} + \delta^{2k+2}\eta \cdot \frac{\gamma_1C_0C_2}{2L^2} + \delta^{2k+2}D_0C_2\right)
\nonumber 
\\
& \qquad\qquad\qquad\qquad 
+ \delta^{2}\eta \cdot C_3 + \delta^{2k+2}\cdot C_4
+ 2\left\Vert \nabla F(\mathbf{x}_*) \right\Vert^2
\nonumber 
\\
& \quad 
= \eta\delta^2\left(C_1 + C_3\right) + \delta^2\eta^2\left(\frac{C_1C_2}{2L^2} \right) + \delta^{2k+2}\eta\left(\gamma_1C_0 + D_0C_2\right) 
\nonumber 
\\
& \qquad\qquad\qquad\qquad 
+ \delta^{2k+2}\eta^2\left(\frac{\gamma_1C_0C_2}{2L^2}\right) +\delta^{2k+2}\left(D_0 + C_4 \right)+ 2\left\Vert \nabla F(\mathbf{x}_*) \right\Vert^2,
\label{last:inequality}
\end{align}
with 
\begin{equation}
C_2:= \frac{2L^4}{N^2\left(\delta^2 +\eta\mu\left(1 - \frac{\eta L}{2}\right) - 1 \right)}\left(\eta + \frac{1 + \eta L}{\mu\left(1-\frac{\eta L}{2}\right)} \right),
\end{equation}
and
\begin{equation}
C_3:= \frac{2L^2}{N} \cdot \frac{\eta\sigma^2 + 2d}{\delta^2 +\eta\mu\left(1 - \frac{\eta L}{2}\right) - 1},\quad C_4:= \frac{2L^2}{\delta^2 +\eta\mu\left(1 - \frac{\eta L}{2}\right) - 1}\mathbb{E}\left[ \left\Vert \overline{e}_{x}^{(0)}\right\Vert^2 \right],
\end{equation}
where we used the fact that $x^{(k)} - \mathbf{x}_* = e^{(k)}_x = \widetilde{x}^{(k)} + 1_N \otimes \overline{e}^{(k)}_x$ in Lemma~\ref{lemma:err}, and the last inequality in \eqref{last:inequality} follows bound in~\eqref{nabla:F:bound:t1} above. The proof is complete.


\subsection{Proof of Corollary~\ref{cor:bound:grad:tilde:v}}

We proved in Lemma~\ref{lemma:bound:grad:tilde:v} such that 
\begin{align}
\mathbb{E}\left[\left\Vert \widetilde{v}^{(k)}\right\Vert^{2}\right]
& \leq h\delta^2 \cdot \frac{C_1\gamma_2}{2L^2} + \delta^{2k+2}h \cdot \frac{C_0\gamma_1\gamma_2}{2L^2}
\nonumber 
\\
& \qquad\qquad+ \delta^{2k+2} \cdot \left((h/\eta)\gamma_2D_0 + C_0\right) + (h/\eta)\delta^2 \cdot \frac{w_2}{N}\left(\eta\sigma^2 + 2d\right),
\\
\mathbb{E}\left[\left\Vert \nabla F\left(x^{(k)}\right) \right\Vert^2 \right] 
& \leq \eta\delta^2\left(C_1 + C_3\right) + \delta^2\eta^2\left(\frac{C_1C_2}{2L^2} \right) + \delta^{2k+2}\eta\left(\gamma_1C_0 + D_0C_2\right) 
\nonumber 
\\
& \qquad\qquad + \delta^{2k+2}\eta^2\left(\frac{\gamma_1C_0C_2}{2L^2}\right) +\delta^{2k+2}\left(D_0 + C_4 \right)+ 2\left\Vert \nabla F(\mathbf{x}_*) \right\Vert^2.
\end{align}
Since $0 < \delta < 1$, for any $K_0\geq 0$ such that for every $k \geq K_0 \geq 0$, it holds that
\begin{align}
\mathbb{E}\left[\left\Vert \widetilde{v}^{(k)} \right\Vert^2\right] & \leq  h\delta^2\Bigg(\frac{C_1\gamma_2}{2L^2} +  \frac{C_0\gamma_1\gamma_2}{2L^2} \Bigg) + (h/\eta)\delta^{2}\left( \gamma_2D_0 + \frac{w_2}{N}\left(\eta\sigma^2 + 2d\right) \right) + \delta^{2K_0}C_0,
\nonumber 
\\
\mathbb{E}\left[\left\Vert \nabla F\left(x^{(k)}\right) \right\Vert^2\right] &  \leq \eta\delta^2\left(C_1 + C_3 + \gamma_1C_0 + D_0C_2\right) + \delta^2\eta^2\left(\frac{C_1C_2}{2L^2} + \frac{\gamma_1C_0C_2}{2L^2}\right) 
\nonumber 
\\
& \qquad\qquad\qquad\qquad\qquad\qquad\qquad\qquad  + \delta^{2K_0} (D_0 + C_4) + 2\left\Vert \nabla F(\mathbf{x}_*)\right\Vert^2.
\end{align}
In particular, we use the inequality $1 - \frac{1}{x} \leq \log(x) \leq x - 1$, and choose the constant $K_0$ as follows:
\begin{equation}
K_0:=\frac{\delta^2}{1 - \delta^2}\left[\left(1 - \frac{\left\Vert \nabla F(\mathbf{x}_*) \right\Vert^2}{D_0 + C_4}\right)  \vee \left(1 - \frac{\left\Vert \nabla F(\mathbf{x}_*) \right\Vert^2}{C_0}\right)  \right] \vee 0,
\end{equation}
which induces that
\begin{equation}
\delta^{2K_0}C_0 \vee \delta^{2K_0}(D_0 + C_4) \leq \left\Vert \nabla F(\mathbf{x}_*) \right\Vert^2.
\end{equation}
Therefore, we can obtain the uniform bounds such that 
\begin{equation}
\mathbb{E}\left[\left\Vert \widetilde{v}^{(k)} \right\Vert^2\right] \leq R_h,\qquad \mathbb{E}\left[\left\Vert \nabla F\left(x^{(k)}\right) \right\Vert^2\right] \leq R_h', 
\end{equation}
for any $k \geq K_0$, where 
\begin{align}
& R_h:=  h\delta^2\left(\frac{C_1\gamma_2}{2L^2} +  \frac{C_0\gamma_1\gamma_2}{2L^2} \right) + (h/\eta)\delta^{2}\left( \gamma_2D_0 + \frac{w_2}{N}\left(\eta\sigma^2 + 2d\right) \right) + \left\Vert \nabla F(\mathbf{x}_*) \right\Vert^2,
\\
& R_h':= \eta\delta^2\left(C_1 + C_3 + \gamma_1C_0 + D_0C_2\right) + \delta^2\eta^2\left(\frac{C_1C_2}{2L^2} + \frac{\gamma_1C_0C_2}{2L^2}\right) 
 + 3\left\Vert \nabla F(\mathbf{x}_*)\right\Vert^2.
\end{align}
This completes the proof.

\end{document}